# Continuous Semantic Caching for Low-Cost LLM Serving

Baran Atalar
Carnegie Mellon University
Pittsburgh, Pennsylvania, USA
batalar@andrew.cmu.edu

Xutong Liu
University of Washington Tacoma
Tacoma, Washington, USA
xutongl@uw.edu

Jinhang Zuo
City University of Hong Kong
Hong Kong, China
jinhang.zuo@cityu.edu.hk

Siwei Wang
Microsoft Research Asia
Beijing, China
siweiwang@microsoft.com

Wei Chen
Microsoft Research Asia
Beijing, China
weic@microsoft.com

Carlee Joe-Wong
Carnegie Mellon University
Pittsburgh, Pennsylvania, USA
cjoewong@andrew.cmu.edu

## Abstract

As Large Language Models (LLMs) become increasingly popular, caching responses so that they can be reused by users with semantically similar queries has become a vital strategy for reducing inference costs and latency. Existing caching frameworks have proposed to decide which query responses to cache by assuming a finite, known universe of discrete queries and learning their serving costs and arrival probabilities. As LLMs' pool of users and queries expands, however, such an assumption becomes increasingly untenable: real-world LLM queries reside in an infinite, continuous embedding space. In this paper, we establish the first rigorous theoretical framework for semantic LLM response caching in continuous query space under uncertainty. To bridge the gap between discrete optimization and continuous representation spaces, we introduce dynamic $\epsilon$-net discretization coupled with Kernel Ridge Regression. This design enables the system to formally quantify estimation uncertainty and generalize partial feedback on LLM query costs across continuous semantic query neighborhoods. We develop both offline learning and online adaptive algorithms optimized to reduce switching costs incurred by changing the cached responses. We prove that our online algorithm achieves a sublinear regret bound against an optimal oracle, which reduces to existing bounds for discrete query models. Extensive empirical evaluations demonstrate that our framework approximates the continuous optimal cache well while also reducing computational and switching overhead compared to existing methods.

## CCS Concepts

• **Computing methodologies → Online learning settings**; • **Theory of computation → Caching and paging algorithms**; • **Information systems → Language models**.

## 1 Introduction

Large language models (LLMs) have recently exploded in popularity, with ChatGPT reportedly reaching 700 million users in July 2025 [3]. LLM queries, however, are expensive [24]: not only can they have seconds-level response latencies, but LLM inference can have significant energy costs and compute utilization, costs that will only grow with query volume. Prior work has suggested *caching* query responses to reduce these costs: cached responses can then be returned directly for semantically similar queries, eliminating the need for a LLM inference pass. Caching is particularly effective for queries that seek factual, non-user-specific information, which make up a significant portion of ChatGPT usage [3].

While caching presents a natural solution to reducing LLM inference costs, deciding *which queries and responses to cache*, which we refer to as the cache design problem, presents a significant **research challenge**: cache sizes are limited, so the most frequently arriving and/or most expensive query-response pairs should be prioritized for caching. These costs and arrival probabilities are often unknown and stochastic, and thus must be estimated from offline data or online experience. Prior work has addressed such challenges in other caching settings [17, 20], but the LLM setting raises novel challenges. Firstly, we should relate *semantically similar* queries in order to reduce cache misses, instead of relying on exact query matches like some existing LLM frameworks [32]. Indeed, some recent work has found that more than 30% of LLM queries are semantically similar [8]. Secondly, initial work on semantic caching for LLMs relies on the assumption that query-response caching candidates are drawn from a known, finite list [2, 8, 16, 28]. One can then apply online or offline learning methods to estimate the cost of serving each query, as well as its arrival probability, and choose what to cache accordingly [18, 19]. As LLMs grow more popular and their queries and users more diverse, it becomes almost impossible to enumerate such a finite set of queries and responses.

We address these challenges by considering semantic caching over *continuous query space*. Learning query costs and arrival probabilities is thus significantly harder: since future queries can come from anywhere in the continuous query space, the cost and arrival probability of each query must be *inferred from those of other queries* instead of directly estimated from the history of the same query. Even if these parameters can be inferred, we must now optimize over an *infinite* space of possible query-response cache candidates, potentially introducing significant suboptimality into the cache design. To the best of our knowledge, **ours is the first work to** **propose a semantic caching framework for low-cost LLM serving in continuous query space with theoretical performance guarantees**.

## 1.1 Related Work

**Low-Cost LLM Serving.** As the deployment of LLM models scales, lowering their high inference costs has become an important research direction. While optimization techniques such as quantization [7, 30] and speculative decoding [15, 27] reduce compute overhead, caching has emerged as a highly effective method to reduce redundant computations entirely. Previous works have investigated caching at multiple levels, from reusing internal KV-states [13, 21, 25] to routing API requests [6, 22]. At the query level, earlier works relied on exact-text matching [19, 32], which is restrictive for natural language prompts. This limitation led to the development of semantic caching [8, 9, 16], allowing similar queries to share responses. While recent systems extend semantic caching to richer settings such as multi-turn contexts [31] and tiered architectures [26], they remain fundamentally heuristic based and assume knowledge of query costs or arrival distributions or do not generalize across a continuous query space. Our work tackles the more complex and realistic scenario of caching over an unbounded, continuous query space with uncertain query costs and arrivals.

**Learning-Based Cache Optimization.** Traditional caching methods, such as LFU (least frequently used) and LRU (least recently used) [14], are simple online heuristics that adapt to observed request patterns without explicitly modeling system parameters. In contrast, cost-aware variants such as LEC [1] estimate quantities like request probabilities and retrieval costs to guide caching decisions. In our previous work [18], we introduced an online adaptation framework for semantic caching, but it remained restricted to a finite, discrete query set. Transitioning to a continuous domain inherently changes the mathematical structure of the problem. Specifically, our continuous caching framework can be modeled as a variant of the Combinatorial Multi-Armed Bandit (CMAB) problem [4]. Unlike standard CMAB settings, our environment introduces a "reverse feedback" challenge: when a query is served from the cache, the true cost of its execution on the LLM remains unobserved.

## 1.2 Our Contributions

Our work makes the following contributions:

**(1) Continuous Semantic Caching Model** (Section 2): We model a sequence of queries, each drawn from an a priori unknown probability distribution over the continuous query space, arriving at a LLM operator (Figure 1). Upon arrival, a query is either served from the cache or from the LLM, depending on whether the *mismatch* cost of serving the cached response of a semantically similar query exceeds the *serving* cost of LLM inference on this query. To handle infinite query space, we introduce the notion of an $\epsilon$-net that partitions the query space into Voronoi regions with semantically similar queries. Given an $\epsilon$-net, one can then discretize the cache design problem by caching at most one representative query per region. Importantly, *we do not assume a pre-defined $\epsilon$-net*, as covering the entire query space is likely intractable. We then face the challenge that solely working with these $\epsilon$-net query candidates introduces not only discretization errors in choosing the optimal cached queries but also additional errors in estimating query costs, as semantically similar queries may still have distinct serving costs. We address these in our theoretical analysis below.

**(2) Algorithm Design:** We consider three cache design settings. First, in the *oracle setting* (Section 3) with full information of query costs and arrivals, we use our $\epsilon$-net definition to provide the `Reverse Greedy` algorithm with provable approximation and discretization guarantees. We then consider the *offline setting* (Section 4), where historical serving and arrival data is available (e.g., from deployment logs [19, 32]). We use this data to build an $\epsilon$-net and introduce a pessimistic learning algorithm, `CUCB-SC-Cont`, that uses kernel ridge regression (KRR) to estimate the serving cost for a given query from those of other queries in the historical data. Finally, in the *online adaptive setting* (Section 5), we propose `CLCB-SC-LS-Cont`, which adaptively constructs an $\epsilon$-net on online query arrivals. The online setting further introduces a low-switching constraint not present in traditional CMAB: changing the cache may require rerunning some queries to get their responses to cache, incurring inference costs. We thus design adaptive thresholds that minimize such cache switching along with traditional algorithm regret.

**(3) Theoretical Guarantees:** After deriving an approximation guarantee for the oracle setting, we provide a suboptimality guarantee for `CUCB-SC-Cont` in the offline setting. To do so, we introduce a novel method to quantify the KRR estimation error for the model parameters and the resulting cache suboptimality. We then use these results to analyze the optimal choice of $\epsilon$, trading off between a lower discretization versus estimation error. Finally, we derive a regret upper bound for `CLCB-SC-LS-Cont`, despite the continuous space from which our queries are drawn.

**(4) Performance Evaluation** (Section 6): We conduct extensive experiments on synthetic and real-world datasets for offline and online settings. We show that `CUCB-SC-Cont` achieves the lowest suboptimality gap in the offline setting, yielding up to a 73.32% improvement over prior methods. In the online setting, `CLCB-SC-LS-Cont` outperforms baselines regarding final average regret, with up to 72.41% and 43.14% reductions observed on synthetic and bursty real-world query streams, respectively. In addition, our algorithm maintains these strong regret results while keeping runtime highly competitive with existing semantic caching baselines.

We defer all theoretical proofs and additional experimental results to our supplementary material. Results numbered with the prefix "A" (e.g., Lemma A.1 and Corollary A.3) refer to this material.

# 2 System Model

In this section, we present the system model of the continuous semantic caching problem for low-cost LLM serving (Fig. 1).

## 2.1 LLM Serving System

We formalize our LLM serving system as follows.

**Queries and Embeddings.** We assume queries, which each represent a natural language prompt, are independently drawn from a probability density function $f : \mathcal{X} \to \mathbb{R}_+$ over a continuous space $\mathcal{X} \subset \mathbb{R}^{d_e}$ such that $\int_{\mathcal{X}} f(x)\,dx = 1$. We assume that $\mathcal{X}$ inherits a metric structure through a distance function $d : \mathcal{X} \times \mathcal{X} \to \mathbb{R}_+$ (e.g., Euclidean distance in the embedding space). We let $p \le d_e$ denote the intrinsic (metric) dimension of $\mathcal{X}$. Specifically, we assume that there exists a constant $c_0 > 0$ such that for any $x \in \mathcal{X}$ and any sufficiently small radius $r > 0$, $\int_{B(x,r)} f(x')dx' \geq c_0 r^p$ where $B(x,r) = \{x' \in \mathcal{X} : d(x,x') \leq r\}$.

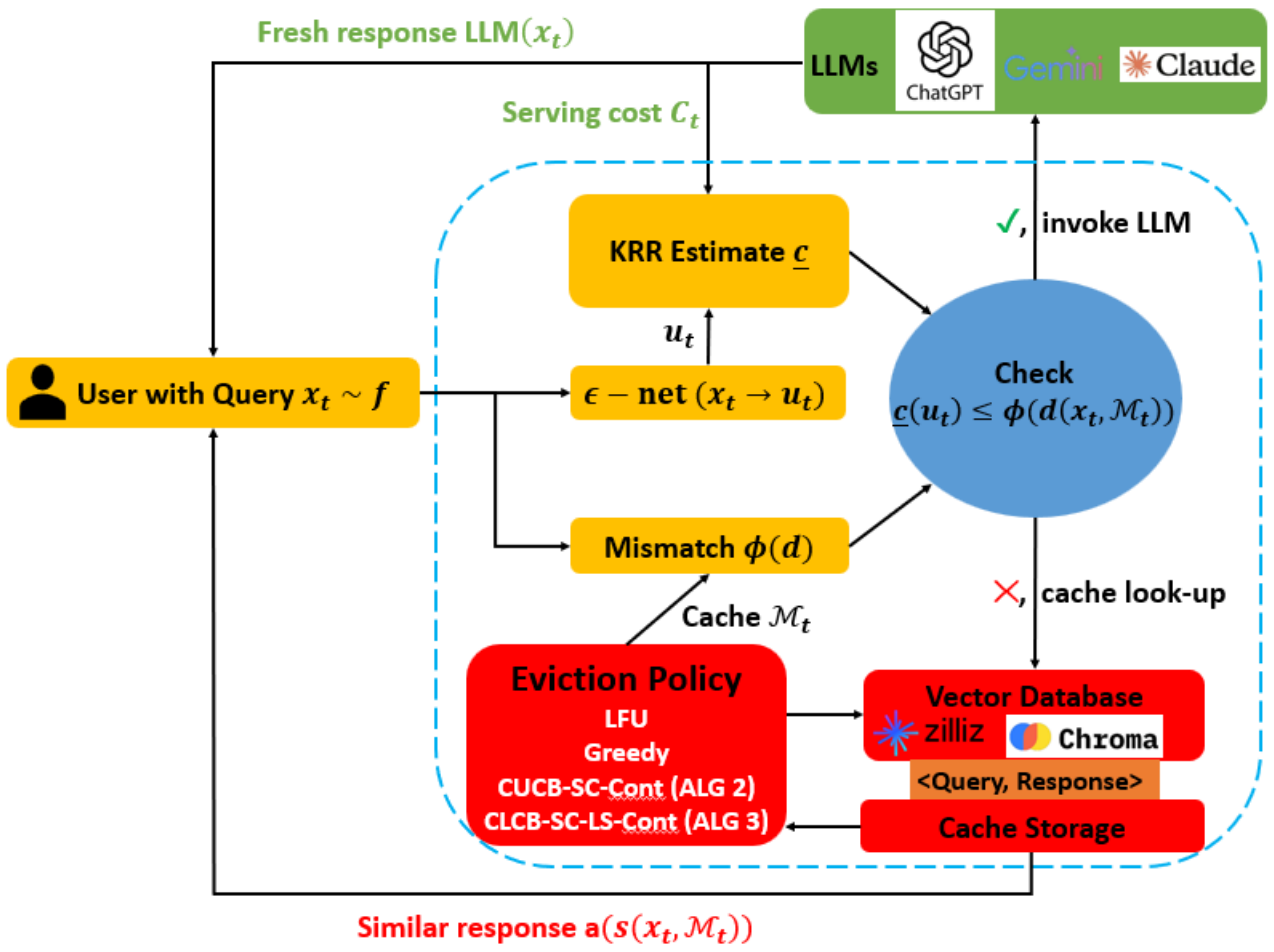

**Figure 1: Continuous semantic caching system for low-cost LLM serving, showing query arrivals, service from the cache or LLM, and cache updates.**

**Serving Costs and Partial Observations.** Each query $x \in \mathcal{X}$ is associated with an unknown expected serving cost $c(x) \in (0,1]$, representing, for example, latency or computational cost of invoking the LLM. When the system queries the LLM with input $x$, it observes a noisy realization $C_t(x) = c(x) + \eta_t(x)$, where the noise sequence $\eta_t(x)$ is conditionally $R$-sub-Gaussian given the history $\mathcal{F}_{t-1}$ (i.e., $\mathbb{E}[\eta_t(x)|\mathcal{F}_{t-1}] = 0$ and $\mathbb{E}[\exp(\xi\eta_t(x))|\mathcal{F}_{t-1}] \leq \exp(\xi^2 R^2/2)$ for all $\xi \in \mathbb{R}$). This cost is only observed when the LLM is invoked.

**Similarity and Mismatch Costs.** For a semantic cache of at most $k$ query-response pairs $\mathcal{M} = \{(u_1, a(u_1)), \dots, (u_k, a(u_k))\}$, $u_i \in \mathcal{X}$, we define the distance from a query $x$ to the cache as $d(x, \mathcal{M}) := \min_{u \in \mathcal{M}} d(x,u)$, and denote the closest cached query by $s(x, \mathcal{M}) := \operatorname{argmin}_{u \in \mathcal{M}} d(x,u)$. Note that to satisfy the Lipschitz continuity required for our theoretical guarantees, $d$ must be a well-defined metric space distance (e.g., the Euclidean distance between $L_2$-normalized embeddings). If a query $x$ is served using the cached response corresponding to $u \in \mathcal{M}$, it incurs an expected mismatch cost $\phi(d(x,u))$, where $\phi : \mathbb{R}_+ \to [0,1]$ is a non-decreasing function. Because queries arrive from a continuous space, a user prompt can be significantly different from all previously cached items, potentially yielding an arbitrarily large geometric distance. Bounding the mismatch penalty to a maximum of 1 explicitly caps the cost of these extreme outliers, ensuring the cache penalty never outscales the normalized cost of querying the LLM, while providing the strict upper bounds required for our theoretical regret guarantees. For example, $\phi$ could be selected as a clipped linear function, $\phi(d) = \min(\zeta d, 1)$ for some scaling parameter $\zeta > 0$. On the other hand, querying the LLM directly on $x$ incurs no mismatch cost.

## 2.2 Semantic Caching

The system maintains a semantic cache $\mathcal{M}$ defined as before. At each round $t$, a query $x \sim f$ arrives. The system chooses between:

- **Cache Lookup:** Serve using the closest cached query $s(x, \mathcal{M})$ and incur mismatch cost $\phi(d(x, \mathcal{M}))$;
- **LLM Query:** Query the LLM, incur and observe cost $c(x)$.

The optimal decision rule compares the two costs:

$$a_t = \begin{cases} \mathtt{LLM}(x) & \text{if } c(x) \leq \phi(d(x, \mathcal{M})), \\ a(s(x, \mathcal{M})) & \text{otherwise.} \end{cases} \tag{1}$$

This raises the central question: *how should we construct the cache $\mathcal{M}$ in a continuous query space?*

**Discretization via $\epsilon$-Net and Voronoi Regions.** Optimizing directly over the continuous space $\mathcal{X}$ is intractable. To obtain a finite representation, we approximate $\mathcal{X}$ using an $\epsilon$-net $\mathcal{S} = \{x_1, \dots, x_m\} \subseteq \mathcal{X}$ such that for every $x \in \mathcal{X}$, there exists $x_i \in \mathcal{S}$ with $d(x, x_i) \leq \epsilon$.

The set $\mathcal{S}$ induces a partition of the space into $m$ regions $\{V_i\}_{i=1}^m$, defined as Voronoi regions:

$$V_i := \{x \in \mathcal{X} : d(x, x_i) \leq d(x, x_j),\ \forall j \neq i\}.$$

Each region $V_i$ contains all queries whose closest representative in $\mathcal{S}$ is $x_i$. This induces a finite partition of the continuous query space. We define the weight of each region $V_i$ as

$$w_i := \int_{V_i} f(x)\, dx,$$

i.e., the probability mass of queries in this region.

**Loss Function over Continuous Query Space.** The expected loss of a cache $\mathcal{M}$ under density $f$ is

$$\mathcal{L}(\mathcal{M}; f, c, d) = \int_{\mathcal{X}} f(x) \min\{c(x), \phi(d(x, \mathcal{M}))\} dx. \tag{2}$$

The cache design problem aims to find:

$$\mathcal{M}^* = \operatorname{argmin}_{\mathcal{M} \subseteq \mathcal{X}, |\mathcal{M}| \leq k} \mathcal{L}(\mathcal{M}; f, c, d) \tag{3}$$

Using the $\epsilon$-net approximation, we obtain the discretized loss:

$$\widehat{\mathcal{L}}(\mathcal{M}; \mathcal{S}, \mathbf{w}, c) = \textstyle\sum_{i=1}^m w_i \min\{c(x_i), \phi(d(x_i, \mathcal{M}))\}. \tag{4}$$

In our formal theoretical approach, which we detail in the next sections, we bound the approximation error between Equation (2) and Equation (4), which enables us to derive guarantees on the discretized loss and relate it to the expected loss under density $f$.

# 3 Approximate Solution for Continuous Semantic Caching with Known Parameters

In this section, we study the semantic caching problem under the assumption that the arrival distribution and cost function are **known**. Our formulation builds on the discrete semantic caching framework of [18], which considers a finite query set and derives a discrete reverse-greedy approximation algorithm. However, since our queries are drawn from a continuous space $\mathcal{X} \subset \mathbb{R}^{d_e}$, we replace [18]'s finite sum objective with Equation 2's integral loss function. We first show that the cache design problem for the optimal solution over this continuous space is computationally intractable:

Lemma 3.1 (NP-hardness). *The continuous semantic caching problem shown in Equation 3 is NP-hard.*

This lemma shows that our formulation is a strict generalization of the discrete setting. In particular, if $\mathcal{X}$ is finite and $f$ is discrete, we recover the setting in [18] exactly. Our problem inherits the NP-hardness but introduces additional complexity due to the infinite

**Algorithm 1** `Continuous Reverse Greedy` [18]: Removing Least Beneficial Cache Items

```
Input: candidates S, weights w, costs c, d, k, φ
Initialize: M_0 ← S, m = |S|. If m ≤ k, Return M = S.
for i = 1 to m − k do
    q_i = argmin_{q∈M_{i−1}} L̂(M_{i−1} \ {q}; S, w, c)
    Update M_i ← M_{i−1} \ {q_i}
end for
Return: M = M_{m−k}
```

**Algorithm 2** `CUCB-SC-Cont`: Combinatorial Upper Confidence Bound Algorithm for the Semantic Caching Problem

1: **Input:** $\mathcal{D} = \{(\mathcal{M}_t, x_t, C_t)\}_{t=1}^{n}$ , $d$, kernel $\kappa$, confidence $\delta$, $\phi$, $k$, $\epsilon$, ridge $\lambda$, split into $\mathcal{D}_1$ ($n_1$ rounds) and $\mathcal{D}_2$ ($n_2$ rounds), $B$, $R$.
2: **Build** $\epsilon$**-net** $\mathcal{S} \subset \{x_t\}_{t\in\mathcal{D}_1}$, covering every observed $x_t$
3: **Weights**: Cells $\{V_i\}$, set $\hat{w}_i = \frac{1}{n_2}\sum_{t\in\mathcal{D}_2} \mathbf{1}\{x_t \in V_i\}\ \forall i \in [m]$
4: **KRR fit of** $c$: Gather all cost observations $(x_t, C_t)$ where LLM was queried, form $K, y$, and compute $\hat{c}(x) = \kappa_x^T(K+\lambda I)^{-1}y$ and $\sigma_\lambda^2 = \kappa(x,x) - \kappa_x^T(K+\lambda I)^{-1}\kappa_x$
5: **Set pessimistic costs**: $\forall x \in \mathcal{S}$, confidence radius $\Delta(x) = \left(B + R\sqrt{\frac{2\log(6m/\delta)}{\lambda}}\right)\sigma_\lambda(x)$ and set $\bar{c}(x) = \min\{1, \hat{c}(x) + \Delta(x)\}$
6: Compute $\hat{\mathcal{A}}$ = `Cont_Reverse_Greedy` $(\mathcal{S}, \hat{\boldsymbol{w}}, \bar{\boldsymbol{c}}, d, k, \phi)$ and query `LLM` for $u \in \hat{\mathcal{A}}$
7: **Return:** $\hat{\mathcal{M}} = \{u, a(u)\}_{u\in\hat{\mathcal{A}}}$

decision space induced by continuous query space. To address this challenge, we use Section 2.2's continuous-to-discrete reduction via an $\epsilon$−net. Next, we bound the resulting discretization error.

*Assumption* 1 (Lipschitz Continuity of the Expected Cost). Define $g_{\mathcal{M}}(x) = \min\{c(x), \phi(d(x,\mathcal{M}))\} \in [0,1]$. Assume there is a constant $L_g > 0$ such that for all cache $\mathcal{M}$, $x \to g_{\mathcal{M}}(x)$ is $L_g$-Lipschitz:

$$|g_{\mathcal{M}}(x) - g_{\mathcal{M}}(y)| \le L_g d(x,y) \quad \forall x, y \in \mathcal{X}$$

Note that a sufficient condition for Assumption 1 to hold is that both $c(x)$ and $\phi(\cdot)$ are themselves Lipschitz continuous.

Lemma 3.2 (Continuous-to-discrete approximation via $\varepsilon$-net). *Let $\mathcal{S}$ be a static $\epsilon$-net covering the space $\mathcal{X}$ and suppose that Assumption 1 holds. Then for every $\mathcal{M}$ with $|\mathcal{M}| \le k$:*

$$|\mathcal{L}(\mathcal{M}) - \widehat{\mathcal{L}}(\mathcal{M};\mathcal{S})| \le L_g\epsilon$$

This lemma shows that the continuous objective can be well approximated by its discrete counterpart. The approximation error scales linearly with the covering radius $\epsilon$ and with this bound, the problem reduces to a finite optimization problem over $\mathcal{S}$.

We directly apply Lemma 2 from [18] to design Algorithm 1. This lemma shows that our discretized loss function is supermodular and non-increasing regarding the set $\mathcal{M}$. Theorem 3.3 provides a guarantee on the performance of Alg. 1 following [10] and combining this result with Lemma 3.2.

Theorem 3.3 (Approximation Guarantee). *Let $\mathcal{M}$ be the solution returned by Alg. 1. Then $\widehat{\mathcal{L}}(\mathcal{M};\mathcal{S},\boldsymbol{w},\boldsymbol{c}) \le \alpha\,\widehat{\mathcal{L}}(\mathcal{M}^*_{\mathcal{S}};\mathcal{S},\boldsymbol{w},\boldsymbol{c})$, where $\mathcal{M}^*_{\mathcal{S}} \in \arg\min_{\mathcal{M}'\subseteq\mathcal{S},|\mathcal{M}'|\le k} \widehat{\mathcal{L}}(\mathcal{M}';\mathcal{S},\boldsymbol{w},\boldsymbol{c})$, $\alpha = \frac{e^{\Gamma}-1}{\Gamma}$, $\Gamma = \frac{s}{1-s}$ and $s \in [0,1]$ is the curvature. Moreover, for the continuous optimum $\mathcal{M}^*$ in Eq.* (3),

$$\mathcal{L}(\mathcal{M}) - \alpha\mathcal{L}(\mathcal{M}^*) \le \widehat{\mathcal{L}}(\mathcal{M};\mathcal{S},\boldsymbol{w},\boldsymbol{c}) - \alpha\widehat{\mathcal{L}}(\mathcal{M}^*_{\mathcal{S}};\mathcal{S},\boldsymbol{w},\boldsymbol{c}) + (1+2\alpha)L_g\epsilon.$$

# 4 Offline Learning for Continuous Semantic Caching Problem with Unknown Parameters

In this section, we consider the semantic caching problem when the cost function $c(x)$ and density $f$ are unknown and must be learned from an offline dataset. Unlike the setting in Section 3, where $f$ and $c$ are assumed known, we are given a dataset of observations $\{(\mathcal{M}_t, x_t, C_t)\}_{t=1}^{n}$, where $C_t$ denotes the cost incurred when serving query $x_t$ and $\mathcal{M}_t$ is the selected cache at round $t$. If $C_t = \emptyset$, it indicates that the cached response from the nearest neighbor in $\mathcal{M}_t$ was used and no cost was observed. If $C_t > 0$, the LLM was invoked and the cost feedback was recorded.

Our formulation extends the discrete semantic caching framework of [18], which assumes a finite query set and estimates costs using empirical means. The key difference from [18] is that their framework does not require generalization across inputs. In contrast, our setting requires learning the cost function over a continuous space and making predictions for previously unseen queries. This introduces two key challenges: (i) estimating $c(x)$ over a continuous domain, and (ii) incorporating estimation uncertainty into the cache optimization problem.

To address these challenges, we model $c(x)$ using kernel ridge regression (KRR) in a reproducing kernel Hilbert space (RKHS). Building on established kernelized bandit frameworks [5], KRR provides both the non-parametric flexibility to learn complex serving cost functions and the closed form posterior variance necessary to construct confidence bounds that enable optimal cache selection.

Since computing the optimal cache is NP-hard even when the parameters are known (Lemma 3.1), we instead aim to learn an approximate solution that competes with the $\alpha$-approximate oracle returned by the Continuous Reverse Greedy algorithm (Alg. 1). This algorithm is consistent with prior work on discrete semantic caching [18], where the same approximation notion is adopted due to computational intractability. Accordingly, our goal is to minimize the approximate suboptimality defined with respect to this oracle:

$$\mathrm{SubOpt}(\mathcal{M}) = \mathcal{L}(\mathcal{M}) - \alpha \cdot \mathcal{L}(\mathcal{M}^*), \tag{5}$$

Before proceeding with the `CUCB-SC-Cont` algorithm (shown in Algorithm 2), we provide the following definitions. For a query $x$, we define the kernel vector $\kappa_x = [\kappa(x_{t_1}, x), \ldots, \kappa(x_{t_\ell}, x)]^\top$, where $\{(x_{t_i}, C_{t_i})\}_{i=1}^{\ell}$ are the subset of queries for which the LLM was invoked and the cost was observed and $\{t_1, ..., t_\ell\} \subseteq \{1, ..., n\}$ are the indices where the LLM was actually queried. Note that $\kappa(x,x)$ is a scalar measuring self-similarity, whereas $\kappa_x$ is a vector of similarities between $x$ and the observed data. We define the kernel (Gram) matrix $K \in \mathbb{R}^{\ell\times\ell}$ as $K_{ij} = \kappa(x_{t_i}, x_{t_j})$, $\forall i, j \in [\ell]$ and the vector of observed costs as $y = [C_{t_1}, ..., C_{t_\ell}]^\top$.

Algorithm 2 presents our offline learning procedure for the continuous semantic caching problem with unknown parameters. The algorithm splits the rounds into $\mathcal{D}_1$ and $\mathcal{D}_2$, constructs an $\epsilon$-net $\mathcal{S}$ over the embeddings in $\mathcal{D}_1$ (Line 2), and estimates each arrival weight $\hat{w}_i$ as the fraction of $\mathcal{D}_2$ queries in cell $V_i$ (Line 3). Building the partition and the estimates on disjoint data keeps the partition independent of the samples it is evaluated on. To learn the unknown serving cost function $c(x)$, we apply KRR using only the subset of queries for which the LLM was actually invoked (non-empty cost values), yielding an estimate $\hat{c}(x)$ along with an uncertainty term $\sigma_\lambda(x)$ (Line 4). Based on this, we construct a confidence radius $\Delta(x)$ and define pessimistic cost estimates $\bar{c}(x)$, which upper bound the true cost with high probability (Line 5). Finally, we apply the reverse greedy oracle on $(\mathcal{M}, \mathcal{S}, \hat{\boldsymbol{w}}, \bar{\boldsymbol{c}})$ to compute an approximate optimal cache (Lines 6-7). We directly estimate $\boldsymbol{w}$ using empirical frequencies in the $\epsilon-$net because every incoming query is observed. In contrast, cost feedback is sparse and recorded only when the LLM is invoked, which necessitates using KRR to generalize across the continuous space. We now present its performance guarantee:

Theorem 4.1 (Suboptimality Gap Bound for Algorithm 2). *Suppose Assumption 1 holds, $\epsilon$-net $\mathcal{S}$ built on $\mathcal{D}_1$ covers $\mathcal{X}$, RBF kernel is used, the conditions of Lemma 4.2 hold, and $n_2 \geq \max_i N_0/(w_i v_i \underline{v}_i)$ ($N_0$ as in Corollary A.3) with $\delta/3$ in place of $\delta$. Then with probability at least $1-\delta$, the cache $\hat{\mathcal{M}}$ returned by* `CUCB-SC-Cont` *satisfies*

$$\mathrm{SubOpt}(\hat{\mathcal{M}}) \leq \underbrace{(\alpha+1)\sqrt{\frac{2m\ln(6/\delta)}{n_2}}}_{Arrival} + \underbrace{(1+2\alpha)L_g\frac{2\sqrt{d_e}}{m^{1/d_e}}}_{Discretization} + \underbrace{2\sqrt{2}\,\alpha(B+c_\lambda)\left(\frac{2\lambda(2\epsilon/\ell_k)^p\,C_v}{n_2}\right)^{\frac{1}{p+2}}}_{Cost},$$

*where $c_\lambda = R\sqrt{2\ln(6m/\delta)/\lambda}$ and $C_v = \sum_i (v_i\underline{v}_i)^{-1}$ and $v_i$ is the local observation propensity of cell $V_i$ and $\underline{v}_i$ the lower-mass constant of $V_i$.*

While the covering number $m$ scales exponentially with the embedding dimension $d_e$, we explicitly manage this dependency in Remark 1 in our supplementary material by deriving an optimal discretization radius $\epsilon^*$ that balances $m$ against the sample size $n$ to ensure the suboptimality gap converges to 0.

Our proposed algorithm and analysis can be viewed as a continuous analogue of the offline CUCB-SC algorithm in [18], but introduce several key differences that also drive our technical contributions. Unlike the discrete setting, we do not assume a known finite query universe and instead approximate the arrival distribution via an $\epsilon$-net over the embedding space, incurring a discretization error absent in the discrete case. Moreover, rather than estimating costs independently per query, we learn a continuous cost function $c(x)$ using KRR, leading to confidence bounds for each candidate query that depend on the number of cost observations near each candidate rather than the total number of queries $n$. These differences require jointly addressing geometric approximation and nonparametric function estimation, yielding three main contributions: (i) quantifying the discretization error induced by the $\epsilon$-net, (ii) replacing per-query empirical estimates with KRR-based cost learning and uncertainty quantification, and (iii) deriving geometry-aware bounds on the regularized KRR posterior width by linking local counts of cost observations near each representative to the width through a bias–variance trade-off under the RBF kernel.

The bound in Theorem 4.1 decomposes into three terms corresponding to distinct sources of error in the continuous semantic caching setting. **(i) Arrival estimation error:** The term $O\left(\sqrt{\frac{m}{n_2}}\right)$ arises from estimating the arrival weights $\boldsymbol{w}$ from $n_2$ samples, and matches the concentration rate in the discrete setting [18]. **(ii) Cost estimation error:** Reflects the difficulty of learning a continuous cost function $c(x)$ from partial observations. Unlike the discrete case, where the estimation error scales as $O(n^{-1/2})$, the continuous setting incurs a slower rate $O(n_2^{-1/(p+2)})$ due to the nonparametric nature of the problem and the need to generalize across nearby queries. This term captures the interaction between kernel smoothness (through $\ell_k$), intrinsic dimension $p$, and the local observation propensities $v_i$. **(iii) Discretization error:** The term $O\left(m^{-1/d_e}\right)$ arises from approximating the continuous space $\mathcal{X}$ using an $\epsilon$-net of size $m$, and vanishes as the discretization becomes finer.

**Comparison to the Discrete Case and Reduction.** In the discrete setting [19], the suboptimality gap scales as $O\left(\sqrt{\frac{\sum_q v(q)^{-1}}{n}}\right)$, reflecting finite-arm estimation with no geometric structure. In contrast, our bound generalizes this result to continuous spaces by introducing both geometric approximation and functional estimation terms. Specifically, while the dependence on $v_i$ is preserved, it is attenuated through the dimension-dependent exponent $1/(p+2)$, and the rate in $n$ degrades from $n^{-1/2}$ to $n_2^{-1/(p+2)}$ due to the nonparametric nature of learning the continuous cost function. Additionally, new dependencies on the kernel parameter $\ell_k$ and discretization level $m$ arise. At the same time, our result recovers the discrete case as a special instance. When the query space is finite and coincides with the $\epsilon$-net (i.e., $\epsilon = 0$ and $m = |\mathcal{Q}|$), and when the kernel does not induce smoothing across distinct queries, the geometric (kernel induced) and discretization terms vanish.

In this regime, taking $r = 0$ in Lemma A.2 gives $\sigma_\lambda^2(x_i) \leq \lambda/N_i$, where $N_i$ is the number of cost observations of $x_i$, so cost estimation reduces to per-query estimation and the bound recovers the $O(n^{-1/2})$ scaling of the discrete CUCB-SC analysis. To establish the result in Theorem 4.1, we use the following lemma with the concentration bound guarantees.

Lemma 4.2 (Concentration of Estimates). *Let $\mathcal{S}$ be built on $\mathcal{D}_1$. Assume $\kappa$ is the unit-variance RBF kernel with length $\ell_k$, $c \in \mathcal{H}_\kappa$ with $\|c\| \leq B$, R-sub-Gaussian noise, a fixed $\lambda$, logging whose decisions do not depend on observed costs, and conditions (P) and (M) of Lemma A.1, $c_\lambda := R\sqrt{2\ln(2m/\delta)/\lambda}$, $\rho_i := 2\lambda(2\epsilon/\ell_k)^p/(n_2 w_i v_i \underline{v}_i)$.*

$$\mathcal{E}_{\mathrm{arvl}} := \{\|\hat{w} - w\|_1 \leq \sqrt{2m\ln(2/\delta)/n_2}\},$$
$$\mathcal{E}_{\mathrm{cost}} := \{0 \leq \bar{c}(x_i) - c(x_i) \leq 2(B+c_\lambda)\,\sigma_\lambda(x_i)\ \forall i\},$$
$$\mathcal{E}_{\mathrm{width}} := \{\sigma_\lambda(x_i) \leq \sqrt{2}\,\rho_i^{1/(p+2)}\ \forall i\}.$$

*If $n_2 \geq \max_i N_0/(w_i v_i \underline{v}_i)$, ($N_0$ defined in Corollary A.3) all three events hold with probability at least $1-3\delta$. On their intersection, for all $i$,*

$$0 \leq \bar{c}(x_i) - c(x_i) \leq \min\{1,\ 2\sqrt{2}\,(B+c_\lambda)\,\rho_i^{1/(p+2)}\}.$$

Lemma 4.2 characterizes a high-probability event under which both arrival weights and costs are accurately estimated. While $\mathcal{E}_{\mathrm{arvl}}$ follows from standard concentration arguments [29], $\mathcal{E}_{\mathrm{cost}}$ and $\mathcal{E}_{\mathrm{width}}$ capture the main technical challenge in the continuous setting. $\mathcal{E}_{\mathrm{cost}}$ is a KRR confidence bound with a fixed ridge, and $\mathcal{E}_{\text{width}}$ controls the regularized posterior width: Lemma A.1 lower-bounds the number of cost observations near each representative, and Lemma A.2 converts these local counts into a width bound through a bias–variance trade-off in the neighborhood radius. This yields an explicit cost estimation error that depends on the intrinsic dimension, kernel bandwidth and local observation propensities.

**Generality of the Kernel Choice.** We derive our geometric bounds (Lemma A.2) using the RBF kernel because it is one of the standard choices for continuous embedding spaces. However, our mathematical framework easily extends to other kernels. The idea of bounding the kernel's norm applies to any distance based kernel. For instance for the Laplacian kernel we would get $n_2^{-1/(2(p+1))}$ instead.

**Optimal $\epsilon$ Value.** We derive the optimal covering radius $\epsilon^*$ in Remark 1, defined as the specific radius that minimizes the algorithm's proved suboptimality gap in Theorem 4.1. We use the fact that $m = O(\epsilon^{-d_e})$ in this derivation. Our analysis shows that this optimal radius shifts based on the total number of samples $n$ and embedding dimension $d_e$. As we collect more data ($n$ increases), we can shrink $\epsilon^*$ to create a finer discretization. However, the geometry of the space also introduces some dependencies: if $d_e$ is large, this requires us to use a corresponding large radius to cover the space. We find the optimal radius to be $\epsilon^* = O(n_2^{-1/(d_e+2)})$

# 5 Online Adaptive Learning for the Continuous Semantic Caching Problem

Having developed an offline learning algorithm for the continuous semantic caching problem, we now turn to the online setting, where the system must learn the optimal cache dynamically from a stream of queries without access to a pre-collected dataset. Queries arrive sequentially, drawn from a continuous embedding space, and the algorithm must simultaneously estimate the arrival distribution and the cost function while making caching decisions in real time. We thus face the challenge of partial feedback: the serving cost is observed only when the LLM is queried. The goal is to adaptively update the cache to minimize long-term serving and mismatch cost while balancing exploration (querying the LLM to improve estimates) and exploitation (serving from the cache using current knowledge). To achieve this, we develop two distinct algorithms. Our primary method, which we introduce in this section, dynamically expands a representative set of queries on the fly to continuously adapt to new semantic regions and uses adaptive stages. We also introduce an alternative algorithm in the supplementary material, `CLCB-Frozen-Cont`, which restricts learning to a fixed frozen candidate pool over stable, discrete stages to provide stricter control over cache updates. Next, we introduce the online learning setup, present the proposed algorithm `CLCB-SC-LS-Cont` in Algorithm 3, and finally provide theoretical guarantees and analysis.

## 5.1 Online Learning Setup

We consider a $T$-round sequential decision-making problem over a continuous embedding space $\mathcal{X} \subset \mathbb{R}^{d_e}$. At the beginning of each round $t \in [T]$, the system maintains a semantic cache $\mathcal{M}_t$ of size $k$ and a user with query $x_t \in \mathcal{X}$ arrives. As described in Section 2.2, the agent either serves the cached response of the nearest neighbor or queries the LLM for a fresh response and observes the cost.

At the end of round $t$, the agent decides whether to update the cache. If $\mathcal{M}_t \neq \mathcal{M}_{t-1}$, the agent incurs a *switching cost* for fetching and storing the updated queries in $\mathcal{M}_t \setminus \mathcal{M}_{t-1}$.

**Learning Objective.** The objective is to minimize the cumulative regret relative to the best fixed $\alpha$-approximate cache over the time horizon, while accounting for both serving and switching costs. We define our comparator as $\mathcal{M}^*_{\mathcal{N}_T}$, the optimal cache of size $k$ selected from the set of all observed queries $\mathcal{N}_T$. This definition is necessary because, in a practical serving system, we can only populate the cache with query-response pairs that we have actually observed. Formally, letting $\mathcal{M}_0 = \emptyset$, the expected cumulative regret with switching costs is defined as:

$$\text{Reg}(T) = \mathbb{E}\left[\sum_{t=1}^{T}[\mathcal{L}(\mathcal{M}_t) - \alpha\mathcal{L}(\mathcal{M}^*_{\mathcal{N}_T}) + \sum_{x\in\mathcal{M}_t\setminus\mathcal{M}_{t-1}} c(x)]\right]$$

where the expectation is taken over the randomness of the continuous query arrivals, the serving costs, and the cache selection.

## 5.2 Algorithm Design of `CLCB-SC-LS-Cont`

In this section, we present `CLCB-SC-LS-Cont`, an online learning algorithm designed for semantic caching over continuous spaces with low switching costs. The pseudocode is provided in Algorithm 3, with subroutines `Switch Cache Continuous` in Algorithm 4 and `Serve and Update Continuous` in Algorithm 5. `CLCB-SC-LS-Cont` proceeds in four main phases: (i) dynamically constructing a representative discrete state space from prior arrivals, (ii) determining when to switch the cache by defining observation thresholds, (iii) actively exploring unknown serving costs using KRR, and (iv) serving queries while refining the kernel dataset.

**Dynamic Representative Set Construction (Algorithm 3, Lines 5-11).** As each query $x_t \sim f(x)$ arrives, the algorithm builds an empirical representative set $\mathcal{S}_t$ on the fly. If $x_t$ is further than the covering radius $\epsilon$ from all existing centers in $\mathcal{S}_{t-1}$, it is added as a new center ($\mathcal{S}_t = \mathcal{S}_{t-1} \cup \{x_t\}$), its counters are initialized, and a cache switch is forced to account for this new region of the space. Otherwise, $x_t$ is mapped to its nearest existing center $u_t$, and the representative set remains unchanged.

**Stage-Based Cache Switching (Algorithm 3, Lines 17-23).** To control switching frequency while ensuring sufficient exploration over the expanding set $\mathcal{S}_t$, we utilize a stage-based switching mechanism. For each center $u \in \mathcal{S}_t$, we maintain a local stage index $\tau_u$. A new stage is triggered and the `Switch` flag is set to true, when the number of queries mapped to $u$ in the current stage exceeds $1 + \sqrt{\frac{T}{m_t}\sum_{\tau<\tau_u}|\mathcal{T}(u,\tau)|}$. Crucially, this threshold dynamically adjusts based on the current number of active centers $m_t = |\mathcal{S}_t|$. A similar global stage threshold $\tau_f$ is maintained to track the arrivals in the current stage.

**Active Exploration via KRR Optimism (Algorithm 4, Lines 3-18).** When a switch is triggered, the agent must estimate the unknown serving costs to rebuild the cache. To efficiently explore the continuous cost space, we apply the principle of optimism in the face of uncertainty using KRR. For each center $u \in \mathcal{S}_t$, the algorithm evaluates the Gram matrix $K_{t-1}$ over the exact historically observed query-cost pairs $O_{t-1}$. It computes the KRR posterior

**Algorithm 3** `CLCB-SC-LS-Cont`: Continuous Semantic Caching with Low Switching

1: **Input:** Kernel $\kappa$, ridge $\lambda$, confidence $\delta$, $k$, $\phi$, distance $d$, radius $\epsilon$
2: **Initialize and define:** $\mathcal{S}_0, \hat{\mathcal{M}}_0, O_0 \leftarrow \emptyset$, $\ell_0 \leftarrow 0$, global stage $\tau_f \leftarrow 1$, $\mathcal{T}(f, \tau_f) \leftarrow \emptyset$, $\pi_{\mathcal{S}}(x) = \text{argmin}_{u \in \mathcal{S}} d(x, u)$
3: **for** $t = 1, \ldots, T$ **do**
4: Receive query $x_t \in \mathcal{X}$. `Switch` $\leftarrow$ `False`
5: **if** $\mathcal{S}_{t-1} = \emptyset$ or $\min_{u \in \mathcal{S}_{t-1}} d(x_t, u) > \epsilon$ **then**
6: $\mathcal{S}_t \leftarrow \mathcal{S}_{t-1} \cup \{x_t\}$, set $u_t \leftarrow x_t$
7: Init counters: $N_{f,t-1}(u_t), N_{c,t-1}(u_t), L_{c,t-1}(u_t) \leftarrow 0$
8: Init local stage: $\tau_{u_t} \leftarrow 1$, $\mathcal{T}(u_t, \tau_{u_t}) \leftarrow \emptyset$. `Switch` $\leftarrow$ `True`
9: **else**
10: $\mathcal{S}_t \leftarrow \mathcal{S}_{t-1}$, set $u_t \leftarrow \arg\min_{u \in \mathcal{S}_t} d(x_t, u)$
11: **end if**
12: **if** $\mathcal{S}_t \neq \mathcal{S}_{t-1}$ **then**
13: $\forall u \in \mathcal{S}_t : N_{f,t-1}(u; \mathcal{S}_t) \leftarrow \sum_{s=1}^{t-1} \mathbf{1}\{\pi_{\mathcal{S}_t}(x_s) = u\}$
14: **else**
15: $\forall u \in \mathcal{S}_t : N_{f,t-1}(u; \mathcal{S}_t) \leftarrow N_{f,t-1}(u; \mathcal{S}_{t-1})$
16: **end if**
17: Let $m_t \leftarrow |\mathcal{S}_t|$. Check local stage limits for all $u \in \mathcal{S}_t$:
18: **if** $\exists u \in \mathcal{S}_t$ s.t. $|\mathcal{T}(u, \tau_u)| \geq 1 + \sqrt{\frac{T}{m_t} \sum_{\tau < \tau_u} |\mathcal{T}(u, \tau)|}$ **then**
19: $\tau_u \leftarrow \tau_u + 1$, $\mathcal{T}(u, \tau_u) \leftarrow \emptyset$, `Switch` $\leftarrow$ `True`
20: **end if**
21: **if** $|\mathcal{T}(f, \tau_f)| \geq 1 + \sqrt{T \sum_{\tau < \tau_f} |\mathcal{T}(f, \tau)|}$ **then**
22: $\tau_f \leftarrow \tau_f + 1$, $\mathcal{T}(f, \tau_f) \leftarrow \emptyset$, `Switch` $\leftarrow$ `True`
23: **end if**
24: **if** `Switch` **then**
25: $\hat{\mathcal{M}}_t \leftarrow$ `Switch_Cache_Continuous`
26: **else**
27: $\hat{\mathcal{M}}_t \leftarrow \hat{\mathcal{M}}_{t-1}$
28: **end if**
29: Call `Serve_and_Update_Continuous`
30: **end for**
31: **Return:** $\{\hat{\mathcal{M}}_t\}_{t=1}^{T}$

**Algorithm 4** `Switch Cache Continuous`

1: **Input:** $\mathcal{S}_t$, $\{N_{f,t-1}(u; \mathcal{S}_t) : u \in \mathcal{S}_t\}$, $O_{t-1}$, $\ell_{t-1}$, $\kappa$, $\lambda$, $\delta$, $k$
2: **Weights:** $\forall u \in \mathcal{S}_t$, set $\hat{w}_{t-1}(u) = \frac{N_{f,t-1}(u; \mathcal{S}_t)}{\max\{1, t-1\}}$.
3: **if** $\ell_{t-1} \geq 1$ **then**
4: Let $O_{t-1} = \{(z_j, y_j)\}_{j=1}^{\ell_{t-1}}$. Form $y_{t-1} = [y_1, \ldots, y_{\ell_{t-1}}]^\top$
5: Form Gram matrix $K_{t-1} = [\kappa(z_i, z_j)]_{i,j=1}^{\ell_{t-1}}$
6: **for** each $u \in \mathcal{S}_t$ **do**
7: Define $k_{u,t-1} = [\kappa(z_1, u), \ldots, \kappa(z_{\ell_{t-1}}, u)]^\top$
8: Compute KRR mean and variance:
9: $\hat{c}_{t-1}(u) = k_{u,t-1}^\top (K_{t-1} + \lambda I)^{-1} y_{t-1}$
10: $\sigma_{t-1,\lambda}^2(u) = \kappa(u, u) - k_{u,t-1}^\top (K_{t-1} + \lambda I)^{-1} k_{u,t-1}$
11: Set optimistic LCB: $\underline{c}_{t-1}(u) = \hat{c}_{t-1}(u) - \left(B + R\sqrt{2(\gamma_t + \ln(2/\delta))}\right) \sigma_{t-1,\lambda}(u)$
12: **end for**
13: **else**
14: **Prior:** $\forall u \in \mathcal{S}_t$, set $\hat{c}_{t-1}(u) \leftarrow 0$, $\sigma_{t-1,\lambda}(u) \leftarrow \sqrt{\kappa(u, u)}$
15: Set $\underline{c}_{t-1}(u) \leftarrow -B \cdot \sigma_{t-1,\lambda}(u)$
16: **end if**
17: $\hat{\mathcal{A}}_t \leftarrow$ `Cont_Reverse_Greedy`$(\mathcal{S}_t, \hat{\mathbf{w}}_{t-1}, \underline{\mathbf{c}}_{t-1}, k)$
18: Query `LLM` for $u \in \hat{\mathcal{A}}_t \setminus \hat{\mathcal{M}}_{t-1}$ and store responses
19: **Return:** $\hat{\mathcal{M}}_t = \hat{\mathcal{A}}_t$

**Algorithm 5** `Serve and Update Continuous`

1: **Input:** Query $x_t$, center $u_t$, cache $\hat{\mathcal{M}}_t$, $\mathcal{S}_t$, $O_{t-1}$, $\ell_{t-1}$, counters
2: Update global arrival: $\forall u \in \mathcal{S}_t : N_{f,t}(u; \mathcal{S}_t) \leftarrow N_{f,t-1}(u; \mathcal{S}_t) + \mathbf{1}\{u = u_t\}$, add $t$ to $\mathcal{T}(f, \tau_f)$
3: Compute current $\underline{c}_{t-1}(u_t)$ via KRR on $O_{t-1}$ (as in Alg. 4)
4: Let $d(x_t, \hat{\mathcal{M}}_t) = \min_{u \in \hat{\mathcal{M}}_t} d(x_t, u)$ (where $d(x_t, \emptyset) = \infty$).
5: **if** $\underline{c}_{t-1}(u_t) \leq \phi(d(x_t, \hat{\mathcal{M}}_t))$ **then**
6: Query `LLM` on $x_t$ to get $a_t$, observe true cost $C_t(x_t)$
7: $O_t \leftarrow O_{t-1} \cup \{(x_t, C_t(x_t))\}$, $\ell_t \leftarrow \ell_{t-1} + 1$
8: $N_{c,t}(u_t) \leftarrow N_{c,t-1}(u_t) + 1$, $L_{c,t}(u_t) \leftarrow L_{c,t-1}(u_t) + C_t(x_t)$
9: Add $t$ to local stage $\mathcal{T}(u_t, \tau_{u_t})$
10: **else**
11: Serve cached response: $a_t \leftarrow a(s(x_t, \hat{\mathcal{M}}_t))$
12: Carry over unmodified: $O_t \leftarrow O_{t-1}$, $\ell_t \leftarrow \ell_{t-1}$
13: Carry over unmodified: $N_{c,t}(u) \leftarrow N_{c,t-1}(u), L_{c,t}(u) \leftarrow L_{c,t-1}(u)$
14: **end if**
15: **Return:** $(a_t, O_t, \ell_t, \{N_{f,t}\}, \{N_{c,t}\}, \{L_{c,t}\}, \{\mathcal{T}(u, \tau_u)\}, \mathcal{T}(f, \tau_f))$

mean $\hat{c}_{t-1}$ and subtracts a confidence radius $\Delta_{t-1}$ which scales with the posterior standard deviation $\sigma_{t-1,\lambda}$ to obtain the optimistic lower confidence bound $\underline{c}_{t-1}$ (Lines 8-11). The algorithm then calls the `Cont_Reverse_Greedy` oracle using these optimistic costs and empirical weights $\hat{w}_{t-1}$ to construct the new cache $\hat{\mathcal{M}}_t$ and queries the LLM to store the responses not existent in the previous cache $\hat{\mathcal{M}}_{t-1}$ (Lines 17-18).

**Serving and Updating (Algorithm 5, Lines 5-14).** After the cache is determined, the agent routes the incoming query $x_t$ (mapped to center $u_t$). It compares the optimistic cost of querying the LLM $\underline{c}_{t-1}(u_t)$, against the mismatch cost $\phi(d(x_t, \hat{\mathcal{M}}))$ of serving the query from the cache (Line 5). If $\underline{c}_{t-1}(u_t) \leq \phi(d(x_t, \hat{\mathcal{M}}_t))$, the agent queries the LLM and observes the exact realized cost $C_t(x_t)$ and appends the tuple $(x_t, C_t(x_t))$ to the observation set $O_t$ (Lines 6-9). This LCB based routing ensures the algorithm actively collects cost feedback in uncertain regions of the continuous space. Otherwise, it serves the user from the cache without incurring LLM costs or updating the KRR dataset (Lines 11-13).

**Novelty Compared to Discrete Case.** While `CLCB-SC-LS-Cont` shares the high-level exploration-exploitation and low-switching logic of its discrete counterpart [18], it introduces fundamental algorithmic changes to handle continuous domains. First, rather than operating over a fixed finite set of $m$ queries, it dynamically constructs an expanding $\epsilon$−net ($\mathcal{S}_t$), smoothly adapting its stage-switching thresholds to the growing dimension $m_t$. Second, while the discrete algorithm relies on independent empirical means to estimate costs, our algorithm leverages KRR to share information across the space. By utilizing a kernel $\kappa(\cdot, \cdot)$, the observation of a cost at $x_t$ reduces the uncertainty $\sigma_{t,\lambda}(u)$ for all nearby centers $u$, thus generalizing the feedback to unobserved regions. Finally, the optimistic confidence intervals transition from standard Hoeffding/Chernoff bounds based strictly on local hit counts to posterior variance, similarly to Theorem 4.1 in the offline case.

### 5.3 Theoretical Results for CLCB-SC-LS-Cont

In this section, we present our assumption and the main theoretical result characterizing the performance of CLCB-SC-LS-Cont.

*Assumption* 2 (Cluster-Based Arrival Distribution). We assume the continuous query arrival distribution $f(x)$ is highly concentrated, such that the queries naturally cluster around a finite set of optimal semantic cache points $\mathcal{M}^*$. Specifically, the maximum distance between any query generated near an optimal cache point and its nearest center in $\mathcal{S}_t$ decays as $\varepsilon_t = O(1/\sqrt{t})$, while the total number of active $\epsilon$-net centers required to cover the high-probability regions of $f(x)$ remains bounded by a constant $m_{\text{eff}}$.

**Justification of Assumption 2.** In a worst-case uniform continuous space, achieving a sublinear discretization error would require an $\epsilon$-net of size $m_{\max} \approx O(T^{d_e/2})$. This exponentially large number of centers would blow up the estimation regret, leading to a rate of $O(T^{\frac{d_e+1}{d_e+2}})$. However, Assumption 2 relies on the practical reality that real-world user queries do not spread out evenly or randomly; instead, they naturally group together around common themes and popular subjects. We empirically validate this behavior in our experimental datasets: as shown in the PCA (principal component analysis) projection of our synthetic stream (Figure 3 in the supplementary material) and the UMAP visualization of the combined Natural Questions and TriviaQA datasets (Figure 4 in the supplementary material), incoming queries strictly concentrate into dense, distinct semantic regions rather than filling the continuous embedding space uniformly. Because of this clustering, the algorithm forms centers within these dense regions. Therefore, the effective number of generated centers, $m_{\text{eff}}$, remains bounded by the geometry of the clusters, and is smaller than the theoretical maximum ($m_{\text{eff}} \le m_{\max}$).

Theorem 5.1 (Regret of CLCB-SC-LS-Cont). *Let $\mathcal{N}_T$ be the set of all unique queries observed up to horizon $T$, and let $\mathcal{M}^*_{\mathcal{N}_T}$ be the optimal cache of size $k$ selected from $\mathcal{N}_T$. Under Assumption 2, and setting the failure probability to $\delta = 1/T$, the cumulative regret of the* CLCB-SC-LS-Cont *algorithm is upper bounded by:*

$$Reg(T) \le O\Bigg(\underbrace{\beta_T\sqrt{T\gamma_T}}_{Cost} + \underbrace{\sqrt{m_{\mathit{eff}}T\log T}}_{Arrival}\Bigg) + O\Bigg(\underbrace{(1+2\alpha)L_g\sqrt{T}}_{Discretization} + \underbrace{km_{max}\log\log T}_{Switching}\Bigg)$$

*which achieves a strictly sublinear regret of $\tilde{O}(\sqrt{T})$. $\beta_T$ is the KRR confidence multiplier, $\gamma_T$ is the maximum information gain, and $\alpha$ is the approximation ratio of the Reverse Greedy oracle.*

**Comparison to the Discrete Regret Bound.** Theorem 5.1 establishes that CLCB-SC-LS-Cont achieves a sublinear regret of $\tilde{O}(\sqrt{T})$ in the continuous domain. This result is nontrivial when compared with discrete semantic caching bound of $O\left(\sqrt{mT\log(mT)}\log\log T\right)$ derived by [18]. In the discrete setting, the regret scales with the known, finite number of unique queries $m$. However, in our continuous setting, the query space is infinite. We replace the dependence on $m$ with two continuous analogues: the maximum RKHS information gain $\gamma_T$, and the effective covering number $m_{eff}$ of the arrival clusters. While the discrete algorithm [18] relies on independent arm estimation, our bound isolates the discretization penalty $O(L_g\sqrt{T})$ from the generalization capability of the KRR model, matching the $\sqrt{T}$ time-scaling of [18].

To prove the theorem, we establish the following lemmas, which overcome specific technical challenges in the continuous domain.

The following confidence bound for the cost relies on the result established in [5]. For any $t \ge 1$ and regularization $\lambda > 0$, let $\hat{c}_t(x) = \kappa_t(x)^\top(K_t+\lambda I)^{-1}y_{1:t}$ and $\sigma^2_{t,\lambda}(x) = \kappa(x,x) - \kappa_t(x)^\top(K_t+\lambda I)^{-1}\kappa_t(x)$ denote the standard KRR posterior mean and variance, respectively. We define the information gain as $\gamma_t = \frac{1}{2}\log\det(I + \frac{1}{\lambda}K_t)$.

Lemma 5.2 (Any-time concentration of cost and probability estimates). *Assume the true cost $c \in \mathcal{H}_k$ with $\|c\|_{\mathcal{H}_k} \le B$, and the noise sequence is conditionally $R$-sub-Gaussian. Let $m_{\mathit{eff}}$ be the effective covering number of $\mathcal{X}$ at radius $\epsilon$. For any $\delta \in (0,1)$ and horizon $T \ge 1$, define the confidence radius $\beta_t(x) = \left(B + R\sqrt{2(\gamma_t + \ln(2/\delta))}\right)\sigma_{t,\lambda}(x)$. The joint event $\mathcal{E} = \mathcal{E}_{\text{arr}} \cap \mathcal{E}_{\text{cost}}$ holds with probability at least $1-\delta$, for all $t \in [T]$, $x \in \mathcal{X}$ and every representative set $\mathcal{S}$ containing at most $m_{\mathit{eff}}$ representatives:*

$$\mathcal{E}_{\text{arr}} := \left\{\|\hat{\boldsymbol{w}}_t(\mathcal{S}) - \boldsymbol{w}(\mathcal{S})\|_1 \le 4\sqrt{\frac{2m_{\mathit{eff}}\ln(2T^2/\delta)}{t}}\right\},$$
$$\mathcal{E}_{\text{cost}} := \{|\hat{c}_{t-1}(x) - c(x)| \le \beta_{t-1}(x)\}.$$

*Consequently, the lower confidence bound $\underline{c}_t(x) = \hat{c}_t(x) - \beta_t(x)$ satisfies $\underline{c}_t(x) \le c(x)$ with high probability.*

Lemma 5.2 highlights a fundamental departure from discrete caching. It allows us to construct valid, any-time optimistic lower bounds $\underline{c}$ that generalize across the continuous space based on kernel similarity, forming the basis of our active exploration strategy.

Lemma 5.3 (Number of cache switches). *Let $m_{\max}$ be the maximum covering number of the compact query space $\mathcal{X}$ at radius $\epsilon$, such that $m_{\max} \le (D\sqrt{d_e}/\epsilon)^{d_e}$, where $D$ is the diameter in the embedding space. The total number of cache switches triggered by the* CLCB-SC-LS-Cont *algorithm up to horizon $T$ is bounded by:*

$$\mathbb{E}\left[\sum_{u\in S_T}\tau_u(T)\right] \le O(m_{\max}\log\log T),$$
$$\mathbb{E}[\tau_f(T)] \le O(\log\log T).$$

Finally, Lemma 5.3 shows how our algorithm avoids excessive cache updates, even as new query clusters are discovered on the fly. In a continuous space, we cannot rely on a fixed set of update rules because the cluster centers are generated dynamically as queries arrive. To solve this, we tie the cache update thresholds to the geometric limits of the space, i.e., the max covering number $m_{\max}$.

# 6 Experiments

In this section, we present the experiments for Alg. 2 and 3. Additional results including ablations may be found in supplementary.

**Experimental Setup.** We consider two types of query universes. The first is a synthetic benchmark consisting of a fixed collection of natural-language prompts ($m = 50$) spanning diverse domains and a broad range of prompt lengths, from short keyword-like requests to long conversational questions (similar to the dataset used in [18] with prompts that are more varied in their token length) with added Gaussian noise to the embeddings of the sampled prompts to better simulate a continuous query space. The second is a real-data benchmark we formed by combining 2,500 queries from Natural Questions (NQ) [12] and 2,500 queries from TriviaQA [11] datasets into a single universe of 5,000 real queries. In both settings, query embeddings $x$ are obtained using sentence-transformer models [23] that map queries to a 384-dimensional ($d_e$) vector representation, and semantic similarity $d$ is measured using Euclidean distance after normalization of embeddings, with distances further normalized to lie in the range $[0, 1]$. Serving costs are derived from GPT-2 tokenizer lengths. For both benchmarks, the expected serving cost $c(x)$ is the min-max normalized token length of each query $x$.

We consider a diverse arrival process across synthetic and real-world datasets. In the synthetic setting, arrivals are generated by first uniformly sampling a query from the discrete query universe and then perturbing its embedding with Gaussian noise. This produces semantically nearby but non-identical arrivals and allows the benchmark to model a continuous query space rather than a purely discrete one. In contrast, in the NQ+TriviaQA setting, each arrival is drawn directly from the pool of real dataset queries, so the stream consists of actual queries from NQ and TriviaQA rather than synthetically perturbed embeddings. To make the real data stream more challenging, the sampling process is bursty and non-uniform where bursts alternate between the two datasets, and within each dataset, queries are sampled from a heavy-tailed popularity distribution.

**Offline Setting.** In the offline setting, we evaluate Algorithm 2 on the synthetic dataset for two ablations using cache size $k = 5$: an $\epsilon$ ablation for a fixed stream length of $t = 1000$, and a stream length ablation where $\epsilon$ is fixed at 0.4 (Figure 2 (a)-(b)). Results are averaged across 20 different seeds for both ablations. We compare against CUCB-SC [18], Greedy, Epsilon-Greedy, and LFU [14]. Greedy selects the cache using the same empirical frequency-cost score as Epsilon-Greedy but always exploits, whereas LFU ignores cost entirely and caches only the most frequently observed queries. In the $\epsilon-$ablation, CUCB-SC-Cont achieves the lowest suboptimality gap for small and moderate $\epsilon$, but degrades when $\epsilon$ becomes too large and the discretization becomes too coarse, showing the value of tuning this parameter. CUCB-SC-Cont displays a 52.95% improvement in suboptimality gap over CUCB-SC when $\epsilon = 0.4$. In the stream length ablation, CUCB-SC-Cont remains the strongest method overall; while CUCB-SC improves steadily with more offline data and eventually becomes competitive, the other baselines do not show consistent improvement with the increase in stream length. CUCB-SC-Cont achieves significantly better performance than the other baselines when the stream length is small, showing the benefit of generalization via $\epsilon-$net construction and KRR (73.32% improvement over CUCB-SC when stream length = 100).

**Online Setting.** In the online setting, the CLCB-SC-LS-Cont algorithm demonstrates superior performance in minimizing average regret across both synthetic and real datasets. For the synthetic online dataset, CLCB-SC-LS-Cont achieves the lowest final average regret at 0.0083, which is a 72.41% decrease compared to the most competitive baseline, CLCB-SC-LS [18], with a runtime increase of 47.8%. Figure 2 (c) illustrates that while traditional baselines like LFU [14] and Epsilon-Greedy exhibit much higher regret peaks and slower convergence, the CLCB-SC-LS-Cont variant converges rapidly. On the combined NQ + TriviaQA dataset, the algorithm maintains this advantage with a final average regret of 0.0061, representing a 63.27% reduction over Greedy. While Figure 2 (f) shows a more significant runtime trade-off for the real combined dataset, the algorithm still outperforms the CLCB-SC-LS baseline by 43.14% in regret while maintaining a nearly identical latency of approximately 250 seconds. The increase in runtime when comparing plots (d) and (f) stems from the fact that for the real dataset, the CLCB-SC-LS-Cont and CLCB-SC-LS trigger more frequent switches.

# 7 Conclusion

We establish the first theoretical framework for semantic caching in a continuous space under uncertainty for low-cost LLM serving. By introducing $\epsilon-$net discretization and KRR, we studied the caching problem across different settings. Our theoretical analysis establishes suboptimality, sublinear regret guarantees, coupled with empirical evaluations demonstrating notable reduction in regret.

# Acknowledgments

This work was supported by the Department of Energy under grant DESC002565 and the National Science Foundation under grant 2533813.

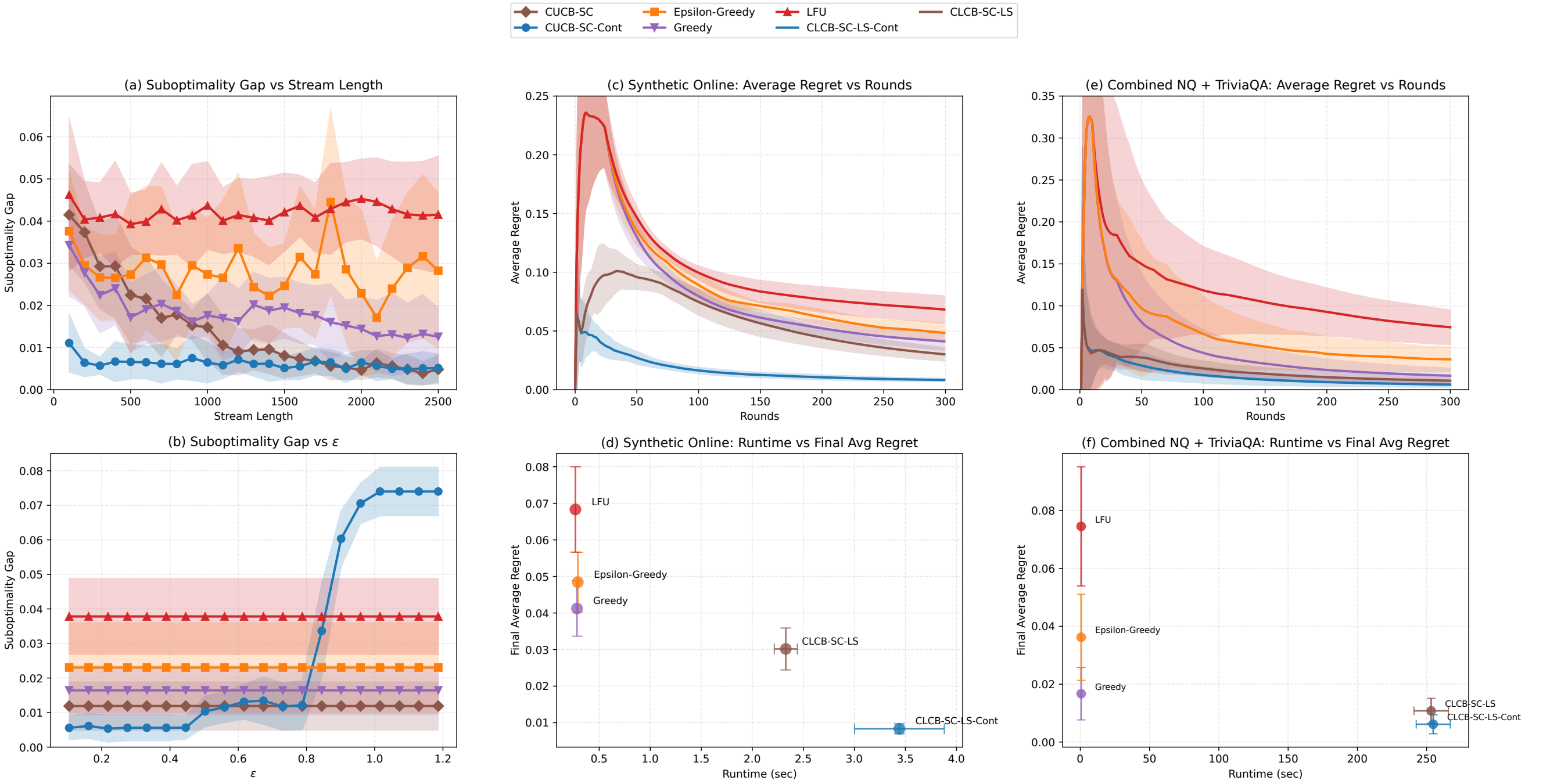

**Figure 2: (a) Variation of the suboptimality gap versus stream length in the offline setting (b) Sensitivity analysis showing the variation of the suboptimality gap with respect to the $\epsilon$-net radius $\epsilon$ (c) Online performance on the synthetic dataset showing the average regret across rounds (d) Efficiency trade-off for synthetic experiments, showing total runtime against the final average regret (e) Online performance on the combined dataset showing the average regret across rounds (f) Efficiency trade-off for real dataset experiments, showing total runtime against the final average regret**

# A Appendix

Here we provide the missing proofs in the main body for the presented algorithms as well as additional empirical results. We also introduce our Frozen algorithm and its corresponding proofs and empirical results.

## A.1 Proofs for CUCB-SC-Cont and CLCB-SC-LS-Cont

Proof of Lemma 3.1. We prove NP-hardness by reduction from the discrete semantic caching problem studied in [18], which is NP-hard. Consider an arbitrary instance of the discrete problem with a finite query set $Q = \{q_1, \ldots, q_n\} \subset \mathcal{X}$, query probabilities $p(q)$, recomputation costs $c(q)$, and mismatch penalty $\phi(d(q, M))$. Construct an instance of the continuous problem by defining the arrival distribution $f$ as a discrete measure supported on $Q$: $f(x) = \sum_{q \in Q} p(q)\, \delta(x - q)$, where $\delta(\cdot)$ denotes the Dirac delta function.

Substituting this choice of $f$ into the continuous loss in (2), we obtain

$$\mathcal{L}(\mathcal{M}; f, c, d) = \int_{\mathcal{X}} f(x) \min\{c(x), \phi(d(x, \mathcal{M}))\}\, dx$$
$$= \sum_{q \in Q} p(q) \min\{c(q), \phi(d(q, \mathcal{M}))\}$$

Thus, on this restricted class of instances, the continuous semantic caching problem reduces exactly to the discrete semantic caching problem of [18]. Therefore, if we could solve the continuous problem in polynomial time, then we could also solve the discrete problem in polynomial time. Since the latter is NP-hard, the continuous semantic caching problem is also NP-hard. □

Proof of Lemma 3.2. Fix an $\epsilon$−net $\mathcal{S} = \{x_i\}_{i=1}^m \subset \mathcal{X}$ and its Voronoi partition $\{V_i\}_{i=1}^m$. Define true cell masses $w_i = \int_{V_i} f(x)dx$. For any cache $\mathcal{M}$ with $|\mathcal{M}| \le k$, given the definition of $g_{\mathcal{M}}(x)$ we have

$$\mathcal{L}(\mathcal{M}) = \int_{\mathcal{X}} f(x) g_{\mathcal{M}}(x) dx, \ \widehat{\mathcal{L}}(\mathcal{M}) = \sum_{i=1}^m w_i g_{\mathcal{M}}(x_i)$$

Given that $g_{\mathcal{M}}(x)$ is $L_g$−Lipschitz, if $c$ is $L_c$−Lipschitz and $\phi \circ d(\cdot, u)$ is $L_\phi$−Lipschitz in $u$, then $L_g = \max(L_c, L_\phi)$. For each cell $V_i$, every $x \in V_i$ satisfies $d(x, x_i) \le \epsilon$, hence $|g_{\mathcal{M}}(x) - g_{\mathcal{M}}(x_i)| \le L_g \epsilon$. Integrating over $V_i$ and summing over $i$ yields $|\mathcal{L}(\mathcal{M}) - \widehat{\mathcal{L}}(\mathcal{M})| \le L_g \epsilon$. □

Proof of Theorem 3.3. The approximation guarantee follows from [18] and [10], since Algorithm 1 optimizes over subsets of $\mathcal{S}$. For the second claim, $\mathcal{L}(\mathcal{M}) - \alpha\mathcal{L}(\mathcal{M}^*) = [\mathcal{L}(\mathcal{M}) - \widehat{\mathcal{L}}(\mathcal{M})] + [\widehat{\mathcal{L}}(\mathcal{M}) - \alpha\widehat{\mathcal{L}}(\mathcal{M}^*_{\mathcal{S}})] + \alpha[\widehat{\mathcal{L}}(\mathcal{M}^*_{\mathcal{S}}) - \widehat{\mathcal{L}}(\mathcal{M}^*)] + \alpha[\widehat{\mathcal{L}}(\mathcal{M}^*) - \mathcal{L}(\mathcal{M}^*)]$. The first and last brackets are at most $L_g\epsilon$ by Lemma 3.2. For the third, map each $u \in \mathcal{M}^*$ to $\pi(u) \in \mathcal{S}$ with $d(u, \pi(u)) \le \epsilon$. Then $d(x, \pi(\mathcal{M}^*)) \le d(x, \mathcal{M}^*) + \epsilon$ for every $x$, so $\widehat{\mathcal{L}}(\mathcal{M}^*_{\mathcal{S}}) \le \widehat{\mathcal{L}}(\pi(\mathcal{M}^*)) \le \widehat{\mathcal{L}}(\mathcal{M}^*) + L_\phi\epsilon \le \widehat{\mathcal{L}}(\mathcal{M}^*) + L_g\epsilon$. □

Proof of Lemma 4.2. Assume a fixed finite partition of the query space $\mathcal{X}$ into $\{V_1, .., V_m\}$. The true weights are defined as $w_i = P_{X\sim f}(X \in V_i) = \int_{V_i} f(x)dx$. Hence we also have $\sum_i w_i = 1$. Conditionally on $\mathcal{D}_1$, from $n_2$ i.i.d. draws $X_1, ..., X_{n_2} \sim f$ we have the empirical weights $\hat{w}_i = \frac{1}{n_2}\sum_{t=1}^{n_2} \mathbf{1}\{X_t \in V_i\}$ for $i = 1, ..., m$.

$$||\hat{\boldsymbol{w}}_i - \boldsymbol{w}_i||_1 = \sum_{i=1}^m |\hat{w}_i - w_i|$$
$$\Delta_i = \hat{w}_i - w_i$$
$$\Delta^+ = \sum_{i:\Delta_i \ge 0} \Delta_i$$
$$\Delta^- = \sum_{i:\Delta_i < 0} (-\Delta_i)$$

Given these definitions it follows that $\sum_i \Delta_i = 0$ since the weights are normalized. From which it follows that $\Delta^+ = \Delta^-$ and $||\hat{\boldsymbol{w}}_i - \boldsymbol{w}_i||_1 = \sum_i |\Delta_i| = 2\Delta^+$. So the event is $||\hat{\boldsymbol{w}}_i - \boldsymbol{w}_i||_1 \ge \epsilon_b$ which means $\Delta^+ \ge \epsilon_b/2$. Define $A : \{i : \Delta_i \ge 0\}$, then $\Delta^+ = \sum_{i\in A}(\hat{w}_i - w_i) = \widehat{W}(A) - W(A)$ where $\widehat{W}(A) = \sum_{i\in A} \hat{w}_i$ and $W(A) = \sum_{i\in A} w_i$. So the event $||\hat{\mathbf{w}}_i - \mathbf{w}_i||_1 \ge \epsilon_b$ is equivalent to $\exists A \subseteq [m]$ s.t. $\widehat{W}(A) - W(A) \ge \epsilon_b/2$. There are $2^m$ possible subsets of $[m]$. Two of them are trivial, i.e. if $A = \emptyset$ or $A = [m]$ which means there is no deviation. So we can union bound over $2^m - 2$ sets. $\widehat{W}(A)$ is the empirical frequency of the binary event $\{X \in \bigcup_{i\in A} V_i\}$, it is a mean of i.i.d. Bernoulli variables $(W(A))$. Hence by Hoeffding's inequality we have:

$$\mathbb{P}(\widehat{W}(A) - W(A) \ge t) \le \exp(-2n_2t^2)$$
$$\mathbb{P}(||\hat{\boldsymbol{w}}_i - \boldsymbol{w}_i||_1 \ge \epsilon_b) \le (2^m - 2)\exp\left(-\frac{n_2\epsilon_b^2}{2}\right)$$

This result directly implies that

$$||\hat{\boldsymbol{w}} - \boldsymbol{w}||_1 \le \sqrt{\frac{2\ln(2^m/\delta)}{n_2}} \le \sqrt{\frac{2m\ln(2/\delta)}{n_2}} \text{ w.p.} \ge 1 - \delta$$

*Cost.* Condition on the design $\mathcal{X} = (X_t, Z_t)_{t\le n}$. With cost-independent logging, the noises are independent, zero-mean and $R$-sub-Gaussian given $\mathcal{X}$. The centers depend only on $\{X_t\}_{t\in\mathcal{D}_1}$, so they are fixed as well. Let $h_x = (K + \lambda I)^{-1}\kappa_x$. Then

$$\hat{c}(x) - c(x) = \underbrace{h_x^\top \mathbf{f} - c(x)}_{\text{bias}} + \underbrace{h_x^\top \eta}_{\text{noise}}.$$

- **Bias.** By the reproducing property and Cauchy–Schwarz, $|\text{bias}| \le B\,\|\kappa(x,\cdot) - \sum_s h_{x,s}\kappa(z_s,\cdot)\|$. The squared norm equals $\sigma_\lambda^2(x) - \lambda\|h_x\|^2 \le \sigma_\lambda^2(x)$, using $\kappa_x^\top h_x = h_x^\top(K + \lambda I)h_x$.
- **Noise.** $h_x^\top\eta$ is $R\|h_x\|$-sub-Gaussian. The same identity gives $\|h_x\| \le \sigma_\lambda(x)/\sqrt{\lambda}$. A union bound over the $m$ centers gives $|h_x^\top\eta| \le c_\lambda\sigma_\lambda(x)$ with probability at least $1 - \delta$.
- **Pessimism.** From $\hat{c} - \Delta \le c \le \hat{c} + \Delta$ and $c \le 1$: $\bar{c} \ge c$ and $\bar{c} - c \le 2\Delta$.

*Posterior Width.* This follows directly from Corollary A.3. Combining all parts gives the stated result. □

Proof of Lemma 5.2. We allocate a failure probability of $\delta/2$ to each of the two events to satisfy a global union bound.

**1. Arrival Concentration ($\mathcal{E}_{\text{arr}}$):**

We first define $\Lambda_T = m_{\text{eff}}\log(T + 1) + m_{\text{eff}}\log 2 + \log\frac{2T}{\delta}$. Fix a sample size $t$ and an ordered tuple of distinct indices $I = (i_1, ..., i_m)$, $m \le m_{\text{eff}}$, from $\{1, ..., \min\{t + 1, T\}\}$. Let $\mathcal{S}_I = (X_{i_1}, ..., X_{i_m})$ be the associated representatives. Condition on the selected query values $X_I = (X_{i_1}, ..., X_{i_m})$. This fixes the partition. Crucially, we do not condition on the event that the algorithm chooses $I$. Consequently, all unselected queries remain independent with density $f$.

For any sign vector $o \in \{-1,1\}^m$, define

$$h_o(x) = \sum_{j=1}^{m} o_j \mathbf{1}\{x \in V(X_{i_j}; \mathcal{S}_I)\}$$

and

$$\theta_o = \int_{\mathcal{X}} h_o(x) f(x) dx = o^\top \boldsymbol{w}(\mathcal{S}_I)$$

Because the cells partition $\mathcal{X}$, $h_o(x) \in \{-1,1\}$. For $s \notin I$, the variables

$$Y_s = h_o(X_s) - \theta_o$$

are conditionally independent, have conditional mean zero, and lie in an interval of width 2. Applying Hoeffding's inequality therefore gives

$$\Pr\left( \sum_{\substack{s<t \\ s \notin I}} Y_s > \sqrt{2t\Lambda_T} \;\middle|\; X_I \right) \le e^{-\Lambda_T}.$$

Using $t$ instead of the number of unselected observations only enlarges the threshold. If there are no unselected observations, the sum is zero. The selected observations among the first $t$ queries satisfy the deterministic bound

$$\sum_{\substack{s<t \\ s \in I}} Y_s \le 2m \le 2m_{\text{eff}}$$

Hence, except on an event of probability at most $e^{-\Lambda_T}$,

$$o^\top (\hat{\boldsymbol{w}}_t(\mathcal{S}) - \boldsymbol{w}(\mathcal{S})) \le \sqrt{\frac{2\Lambda_T}{t}} + \frac{2m_{\text{eff}}}{t}$$

There are at most $(T+1)^{m_{\text{eff}}}$ possible representative-index tuples, at most $2^{m_{\text{eff}}}$ sign vectors per tuple and $T$ sample sizes. Hence the union-bound failure probability is at most $T(T+1)^{m_{\text{eff}}} 2^{m_{\text{eff}}} e^{-\Lambda_T} = \delta/2$. Finally, use

$$\|z\|_1 = \max_{o \in \{-1,1\}^m} o^\top z$$

and the fact that two probability vectors have $\ell_1$ distance at most 2. It can also be shown that

$$\begin{aligned} \omega_t &= \min\left\{2, \sqrt{\frac{2\Lambda_T}{t}} + \frac{2m_{\text{eff}}}{t}\right\} \le 4\sqrt{\frac{2m_{\text{eff}} \log(2T^2/\delta)}{t}} \\ &= O\left(\sqrt{\frac{m_{\text{eff}} \log(2T^2/\delta)}{t}}\right) \end{aligned}$$

which finalizes the claim.

**2. Cost Concentration ($\mathcal{E}_{\text{cost}}$):**

Instead of analyzing the fixed-sample KRR bias and noise terms as in Lemma 4.2, the online setting requires an any-time bound. This follows directly from Theorem 2 in [5]. Applying this result with a failure probability of $\delta/2$ directly yields $\mathcal{E}_{\text{cost}}$. By the union bound, the two events hold simultaneously with probability at least $1-\delta$. □

Lemma A.1 (Local count). *Fix the partition, i.e., condition on $\mathcal{D}_1$. Assume the following where $Z_t = 1$ if the logging policy invokes the LLM for $x_t$ and observes its serving cost and 0 otherwise:*

- *(P) Propensity: $P(Z_t = 1 | X_t = x, \mathcal{F}_{t-1}) \ge \nu_i$ for all $x \in V_i$*
- *(M) Lower mass: $P(X \in B(x_i, v) \cap V_i) \ge w_i \underline{\nu}_i (v/D_i)^p$ for $v \in (0, D_i]$.*

*We define $D_i := \sup_{x,y \in V_i} \|x-y\|$. Since every point of $V_i$ is within $\epsilon$ of $x_i$ by covering, $D_i \le 2\epsilon$ and $V_i \subseteq B(x_i, r)$ for $r \ge D_i$, (M) implies*

$$P(X \in B(x_i, r) \cap V_i) \ge w_i \underline{\nu}_i (r/2\epsilon)^p \quad \text{for all } r \in (0, 2\epsilon].$$

*Let $N_i(r) = \sum_{t \in \mathcal{D}_2} \mathbf{1}\{X_t \in B(x_i, r) \cap V_i, Z_t = 1\}$. If $n_2 w_i \nu_i \underline{\nu}_i (r/2\epsilon)^p \ge 8\ln(1/\delta)$, then*

$$P\left(N_i(r) < \frac{1}{2} n_2 w_i \nu_i \underline{\nu}_i (r/2\epsilon)^p\right) \le \delta$$

Proof. Let $B_t$ be the indicator in $N_i(r)$. Since $X_t$ is independent of $\mathcal{F}_{t-1}$, (P) and (M) give

$$\mathbb{E}[B_t | \mathcal{F}_{t-1}] \ge \nu_i P(X \in B(x_i, r) \cap V_i) \ge q := w_i \nu_i \underline{\nu}_i (r/2\epsilon)^p.$$

For $\theta > 0$, convexity gives $\mathbb{E}[e^{-\theta B_t} | \mathcal{F}_{t-1}] \le \exp(-(1-e^{-\theta})q)$, so $\mathbb{E} e^{-\theta N_i} \le e^{-n_2(1-e^{-\theta})q}$. Take $\theta = \ln 2$ and apply Markov's inequality:

$$P(N_i \le n_2 q/2) \le e^{-n_2 q(1-\ln 2)/2} \le e^{-n_2 q/8}$$

□

Lemma A.2 (Regularized width via local counts). *Let $\kappa$ be the unit-variance RBF kernel with length $\ell_k$. Fix $\lambda > 0$. Let $O = \{z_1, \dots, z_\ell\}$ be the inputs with observed cost and define*

$$\sigma_\lambda^2(x) = \kappa(x,x) - \kappa_x^\top (K + \lambda I)^{-1} \kappa_x.$$

*For $r > 0$, let $N_x(r) = \#\{s : \|z_s - x\| \le r\}$. Then*

$$\sigma_\lambda^2(x) \le \min\left\{1, 2(1 - e^{-r^2/(2\ell_k^2)}) + \frac{\lambda}{N_x(r)}\right\} \le \min\left\{1, \frac{r^2}{\ell_k^2} + \frac{\lambda}{N_x(r)}\right\}$$

*More generally, for any subset $I$ of the observations,*

$$\sigma_\lambda^2(x) \le \max_{s \in I} \|\kappa(x,\cdot) - \kappa(z_s,\cdot)\|_{\mathcal{H}}^2 + \frac{\lambda}{|I|}$$

Proof. Define

$$\begin{aligned} Q(a) &= \left\| \kappa(x,\cdot) - \sum_s a_s \kappa(z_s,\cdot) \right\|_{\mathcal{H}}^2 + \lambda \|a\|_2^2 \\ &= \kappa(x,x) - 2a^\top \kappa_x + a^\top (K + \lambda I) a. \end{aligned}$$

$Q$ is strictly convex, with minimizer $a^* = (K+\lambda I)^{-1}\kappa_x$ and minimum value $\sigma_\lambda^2(x)$. So $\sigma_\lambda^2(x) \le Q(a)$ for every $a$. Put $a_s = 1/|I|$ on $I$ and 0 elsewhere. The first term is at most $\max_{s \in I} \|\kappa_x - \kappa_{z_s}\|^2$ because the norm of an average is at most the largest norm. The penalty term equals $\lambda |I|/|I|^2 = \lambda/|I|$. For the RBF kernel, $\|\kappa_x - \kappa_z\|^2 = 2(1 - e^{-\|x-z\|^2/2\ell_k^2}) \le \|x-z\|^2/\ell_k^2$. Take $I = \{s : \|z_s - x\| \le r\}$. Setting $a = 0$ gives the bound 1. □

Corollary A.3 (Control of the posterior width). *Under (P) and (M), with probability at least $1-\delta$, simultaneously for all $i \in [m]$:*

$$\sigma_\lambda(x_i) \le \sqrt{2} \left( \frac{2\lambda (2\epsilon/\ell_k)^p}{n_2 w_i \nu_i \underline{\nu}_i} \right)^{\frac{1}{p+2}},$$

*provided*

$$n_2 w_i \nu_i \underline{\nu}_i \ge N_0 := \max\left\{ (2\epsilon/\ell_k)^p (2\lambda)^{-p/2} \big(8\ln(m/\delta)\big)^{\frac{p+2}{2}}, \frac{\lambda \ell_k^2}{2\epsilon^2} \right\}.$$

*The KRR may use all cost observations, including those in $\mathcal{D}_1$.*

Proof. Let

$$X = n_2 w_i v_i \underline{v_i}, \qquad b_i = \frac{2\lambda(2\epsilon)^p}{X}, \qquad r_i = (\ell_k^2 b_i)^{1/(p+2)}.$$

(1) The second burn-in condition is equivalent to $r_i \le 2\epsilon$.
(2) A direct computation gives

$$X\left(\frac{r_i}{2\epsilon}\right)^p = X^{2/(p+2)}(2\lambda\ell_k^2)^{p/(p+2)}(2\epsilon)^{-2p/(p+2)}.$$

So the first burn-in condition is exactly

$$X\left(\frac{r_i}{2\epsilon}\right)^p \ge 8\ln(m/\delta).$$

(3) Lemma A.1 at level $\delta/m$ then gives

$$N_{x_i}(r_i) \ge \frac{1}{2}X\left(\frac{r_i}{2\epsilon}\right)^p.$$

since $N_{x_i}(r_i) \ge N_i(r_i)$.
(4) By Lemma A.2,

$$\sigma^2 \le \frac{r_i^2}{\ell_k^2} + b_i r_i^{-p}.$$

At $r_i$ the two terms are equal, each $(b_i/\ell_k^p)^{2/(p+2)}$, so

$$\sigma \le \sqrt{2}(b_i/\ell_k^p)^{1/(p+2)}.$$

(5) Take a union bound over $i$.

□

Proof of Theorem 4.1. We ignore $\mathcal{S}$ in the loss function when contexts are clear $\widehat{\mathcal{L}}(\mathcal{M};\mathcal{S},\boldsymbol{w},\boldsymbol{c}) = \widehat{\mathcal{L}}(\mathcal{M};\boldsymbol{w},\boldsymbol{c})$. We apply Lemma 4.2 with confidence $\delta/3$ for each of its three events. Since we have that $n_2 \ge \max_i N_0/(w_i v_i \underline{v_i})$, the events $\mathcal{E}_{\text{arvl}}$, $\mathcal{E}_{\text{cost}}$ and $\mathcal{E}_{\text{width}}$ hold simultaneously with probability at least $1-\delta$, and on them, for all $i \in [m]$,

$$\begin{aligned}
&\|\hat{\boldsymbol{w}} - \boldsymbol{w}\|_1 \le \sqrt{\frac{2m\ln(6/\delta)}{n_2}}, \\
&0 \le \bar{c}(x_i) - c(x_i) \le 2(B + c_\lambda)\,\sigma_\lambda(x_i), \\
&\sigma_\lambda(x_i) \le \sqrt{2}\left(\frac{2\lambda(2\epsilon/\ell_k)^p}{n_2 w_i v_i \underline{v_i}}\right)^{\frac{1}{p+2}},
\end{aligned}$$

where $c_\lambda = R\sqrt{2\ln(6m/\delta)/(\lambda)}$.

*Step 1: Discrete subopt.* Let $\mathcal{M}_{\mathcal{S}}^* \in \arg\min_{\mathcal{M}\subseteq\mathcal{S},\,|\mathcal{M}|\le k} \widehat{\mathcal{L}}(\mathcal{M};\boldsymbol{w},\boldsymbol{c})$ be the best cache among the representatives. Under the events above,

$$\begin{aligned}
&\widehat{\mathcal{L}}(\hat{\mathcal{M}};\boldsymbol{w},\boldsymbol{c}) - \alpha\widehat{\mathcal{L}}(\mathcal{M}_{\mathcal{S}}^*;\boldsymbol{w},\boldsymbol{c}) \\
&= \underbrace{\alpha\widehat{\mathcal{L}}(\mathcal{M}_{\mathcal{S}}^*;\hat{\boldsymbol{w}},\bar{\boldsymbol{c}}) - \alpha\widehat{\mathcal{L}}(\mathcal{M}_{\mathcal{S}}^*;\boldsymbol{w},\boldsymbol{c})}_{\text{estimation gap}} \\
&\quad + \underbrace{\widehat{\mathcal{L}}(\hat{\mathcal{M}};\hat{\boldsymbol{w}},\bar{\boldsymbol{c}}) - \alpha\widehat{\mathcal{L}}(\mathcal{M}_{\mathcal{S}}^*;\hat{\boldsymbol{w}},\bar{\boldsymbol{c}})}_{\text{greedy gap}} \\
&\quad + \underbrace{\widehat{\mathcal{L}}(\hat{\mathcal{M}};\boldsymbol{w},\boldsymbol{c}) - \widehat{\mathcal{L}}(\hat{\mathcal{M}};\hat{\boldsymbol{w}},\bar{\boldsymbol{c}})}_{\text{pessimism gap}} \\
&\le \alpha\widehat{\mathcal{L}}(\mathcal{M}_{\mathcal{S}}^*;\hat{\boldsymbol{w}},\bar{\boldsymbol{c}}) - \alpha\widehat{\mathcal{L}}(\mathcal{M}_{\mathcal{S}}^*;\boldsymbol{w},\boldsymbol{c}) \\
&\quad + \widehat{\mathcal{L}}(\hat{\mathcal{M}};\boldsymbol{w},\bar{\boldsymbol{c}}) - \widehat{\mathcal{L}}(\hat{\mathcal{M}};\hat{\boldsymbol{w}},\bar{\boldsymbol{c}}) \\
&\le \alpha\sum_{i=1}^m w_i\big(\bar{c}(x_i) - c(x_i)\big) + (\alpha+1)\,\|\hat{\boldsymbol{w}} - \boldsymbol{w}\|_1 \\
&\le \alpha\sum_{i=1}^m w_i\big(\bar{c}(x_i) - c(x_i)\big) + (\alpha+1)\sqrt{\frac{2m\ln(6/\delta)}{n_2}},
\end{aligned}$$

where the equality is due to adding and subtracting $\alpha\widehat{\mathcal{L}}(\mathcal{M}_{\mathcal{S}}^*;\hat{\boldsymbol{w}},\bar{\boldsymbol{c}})$ and $\widehat{\mathcal{L}}(\hat{\mathcal{M}};\hat{\boldsymbol{w}},\bar{\boldsymbol{c}})$. For the first inequality, the greedy gap is $\le 0$ by the approximation ratio guarantee of `Cont_Reverse_Greedy`, since $\hat{\mathcal{M}}$ is its output on $(\mathcal{S},\hat{\boldsymbol{w}},\bar{\boldsymbol{c}})$ and $\mathcal{M}_{\mathcal{S}}^* \subseteq \mathcal{S}$ is feasible; moreover $\widehat{\mathcal{L}}(\hat{\mathcal{M}};\boldsymbol{w},\boldsymbol{c}) \le \widehat{\mathcal{L}}(\hat{\mathcal{M}};\boldsymbol{w},\bar{\boldsymbol{c}})$ because $\bar{\boldsymbol{c}} \ge \boldsymbol{c}$ by Lemma 4.2 and $\widehat{\mathcal{L}}(\mathcal{M};\boldsymbol{w},\boldsymbol{c})$ is monotone in $\boldsymbol{c}$. For the second inequality, we write the estimation gap as $\alpha[\widehat{\mathcal{L}}(\mathcal{M}_{\mathcal{S}}^*;\hat{\boldsymbol{w}},\bar{\boldsymbol{c}}) - \widehat{\mathcal{L}}(\mathcal{M}_{\mathcal{S}}^*;\boldsymbol{w},\bar{\boldsymbol{c}})] + \alpha[\widehat{\mathcal{L}}(\mathcal{M}_{\mathcal{S}}^*;\boldsymbol{w},\bar{\boldsymbol{c}}) - \widehat{\mathcal{L}}(\mathcal{M}_{\mathcal{S}}^*;\boldsymbol{w},\boldsymbol{c})]$. Both weight differences (here and in the pessimism gap) are bounded by $\|\hat{\boldsymbol{w}} - \boldsymbol{w}\|_1$ by Lemma 4 in [18]. The cost difference is bounded by $\sum_i w_i(\bar{c}(x_i) - c(x_i))$, since $\min\{a,t\} - \min\{b,t\} \le a - b$ for $a \ge b$. The last inequality is due to $\mathcal{E}_{\text{arvl}}$.

*Step 2: Cost estimation term.* On $\mathcal{E}_{\text{cost}} \cap \mathcal{E}_{\text{width}}$,

$$\begin{aligned}
\sum_{i=1}^m w_i\big(\bar{c}(x_i) - c(x_i)\big) &\le 2(B + c_\lambda)\sum_{i=1}^m w_i\,\sigma_\lambda(x_i) \\
&\le 2\sqrt{2}(B + c_\lambda)\left(\frac{2\lambda(2\epsilon/\ell_k)^p}{n_2}\right)^{\frac{1}{p+2}} \\
&\quad \cdot \sum_{i=1}^m w_i^{\frac{p+1}{p+2}}(v_i\underline{v_i})^{-\frac{1}{p+2}}.
\end{aligned}$$

We apply Hölder's inequality with $u = \frac{p+2}{p+1}$, $q = p+2$ (so that $1/u + 1/q = 1$), $a_i = w_i^{(p+1)/(p+2)}$ and $b_i = (v_i\underline{v_i})^{-1/(p+2)}$:

$$\begin{aligned}
\sum_{i=1}^m a_i b_i &\le \left(\sum_{i=1}^m a_i^u\right)^{1/u}\left(\sum_{i=1}^m b_i^q\right)^{1/q} \\
&= \left(\sum_{i=1}^m w_i\right)^{\frac{p+1}{p+2}}\left(\sum_{i=1}^m (v_i\underline{v_i})^{-1}\right)^{\frac{1}{p+2}} = C_v^{\frac{1}{p+2}},
\end{aligned}$$

where we used $\sum_i w_i = 1$ and $C_v = \sum_{i=1}^m (v_i\underline{v_i})^{-1}$. Hence

$$\sum_{i=1}^m w_i\big(\bar{c}(x_i) - c(x_i)\big) \le 2\sqrt{2}(B + c_\lambda)\left(\frac{2\lambda(2\epsilon/\ell_k)^p\,C_v}{n_2}\right)^{\frac{1}{p+2}},$$

and combining with Step 1,

$$\widehat{\mathcal{L}}(\hat{\mathcal{M}};\boldsymbol{w},\boldsymbol{c}) - \alpha\widehat{\mathcal{L}}(\mathcal{M}^*_{\mathcal{S}};\boldsymbol{w},\boldsymbol{c}) \le 2\sqrt{2}\,\alpha(B+c_\lambda)\left(\frac{2\lambda(2\epsilon/\ell_k)^p\,C_v}{n_2}\right)^{\frac{1}{p+2}} + (\alpha+1)\sqrt{\frac{2m\ln(6/\delta)}{n_2}}.$$

*Step 3: Relating $\epsilon$ to $m$.* We can also relate $\epsilon$ to $m$. Let's define the diameter of the query space $(\mathcal{X}, d)$ as $D$. Assuming we are using Euclidean distance with unit norm embeddings, it follows that $D \le 2$. The following relation is known for an $\epsilon-$net's covering number, which we defined as $m$:

$$m \le \left(\frac{D\sqrt{d_e}}{\epsilon}\right)^{d_e},$$

from which it follows that

$$\epsilon \le \frac{D\sqrt{d_e}}{m^{1/d_e}} \le \frac{2\sqrt{d_e}}{m^{1/d_e}}.$$

*Step 4: From the discrete to the continuous objective.* We further assume that $\phi$ is $L_\phi$-Lipschitz, so that $u \mapsto \phi(d(x,u))$ is $L_\phi$-Lipschitz for every $x$, and without loss of generality $L_\phi \le L_g$ (under the sufficient condition above, $L_g = \max\{L_c, L_\phi\}$). Under the results of Theorem 3.3, summing with the weights $w_i$ and using the optimality of $\mathcal{M}^*_{\mathcal{S}}$ over subsets of $\mathcal{S}$,

$$\widehat{\mathcal{L}}(\mathcal{M}^*_{\mathcal{S}}) \le \widehat{\mathcal{L}}(\pi(\mathcal{M}^*)) \le \widehat{\mathcal{L}}(\mathcal{M}^*) + L_\phi\epsilon.$$

Therefore

$$\begin{aligned}
&\mathcal{L}(\hat{\mathcal{M}}) - \alpha\mathcal{L}(\mathcal{M}^*)\\
&= \big[\mathcal{L}(\hat{\mathcal{M}}) - \widehat{\mathcal{L}}(\hat{\mathcal{M}})\big] + \big[\widehat{\mathcal{L}}(\hat{\mathcal{M}}) - \alpha\widehat{\mathcal{L}}(\mathcal{M}^*_{\mathcal{S}})\big]\\
&\quad + \alpha\big[\widehat{\mathcal{L}}(\mathcal{M}^*_{\mathcal{S}}) - \widehat{\mathcal{L}}(\mathcal{M}^*)\big] + \alpha\big[\widehat{\mathcal{L}}(\mathcal{M}^*) - \mathcal{L}(\mathcal{M}^*)\big]\\
&\le \widehat{\mathcal{L}}(\hat{\mathcal{M}}) - \alpha\widehat{\mathcal{L}}(\mathcal{M}^*_{\mathcal{S}}) + (1+\alpha)L_g\epsilon + \alpha L_\phi\epsilon\\
&\le \widehat{\mathcal{L}}(\hat{\mathcal{M}}) - \alpha\widehat{\mathcal{L}}(\mathcal{M}^*_{\mathcal{S}}) + (1+2\alpha)L_g\epsilon,
\end{aligned}$$

using $L_\phi \le L_g$. Combining with Steps 2 and 3,

$$\mathcal{L}(\hat{\mathcal{M}}) - \alpha\mathcal{L}(\mathcal{M}^*) \le (\alpha+1)\sqrt{\frac{2m\ln(6/\delta)}{n_2}} + (1+2\alpha)L_g\frac{2\sqrt{d_e}}{m^{1/d_e}} + 2\sqrt{2}\,\alpha(B+c_\lambda)\left(\frac{2\lambda(2\epsilon/\ell_k)^p\,C_v}{n_2}\right)^{\frac{1}{p+2}},$$

which completes the proof. □

*Remark* 1. To determine the optimal value of the discretization radius $\epsilon^*$, we analyze the asymptotic behavior of the suboptimality gap bound derived in the offline setting. First we substitute the covering number relation $m = O(\epsilon^{-d_e})$ into the three main error terms of the bound. The term due to discretization error scales linearly with $\epsilon$ so it is $O(\epsilon)$. The generalization gap term arising from the weight concentration scales as $O\left(\sqrt{m/n_2}\right) = O(\epsilon^{-d_e/2}n_2^{-1/2})$ and the cost estimation gap arising from the estimation of the cost function scales as $O((\epsilon^{p-d_e}/n_2)^{1/(p+2)})$ when $v_i\underline{v}_i \ge v_0$ so $C_v \le m/v_0$.

Arrival and cost both balance against discretization at the same point $\epsilon^* = n_2^{-1/(d_e+2)}$. All three terms are $O(n_2^{-1/(d_e+2)})$ so it follows that $\mathrm{SubOpt}(\hat{\mathcal{M}}) = \tilde{O}(n_2^{-1/(d_e+2)})$.

Proof of Lemma 5.3. At any round $t \in [T]$, the algorithm maintains a dynamically growing representative set $\mathcal{S}_t$. By the definition of an $\epsilon$-net over a compact metric space $\mathcal{X}$ with diameter $D$ and intrinsic dimension $p$, the maximum number of $\epsilon$-balls required to cover the space is strictly finite. Therefore, the number of centers $m_t = |\mathcal{S}_t|$ is monotonically non-decreasing but upper-bounded by the maximum covering number $m_{\max}$ for all $t$:

$$m_t \le m_{\max} \le \left(\frac{D\sqrt{d_e}}{\epsilon}\right)^{d_e}. \tag{6}$$

According to the `CLCB-SC-LS-Cont` algorithm, a new stage $\tau_u \to \tau_u + 1$ is triggered for a specific center $u \in \mathcal{S}_t$ if and only if the number of new observations in the current stage satisfies:

$$|\mathcal{T}(u,\tau_u)| \ge 1 + \sqrt{\frac{T}{m_t}\sum_{\tau=1}^{\tau_u-1}|\mathcal{T}(u,\tau)|}. \tag{7}$$

Because the denominator $m_t$ is bounded such that $m_t \le m_{\max}$, the ratio $\frac{T}{m_t}$ is strictly bounded below by $\frac{T}{m_{\max}}$. We can therefore establish a strict necessary condition for a cache switch:

$$|\mathcal{T}(u,\tau_u)| \ge 1 + \sqrt{\frac{T}{m_{\max}}\sum_{\tau=1}^{\tau_u-1}|\mathcal{T}(u,\tau)|}. \tag{8}$$

At this point, the continuous trigger condition is mathematically identical to the static trigger condition analyzed in the discrete `CLCB-SC-LS` algorithm in [18], with the dynamic variable $m_t$ replaced by the constant $m_{\max}$.

Because the recurrence relation is now structurally identical, the remainder of the proof follows identically to the proof of the discrete low-switching lemma.

For the global arrival probability stages $\tau_p$, the update rule relies on a single global tracker, making it equivalent to the discrete proof [18] where the number of arms is 1, yielding $\mathbb{E}[\tau_p(T)] \le O(\log\log T)$. □

Proof of Theorem 5.1. By Assumption 2, the continuous queries are highly concentrated. The maximum distance between any query $x$ generated from these clusters and its nearest empirical center in $\mathcal{S}_t$ is bounded by a dynamic radius $\varepsilon_t = O(1/\sqrt{t})$. Integrating this over the probability mass of the clusters, the instantaneous discretization error satisfies $|\mathcal{L}(\mathcal{M}) - \widehat{\mathcal{L}}(\mathcal{M})|_t \le L_g\varepsilon_t$.

Next, we bound the regret against the optimal observed cache $\mathcal{M}^*_{\mathcal{N}_T}$. Let $\mathcal{M}^*_{\mathcal{S}_t}$ be the optimal cache restricted to the centers in $\mathcal{S}_t$. By mapping each item in $\mathcal{M}^*_{\mathcal{N}_T}$ to its nearest neighbor in $\mathcal{S}_t$, the mismatch cost increases by at most $L_\phi\varepsilon_t$. Because the loss function evaluates the minimum over the cache, this error does not scale with $k$ and applies only when the cache is hit. The instantaneous comparator gap is bounded by $L_g\varepsilon_t$. Therefore, at any round $t$:

$$\widehat{\mathcal{L}}(\mathcal{M}^*_{\mathcal{S}_t}) \le \widehat{\mathcal{L}}(\mathcal{M}^*_{\mathcal{N}_T}) + L_g\varepsilon_t \tag{9}$$

The expected cumulative regret decomposes into the discrete proxy regret, the dynamic discretization penalties, and the switching cost:

$$\mathrm{Reg}(T) \le \mathbb{E}\left[\sum_{t=1}^{T}\left(\widehat{\mathcal{L}}(\mathcal{M}_t;\mathbf{w},c) - \alpha\widehat{\mathcal{L}}(\mathcal{M}^*_{\mathcal{S}_t};\mathbf{w},c)\right)\right] + \sum_{t=1}^{T}(1+\alpha)L_g\varepsilon_t + \sum_{t=1}^{T}\alpha L_g\varepsilon_t + \mathbb{E}[C_{switch}]$$

Since $\varepsilon_t = O(1/\sqrt{t})$, summing these penalties over $T$ yields $\sum_{t=1}^{T}\frac{L_g}{\sqrt{t}} \le 2L_g\sqrt{T} = O(\sqrt{T})$.

We condition on the high-probability event $\mathcal{E} = \mathcal{E}_{\text{arr}} \cap \mathcal{E}_{\text{cost}}$ from Lemma 5.2, which holds with probability at least $1 - 1/T$. Adding and subtracting intermediate proxy losses evaluated on the estimated weights $\hat{\boldsymbol{w}}_t$ and LCB costs $\underline{c}_{t-1}$ yields:

$$\begin{aligned}
&\widehat{\mathcal{L}}(\mathcal{M}_t;\boldsymbol{w},c) - \alpha\widehat{\mathcal{L}}(\mathcal{M}^*_{\mathcal{S}_t};\boldsymbol{w},c)\\
&= \left[\widehat{\mathcal{L}}(\mathcal{M}_t;\boldsymbol{w},c) - \widehat{\mathcal{L}}(\mathcal{M}_t;\boldsymbol{w},\underline{c}_{t-1})\right] \quad \text{(Optimistic Gap)}\\
&\quad + \left[\widehat{\mathcal{L}}(\mathcal{M}_t;\boldsymbol{w},\underline{c}_{t-1}) - \widehat{\mathcal{L}}(\mathcal{M}_t;\hat{\boldsymbol{w}}_t,\underline{c}_{t-1})\right] \quad \text{(Arrival Gap 1)}\\
&\quad + \left[\widehat{\mathcal{L}}(\mathcal{M}_t;\hat{\boldsymbol{w}}_t,\underline{c}_{t-1}) - \alpha\widehat{\mathcal{L}}(\mathcal{M}^*_{\mathcal{S}_t};\hat{\boldsymbol{w}}_t,\underline{c}_{t-1})\right] \quad \text{(Greedy Gap)}\\
&\quad + \alpha\left[\widehat{\mathcal{L}}(\mathcal{M}^*_{\mathcal{S}_t};\hat{\boldsymbol{w}}_t,\underline{c}_{t-1}) - \widehat{\mathcal{L}}(\mathcal{M}^*_{\mathcal{S}_t};\hat{\boldsymbol{w}}_t,c)\right] \quad \text{(LCB Gap)}\\
&\quad + \alpha\left[\widehat{\mathcal{L}}(\mathcal{M}^*_{\mathcal{S}_t};\hat{\boldsymbol{w}}_t,c) - \widehat{\mathcal{L}}(\mathcal{M}^*_{\mathcal{S}_t};\boldsymbol{w},c)\right] \quad \text{(Arrival Gap 2)}
\end{aligned}$$

By the approximation guarantee of `Cont_Reverse_Greedy`, the Greedy Gap is $\le 0$. Under $\mathcal{E}_{\text{cost}}$, $\underline{c}_{t-1}(u) \le c(u)$, making the LCB Gap $\le 0$. By the weight sensitivity bound (Lemma 4 in [18]), the Arrival Gaps are bounded by $(1+\alpha)\|\hat{\boldsymbol{w}}_t - \boldsymbol{w}\|_1$. By the sensitivity of optimistic costs (Lemma 6 from [18]), the Optimistic Gap is bounded by $\sum_{i:\underline{c}_{t-1}(u_i)\le\phi(d(u_i,\mathcal{M}_t))} w_i(c(u_i) - \underline{c}_{t-1}(u_i))$.

Under Assumption 2, the number of active centers is bounded by $m_{\text{eff}}$. Summing the bound provided by $\mathcal{E}_{\text{arr}}$ over $T$ rounds:

$$\mathbb{E}\left[\sum_{t=1}^{T}(1+\alpha)\|\hat{\boldsymbol{w}}_t - \boldsymbol{w}\|_1\right] \le (1+\alpha)\sum_{t=1}^{T}\sqrt{\frac{2m_{\text{eff}}\ln(2T^3)}{t}} \le O\left(\sqrt{m_{\text{eff}}T\log T}\right)$$

For the Optimistic Gap, under $\mathcal{E}$, $c(u_i) - \underline{c}_{t-1}(u_i) \le 2\beta_{t-1}\sigma_{t-1,\lambda}(u_i)$. Let $\mathbb{I}_{i,t}$ indicate that cell $i$ is chosen and the LLM is queried. Given the history $\mathcal{F}_{t-1}$, $\mathbb{E}[\mathbb{I}_{i,t} \mid \mathcal{F}_{t-1}] = w_i\mathbb{I}\{\underline{c}_{t-1}(u_i) \le \phi(d(u_i,\mathcal{M}_t))\}$. By the triggering probability equivalence trick, the expected sum maps to the $\ell_T$ queries where the LLM was actually invoked ($z_j$):

$$\mathbb{E}\left[\sum_{t=1}^{T}\text{Optimistic Gap}\right] \le \mathbb{E}\left[\sum_{j=1}^{\ell_T}2\beta_{t_j-1}\sigma_{j-1,\lambda}(z_j)\right] \tag{10}$$

Since $\beta_t \le \beta_T$, applying the Cauchy-Schwarz inequality and the Elliptical Potential Lemma for RKHS yields:

$$\sum_{j=1}^{\ell_T}\sigma_{j-1,\lambda}(z_j) \le \sqrt{\ell_T\sum_{j=1}^{\ell_T}\sigma^2_{j-1,\lambda}(z_j)} \le \sqrt{T\frac{2}{\ln(1+1/\lambda)}\gamma_T} = O\left(\sqrt{T\gamma_T}\right)$$

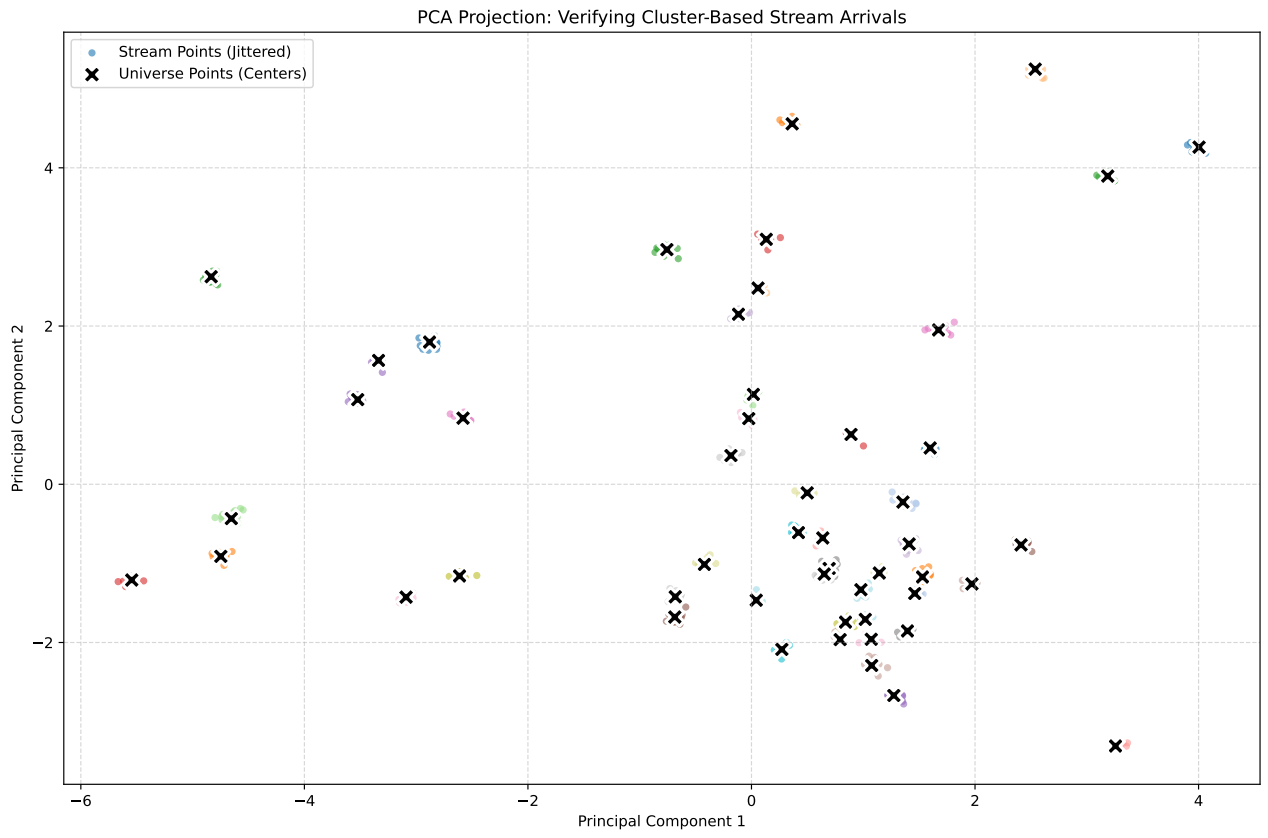

**Figure 3: PCA on the synthetic dataset**

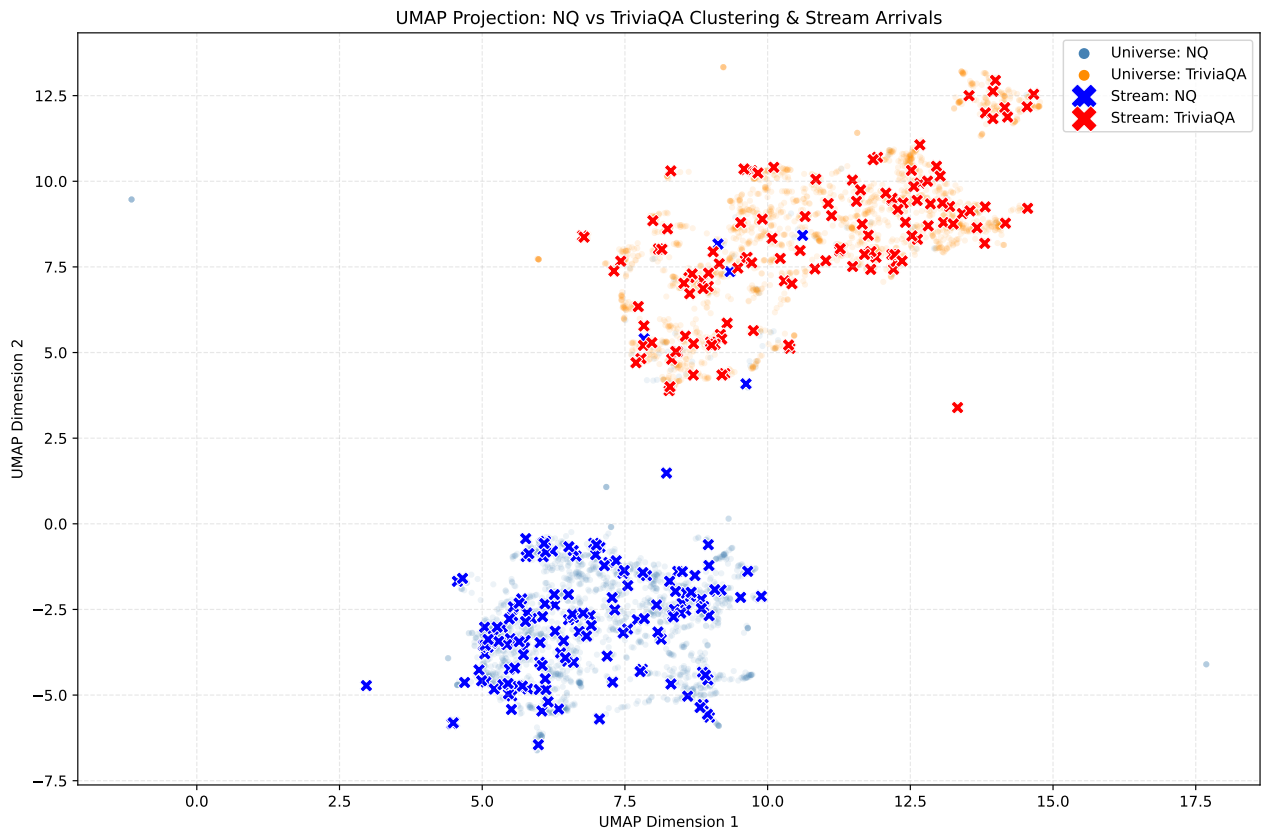

**Figure 4: UMAP on the NQ+TriviaQA dataset**

By Lemma 5.3, maintaining the effective centers triggers at most $O(m_{\max}\log\log T)$ cache switches. At a maximum cost of 1 for $k$ items, $\mathbb{E}[C_{switch}] \le O(km_{\max}\log\log T)$.

If the high-probability event $\mathcal{E}$ fails (probability $\le 1/T$), the maximum regret per round is bounded by 1, adding a constant penalty of $(1/T)\times T = 1$. Combining the Arrival Gap, Optimistic Gap, switching costs, and the dynamically integrated $O(\sqrt{T})$ discretization penalties from Step 1 yields the final stated bound. □

## A.2 `CLCB-Frozen-Cont` algorithm

The `CLCB-Frozen-Cont` algorithm is designed to optimize continuous semantic caching by operating in discrete, stable stages to balance learning query costs with exploiting the cache. Algorithm 6 processes incoming queries and relies on Algorithm 8 to track empirical query frequencies against a fixed *frozen* candidate pool. This subroutine maps incoming queries to their nearest neighbors within a membership radius ($\rho$) (Algorithm 8, lines 2-6) and computes confidence bounds on these weights. It triggers a cache update (SWITCH) only when the uncertainty of the empirical estimates falls below exponentially decaying stage tolerances (Algorithm 8, lines 15-16), which limits the number of updates. Here $\beta_t = B + R\sqrt{2(\gamma_t + \ln(1/\delta_t))}$ and $\gamma_t := \frac{1}{2}\log\det(I + K_t/\lambda)$ is the standard information gain. When a switch occurs, the algorithm updates the frozen pool and determines the new optimal cache subset of size $k$. It does this by utilizing KRR to estimate the LCB of the costs and feeding those into CONT_REVERSE_GREEDY. Finally for the actual routing of each query Algorithm 7 makes an optimistic online decision: it queries the LLM if the LCB of the cost is lower than the mismatch penalty of utilizing the nearest cached item, otherwise, it serves the response directly from the cache.

*A.2.1 Theoretical Guarantees for* `CLCB-Frozen-Cont`.

LEMMA A.4 (STAGE-FROZEN ARRIVAL-WEIGHT CONCENTRATION). *Fix a stage $s$ and let its start time be $t_s$. Let the* frozen *candidate pool be $\bar{U}_s$ and define the stage alphabet*

$$\mathcal{A}_s \triangleq \bar{U}_s \cup \{\perp\}, \qquad m_s \triangleq |\mathcal{A}_s| = |\bar{U}_s| + 1.$$

*For each round $t \geq t_s$ in stage $s$, define the (measurable) symbol assignment $z_t \in \mathcal{A}_s$ by*

$$z_t = \begin{cases} \arg\min_{u\in\bar{U}_s} d(x_t,u), & \text{if } \min_{u\in\bar{U}_s} d(x_t,u) \leq \rho, \\ \perp, & \text{otherwise,} \end{cases}$$

*and assume $\{z_t\}_{t=t_s}^{t_{s+1}-1}$ are i.i.d. with distribution $w_s^\star \in \Delta(\mathcal{A}_s)$. That is $Pr(z_t = a) = w_s^*(a)$ for all $a \in A_s$ and all $t \in [t_s, t_{s+1})$. Let $n \triangleq t - t_s + 1$ be the number of arrivals observed in stage $s$ up to time $t$, and define the empirical distribution*

$$\hat{w}_{s,t}(a) \triangleq \frac{1}{n}\sum_{\tau=t_s}^{t} \mathbf{1}\{z_\tau = a\}, \qquad a \in \mathcal{A}_s.$$

*Then, assume the algorithm uses a confidence schedule $\delta_t = O(1/t^2)$, then there exists a constant $C > 0$ such that with probability at least $1 - O(\delta)$, for all stages $s$ and for all times $t \geq t_s$ :*

$$\left\|\hat{w}_{s,t} - w_s^\star\right\|_1 \leq C\sqrt{\frac{m_s \log 2 - \log(\delta) + \log t}{t - t_s + 1}}.$$

PROOF. For any fixed $n \geq 1$ and $\varepsilon > 0$, a classical bound for empirical distributions on a finite alphabet (from Theorem 2.1 in [29]) states that for i.i.d. samples on an alphabet of size $m_s$,

$$\Pr\Big(\left\|\hat{w}_{s,t} - w_s^\star\right\|_1 \geq \varepsilon\Big) \leq 2^{m_s}\exp\Big(-\frac{n\varepsilon^2}{2}\Big). \tag{11}$$

where $n = t - t_s + 1$ (Here $z_{t_s}, \ldots, z_{t_s+n-1}$ are i.i.d. on $\mathcal{A}_s$, so (11) applies directly.)

Solving for $\epsilon$, this implies that for any failure probability $\Upsilon > 0$,

$$\left\|\hat{w}_{s,t} - w_s^\star\right\|_1 \leq \sqrt{\frac{2\big(m_s \log 2 + \log(1/\Upsilon)\big)}{n}}.$$

Now to have the bound hold simultaneously for all times $t \geq t_s$ within a stage, we allocate a time-dependent failure budget:

$$\Upsilon_t = \frac{\delta}{2t^2}$$

a union bound over all $t \geq t_s$ implies that with probability at least $1 - O(\delta)$ the inequality above holds for all $t \geq t_s$ simultaneously with $\Upsilon = \Upsilon_t$. Substituting $\Upsilon_t$ into the bound yields

$$\left\|\hat{w}_{s,t} - w_s^\star\right\|_1 \leq \sqrt{\frac{2\big(m_s \log 2 + \log(2t^2/\delta)\big)}{t - t_s + 1}}.$$

using $\log(t^2) = 2\log t$ and absorbing constants gives the bound. Finally we can apply a union bound across stages. Since the algorithm uses a low-switching schedule, the number of stages up to time $T$ is at most logarithmic in $T$. Alternatively, assigning a summable stage-level failure budget such as $\delta_t := \delta/(2s^2)$ gives that with probability at least $1 - O(\delta)$ the stated inequality holds for all stages $s$ and all times $t$. □

*Assumption* 3 (Random Cover). Let $\bar{U}_s$ denote a frozen pool of $n_s^{\text{total}}$ unique queries drawn i.i.d from $f$. By standard $\epsilon-$net arguments for metric spaces, there exists a universal constant $C_{cov} > 0$ such that, with high probability, the outlier probability mass $p_{\perp,s}-$defined as the probability of drawing a query outside the union of balls of radius $\rho$ centered at the points in $\bar{U}_s-$is bounded by:

$$p_{\perp,s} \leq \tilde{O}\left(C_{cov}\frac{(1/\rho)^p}{n_s^{\text{total}}}\right)$$

LEMMA A.5 (TOLERANCE-BOUNDED QUANTIZATION ERROR). *Let $\mathcal{L}(\mathcal{M}) = \mathbb{E}_{x\sim f}[g_{\mathcal{M}}(x)]$ denote the true expected continuous loss where and let $\mathcal{L}_s(\mathcal{M}) = \mathbb{E}_{z\sim w_s^*}[g_{\mathcal{M}}(z)]$ denote the empirical proxy loss evaluated over the frozen pool $\bar{U}_s$, where the loss function $g$ is $L_g-$Lipschitz as established before and bounded by 1. Assume the algorithm defines a dynamic membership radius parameterized by the stage tolerance $e_s > 0$:*

$$\rho_s \triangleq \frac{e_s}{2L_g}$$

*Furthermore, assume that at the beginning of stage $s$, the total number of unique historical queries $n_s^{total} = |\bar{U}_s|$ satisfies the following lower bound:*

$$n_s^{total} \geq \tilde{O}\left(\frac{2C_{cov}(2L_g)^p}{e_s^{p+1}}\right)$$

*Then, with high probability, the quantization bias for any cache $\mathcal{M}$ is strictly upper bounded by the stage tolerance:*

$$\sup_{\mathcal{M}} |\mathcal{L}(\mathcal{M}) - \mathcal{L}_s(\mathcal{M})| \leq e_s$$

PROOF. We begin by establishing a uniform upper bound on the quantization bias. By definition, the deviation between the continuous expected loss and the empirical proxy loss can be bounded by moving the absolute value inside the expectation:

$$\begin{aligned}|\mathcal{L}(\mathcal{M}) - \mathcal{L}_s(\mathcal{M})| &= \Big|\mathbb{E}_{x\sim f}[\ell(x;\mathcal{M})] - \mathbb{E}_{z\sim w_s^\star}[\ell(z;\mathcal{M})]\Big| \\ &\leq \mathbb{E}_{x\sim f}\left[|\ell(x;\mathcal{M}) - \ell(z(x);\mathcal{M})|\right]\end{aligned}$$

By applying the Law of Total Expectation, we partition the integration space into an anchored region where the queries are mapped to a nearest neighbor within $\rho_s$, bounded by the $L_g-$Lipschitz property and an outlier region where the queries are unmapped, denoted by $\perp$ and bounded by 1. Taking the supremum over all caches $\mathcal{M}$ yields:

$$\sup_{\mathcal{M}} |\mathcal{L}(\mathcal{M}) - L_s(\mathcal{M})| \leq L_g\rho_s \cdot (1 - p_{\perp,s}) + 1 \cdot p_{\perp,s} \leq L_g\rho_s + p_{\perp,s} \tag{12}$$

Now we bound each of these two additive terms by $e_s/2$.

By the definition of our dynamic radius schedule, we substitute $\rho_s = \frac{e_s}{2L_g}$ directly into the anchored error term, yielding:

$$L_g \rho_s = L_g \left( \frac{e_s}{2L_g} \right) = \frac{e_s}{2}$$

To bound the outlier error term, using Assumption 3, the outlier probability $p_{\perp,s}$ for the frozen pool $\bar{U}_s$ evaluated at the scheduled radius $\rho_s$ is given by:

$$p_{\perp,s} \le \tilde{O}\left( C_{cov} \frac{(1/\rho)^p}{n_s^{\text{total}}} \right) = \tilde{O}\left( C_{cov} \frac{(2L_g/e_s)^p}{n_s^{\text{total}}} \right)$$

To ensure this error satisfies $p_{\perp,s} \le \frac{e_s}{2}$, we impose the following inequality:

$$\tilde{O}\left( C_{cov} \frac{(2L_g/e_s)^p}{n_s^{\text{total}}} \right) \le \frac{e_s}{2}$$

Rearranging this inequality to solve for the threshold of the pool size $n_s^{\text{total}}$, we obtain:

$$n_s^{\text{total}} \ge \tilde{O}\left( \frac{2C_{cov}(2L_g)^p}{e_s^{p+1}} \right)$$

By the premise of the lemma, $n_s^{\text{total}}$ explicitly satisfies this threshold at the start of stage $s$. Hence, the condition $p_{\perp,s} \le \frac{e_s}{2}$ holds with high probability. Therefore

$$L_g \rho_s + p_{\perp,s} \le e_s$$

□

Lemma A.6 (Low-Switching from Decaying Uncertainty). *Let the stage tolerance be defined by the schedule $e_s = 2^{-2^s}$ for $s \ge 1$. A stage $s$ ends at the first time step $t$ such that the joint uncertainty envelope satisfies $U_t \le e_s$, where $U_t = \max\{U_t^{(p)}, U_t^{(c)}\}$. Assume the algorithm's pool and continuous uncertainties satisfy the following upper bounds for some constant $j \ge 1$:*

$$U_t^{(p)} \le C_p \sqrt{\frac{\log T}{t - t_s + 1}}, \quad \text{and} \quad U_t^{(c)} \le C_c \frac{(\log T)^j}{\sqrt{t}}$$

*where $t_s$ is the start time of stage $s$. Then the total number of stage switches up to time $T$, denoted $S(T)$, is bounded by $O(\log \log T)$.*

Proof. For the algorithm to trigger a switch and end stage $s$, the joint uncertainty must satisfy $U_t \le e_s$. Because $U_t = \max\{U_t^{(p)}, U_t^{(c)}\}$, this strict threshold requires both conditions to hold simultaneously:

$$U_t^{(p)} \le e_s \quad \text{and} \quad U_t^{(c)} \le e_s$$

Let $n_s$ denote the total number of rounds spent in stage $s$. The uncertainty condition requires that $C_p \sqrt{\frac{\log T}{t - t_s + 1}} \le e_s$. Since $t - t_s + 1$ measures the time elapsed within the current stage, this implies the stage cannot end until at least:

$$n_s \ge \frac{C_p^2 \log T}{e_s^2}$$

While the continuous uncertainty $U_t^{(c)}$ may require the absolute time $t$ to be large, the stage duration $n_s$ is strictly lower-bounded by the time required for the pool uncertainty to decay. If $U_t^{(c)}$ decays slower than $U_t^{(p)}$, it only forces the stage to last longer, which would further decrease the total number of switches. Therefore, taking the minimum required duration is a valid conservative bound.

Substituting the tolerance schedule $e_s = 2^{-2^s}$, we get $e_s^2 = 2^{-2^{s+1}}$, yields the lower bound:

$$n_s \ge C_p^2 (\log T) 2^{2^{s+1}}$$

Let $S$ be the total number of stages initiated up to time $T$. The total time horizon $T$ is the sum of the lengths of all completed stages, plus the final incomplete stage. Therefore, $T$ is strictly lower bounded by the duration of the second-to-last stage, $n_{S-1}$:

$$T = \sum_{s=1}^{S} n_s \ge n_{S-1} \ge C_p^2 (\log T) 2^{2^S}$$

Rearranging to isolate $S$:

$$2^{2^S} \le \frac{T}{C_p^2 \log T} \le T$$

Taking the logarithm of both sides yields $2^S \le \log_2 T$. Taking the logarithm a second time gives:

$$S \le \log_2 \log_2 T$$

Hence, the total number of switches by time $T$ is bounded by $O(\log \log T)$. □

Theorem A.7 (Regret Bound for CLCB-Frozen-Cont). *Suppose the algorithm is run for $T$ rounds with a cache size $k$ and a maximum item cost $c_{\max} \le 1$. Let the stage tolerance follow the schedule $e_s = 2^{-2^s}$. Assume the pool uncertainty decays as $U_t^{(p)} = O(\sqrt{(\log T)/n})$ where $n$ is the time elapsed in the current stage, and the continuous cost function satisfies a kernel growth condition with information gain $\gamma_t \le C_\gamma (\log(1 + t/\lambda))^j$ for some $j \ge 1$. Let $\alpha$ be the approximation ratio of the Continuous Reverse Greedy subroutine, and $\mathcal{M}_t^*$ denote the optimal cache in hindsight at time $t$.*

*Then, with high probability, the cumulative $\alpha$-regret of the algorithm according to the definition in section 5.1 is:*

$$Regret(T) = \tilde{O}\left( \sqrt{T} \right)$$

Proof. We prove the claimed bound by (i) controlling per round $\alpha-$suboptimality within each stage via the stage envelope, and (ii) summing the resulting stage regrets. Fix a stage index $s$ and consider any round $t$ that belongs to stage $s$, by construction of the low-switching scheme, the cache is constant within the stage $\mathcal{M}_t = \mathcal{M}_s$ for all such $t$. By definition of stage $s$, we have for all $t$ in stage $s$,

$$U_t^{(c)} \le e_s, \quad U_t^{(p)} \le e_s \quad \Rightarrow \quad U_t \le e_s \qquad \text{(StageBound)}$$

where $U_t \triangleq \max\{U_t^{(p)}, U_t^{(c)}\}$. $\widehat{\mathcal{L}}(\mathcal{M}; \mathcal{U}, \boldsymbol{w}, c)$ is the discretized loss evaluates on the candidate pool $\mathcal{U}$ with weights $\boldsymbol{w}$ and costs $c$. Letting $\boldsymbol{w}^*$ denote the true arrival weights and $\hat{\boldsymbol{w}}_t$ denote the estimated weights used at time $t$. By Lemma 4 in [18], we have that for any cache $\mathcal{M}$ and any cost $c$:

$$|\widehat{\mathcal{L}}(\mathcal{M}; \boldsymbol{w}^*, c) - \widehat{\mathcal{L}}(\mathcal{M}; \hat{\boldsymbol{w}}_t, c)| \le \|\boldsymbol{w}^* - \hat{\boldsymbol{w}}_t\|_1 \le U_t^{(p)} \le e_s \qquad \text{(S1)}$$

Defining $\hat{c}_{t-1}$ as in Algorithm 4 and working under the high probability event on which for all $t$ and all $u \in \mathcal{U}_t$ and defining $c_{t-1}^{\text{LCB}}(u) \triangleq$

$\hat{c}_{t-1}(u) - \beta_t \sigma_{t-1}(u)$ for all $u \in \mathcal{U}$ then:

$$|c(u) - \hat{c}_{t-1}(u)| \le \beta_t \sigma_{t-1}(u)$$

On this event, we have $c_{t-1}^{\text{LCB}}(u) \le c(u)$ and

$$0 \le c(u) - c_{t-1}^{\text{LCB}}(u) = c(u) - \hat{c}_{t-1}(u) + \beta_t \sigma_{t-1}(u) \le 2\beta_t \sigma_{t-1}(u)$$

Using Lemma 4 from [18] again, we get

$$\begin{aligned} |\widehat{\mathcal{L}}(\mathcal{M}; \hat{\boldsymbol{w}}_t, c) - \widehat{\mathcal{L}}(\mathcal{M}; \hat{\boldsymbol{w}}_t, c_{t-1}^{\text{LCB}})| &\le \sum_{u \in \mathcal{U}} \hat{w}_t(u) \left| c(u) - c_{t-1}^{\text{LCB}}(u) \right| \\ &= \sum_{u \in \mathcal{U}} \hat{w}_t(u) \left( c(u) - c_{t-1}^{\text{LCB}}(u) \right) \\ &\le 2\beta_t \sum_{u \in \mathcal{U}} \hat{w}_t(u) \sigma_{t-1}(u) \;\triangleq\; U_t^c \;\le\; e_s. \end{aligned} \tag{S2}$$

Combining (S1)-(S2) and applying the triangle inequality yields for any cache $\mathcal{M}$

$$\widehat{\mathcal{L}}(\mathcal{M}, \boldsymbol{w}^*, c) \le \widehat{\mathcal{L}}(\mathcal{M}, \hat{\boldsymbol{w}}_t, c_{t-1}^{\text{LCB}}) + 2e_s$$

$$\widehat{\mathcal{L}}(\mathcal{M}, \hat{\boldsymbol{w}}_t, c_{t-1}^{\text{LCB}}) \le \widehat{\mathcal{L}}(\mathcal{M}, \boldsymbol{w}^*, c) + 2e_s$$

At the beginning of stage $s$, the algorithm computes $\mathcal{M}_s$ by applying Continuous Reverse Greedy to the estimated problem with $(\hat{\boldsymbol{w}}_t, c_{t-1}^{\text{LCB}})$, hence by the $\alpha-$approximation guarantee,

$$\widehat{\mathcal{L}}(\mathcal{M}_s, \hat{\boldsymbol{w}}_t, c_{t-1}^{\text{LCB}}) \le \alpha \min_{|\mathcal{M}| \le k} \widehat{\mathcal{L}}(\mathcal{M}; \hat{\boldsymbol{w}}_t, c_{t-1}^{\text{LCB}}) \tag{RG}$$

Let $\mathcal{M}_t^*$ denote the optimal cache for the true objective (from the queries observed up to time $t$). Since $\mathcal{M}^*$ is feasible for the minimization in (RG),

$$\widehat{\mathcal{L}}(\mathcal{M}_s, \hat{\boldsymbol{w}}_t, c_{t-1}^{\text{LCB}}) \le \alpha \widehat{\mathcal{L}}(\mathcal{M}^*, \hat{\boldsymbol{w}}_t, c_{t-1}^{\text{LCB}}) \tag{RG2}$$

Since the set we use to compute $\mathcal{M}_t^*$ is smaller than that of $\mathcal{M}^*$ we have that $\widehat{\mathcal{L}}(\mathcal{M}^*; \boldsymbol{w}^*, c) \le \widehat{\mathcal{L}}(\mathcal{M}_t^*; \boldsymbol{w}^*, c)$. Hence

$$\begin{aligned} \widehat{\mathcal{L}}(\mathcal{M}^*, \hat{\boldsymbol{w}}_t, c_{t-1}^{\text{LCB}}) \le \widehat{\mathcal{L}}(\mathcal{M}^*, \boldsymbol{w}^*, c) + 2e_s &\le \widehat{\mathcal{L}}(\mathcal{M}_t^*, \boldsymbol{w}^*, c) + 2e_s \\ &\le \widehat{\mathcal{L}}(\mathcal{M}_t^*, \hat{\boldsymbol{w}}_t, c_{t-1}^{\text{LCB}}) + 4e_s \end{aligned}$$

Now combine this result with (RG2) to get:

$$\widehat{\mathcal{L}}(\mathcal{M}_s; \hat{\boldsymbol{w}}_t, c_{t-1}^{\text{LCB}}) \le \alpha(\widehat{\mathcal{L}}(\mathcal{M}_t^*, \hat{\boldsymbol{w}}_t, c_{t-1}^{\text{LCB}}) + 4e_s)$$

Using this result we obtain

$$\widehat{\mathcal{L}}(\mathcal{M}_s, \boldsymbol{w}^*, c) \le \alpha \widehat{\mathcal{L}}(\mathcal{M}_t^*, \boldsymbol{w}^*, c) + (6\alpha + 2)e_s \tag{StageOpt}$$

Recall from Lemma A.5 that $\mathcal{L}_s(\mathcal{M}) = \mathbb{E}_{z \sim w_s^*}[g_{\mathcal{M}}(z)]$ represents the empirical proxy loss on the frozen pool $\bar{U}_s$ under true weights. This is notationally equivalent to our pool evaluated loss $\widehat{\mathcal{L}}(\mathcal{M}; \boldsymbol{w}^*, c)$. Therefore we can rewrite StageOpt in terms of the proxy loss as

$$\mathcal{L}_s(\mathcal{M}_s) \le \alpha \mathcal{L}_s(\mathcal{M}_t^*) + (6\alpha + 2)e_s$$

We now bound the instantaneous $\alpha$-suboptimality gap on the true continuous loss $\mathcal{L}$. By adding and subtracting the proxy losses $\mathcal{L}_s$, we decompose the true gap into the empirical proxy gap and the quantization biases for both the chosen cache and the optimal cache:

$$\begin{aligned} \mathcal{L}(\mathcal{M}_s) - \alpha \mathcal{L}(\mathcal{M}_t^*) = &\underbrace{(\mathcal{L}(\mathcal{M}_s) - \mathcal{L}_s(\mathcal{M}_s))}_{\text{Quantization Bias } (\mathcal{M}_s)} \\ &+ \underbrace{(\mathcal{L}_s(\mathcal{M}_s) - \alpha \mathcal{L}_s(\mathcal{M}_t^*))}_{\text{Empirical Proxy Gap}} \\ &+ \underbrace{\alpha(\mathcal{L}_s(\mathcal{M}_t^*) - \mathcal{L}(\mathcal{M}_t^*))}_{\text{Quantization Bias } (\mathcal{M}_t^*)} \end{aligned}$$

From our rewritten (StageOpt), the Empirical Proxy Gap is bounded by $(6\alpha + 2)e_s$. By the Tolerance-Bounded Quantization Error lemma, we have $\sup_{\mathcal{M}} |\mathcal{L}(\mathcal{M}) - \mathcal{L}_s(\mathcal{M})| \le e_s$. Applying this uniform bound to both $\mathcal{M}_s$ and $\mathcal{M}_t^*$ yields:

$$\mathcal{L}(\mathcal{M}_s) - \mathcal{L}_s(\mathcal{M}_s) \le e_s$$

$$\alpha(\mathcal{L}_s(\mathcal{M}_t^*) - \mathcal{L}(\mathcal{M}_t^*)) \le \alpha e_s$$

Summing these components, the per-round true $\alpha$-regret is bounded by:

$$\mathcal{L}(\mathcal{M}_s) - \alpha \mathcal{L}(\mathcal{M}_t^*) \le e_s + (6\alpha + 2)e_s + \alpha e_s = (7\alpha + 3)e_s$$

This yields the per-round $\alpha$-gap bound on the true continuous loss. Defining $C_0 \triangleq 7\alpha + 3$, we now sum across all rounds $t$ in stage $s$. Letting stage $s$ contain $n_s$ rounds, we obtain:

$$\sum_{t \in \text{stage } s} (\mathcal{L}(\mathcal{M}_s) - \alpha \mathcal{L}(\mathcal{M}_t^*)) \le C_0 n_s e_s \tag{StageReg}$$

Now it remains to upper bound $n_s$ as a function of $e_s$. By definition, stage $s$ ends at the first time $t$ such that that joint envelope satisfies $U_t \le e_s$. We know from Lemma A.4 that with high probability

$$U_t^{(p)} \le C_p \sqrt{\frac{\log T}{t - t_s + 1}} \quad \forall t \in [t_s, t_{s+1})$$

To end stage $s$, it suffices that $U_t^{(p)} \le e_s$. A sufficient condition then is

$$U_t^{(p)} \le C_p \sqrt{\frac{\log T}{t - t_s + 1}} \le e_s \quad \Rightarrow n \ge n_s^{(p)} := \frac{C_p^2 \log T}{e_s^2}$$

which implies the bound $n_s \le n_s^{(p)}$. We also require $U_t^{(c)} \le e_s$ to end the stage. Next, we prove a bound on $U_t^{(c)}$.

Let $x_1, \ldots, x_t$ be the (distinct) contexts at which the learner queried the LLM and observed costs, and let $K_t \in \mathbb{R}^{t \times t}$ be the Gram matrix with entries $[K_t]_{ij} = k(x_i, x_j)$. Define the standard information gain

$$\gamma_t := \frac{1}{2} \log \det(I + K_t / \lambda).$$

For a kernelized ridge-regression / GP-style confidence radius, a standard upper bound can be written as

$$U_t^{(c)} \;=\; \beta_t \sqrt{\frac{2\gamma_t}{t}}, \tag{13}$$

where $\beta_t = B + R\sqrt{2(\gamma_t + \ln(1/\delta_t))}$ is the exploration factor as defined before. Assume the kernel class satisfies the growth condition:

$$\gamma_t \le C_\gamma \big(\log(1 + t/\lambda)\big)^j \qquad \forall t \ge 1, \tag{14}$$

for some kernel-dependent constants $C_\gamma > 0$ and $j \geq 1$. Let $L_t := \log(1 + t/\lambda)$. Then $\sqrt{\gamma_t} \leq \sqrt{C_\gamma}\, L_t^{j/2}$. Moreover, for the usual choice $\delta_t = O(1/t^2)$, one has $\log(1/\delta_t) = O(\log t) = O(L_t)$, so $\beta_t$ is at most polynomial in $L_t$. Plugging into (13) and collecting the largest exponent in $L_t$ gives that there exists a constant $C_c > 0$ such that for all $t \geq 2$,

$$U_t^{(c)} \;\leq\; C_c \,\frac{L_t^j}{\sqrt{t}} \;\leq\; C_c \,\frac{(\log t)^j}{\sqrt{t}}. \tag{15}$$

then $U_t^{(c)} \leq e_s$ holds once

$$C_c \,\frac{(\log T)^j}{\sqrt{t}} \leq e_s \quad\Rightarrow\quad t \geq \left(\frac{C_c(\log T)^j}{e_s}\right)^2$$

which again implies that

$$n_s \leq n_s^{(c)} := \left(\frac{C_c(\log T)^j}{e_s}\right)^2$$

Since we need both to be $\leq e_s$ we take $n_s \leq \max\{n_s^{(p)}, n_s^{(c)}\}$. Hence for some constant $\tilde{C}_1$

$$n_s \leq \tilde{C}_1 \frac{(\log T)^{2j}}{e_s^2}$$

We have from (StageReg) that

$$\text{Regret}_s \leq C_0 n_s e_s \leq C_2 \frac{(\log T)^{2j}}{e_s}$$

Now summing over stages $s = 1, \ldots, S$ where $S$ is the final stage index by time $T$:

$$\text{Regret}(T) = \sum_{s=1}^{S} \text{Regret}_s \leq C_2(\log T)^{2j} \sum_{s=1}^{S} \frac{1}{e_s}$$

Now relate $e_s$ to $T$ using the fact that the total time is $T$:

$$T = \sum_{s=1}^{S} n_s \leq \sum_{s=1}^{S} \tilde{C}_1 \frac{(\log T)^{2j}}{e_s^2}$$

and since $e_s = 2^{-2^s}$ which is a fast decrease, the sum is dominated by the stage hence

$$T \lesssim \tilde{C}_1 \frac{(\log T)^{2j}}{e_S^2} \quad\Rightarrow\quad e_S \lesssim \sqrt{\frac{\tilde{C}_1(\log T)^{2j}}{T}}$$

Similarly $\sum_{s\leq S} 1/e_s$ is dominated by $1/e_S$

$$\sum_{s=1}^{S} \frac{1}{e_s} \lesssim \frac{1}{e_S} \lesssim \sqrt{\frac{T}{\tilde{C}_1(\log T)^{2j}}}$$

Hence we have the following bound on the regret over $T$ rounds:

$$\text{Regret}(T) \leq C_2(\log T)^{2j} \sqrt{\frac{T}{\tilde{C}_1(\log T)^{2j}}} = \tilde{O}(\sqrt{T})$$

Finally, we account for the algorithmic switching costs. Let $c_{\max} \leq 1$ denote the maximum cost of querying a single item, and let $k$ be the cache capacity. During each stage switch, updating the cache from $\mathcal{M}_{s-1}$ to $\mathcal{M}_s$ incurs a maximum switching cost of $k \cdot c_{\max}$. By Lemma A.6, the algorithm performs at most $S(T) = O(\log\log T)$ switches. Thus, the cumulative switching cost over $T$ rounds is bounded by:

$$\text{Total Switching Cost} \leq S(T) \cdot k \cdot c_{\max} = O(k \log\log T)$$

Adding this to our instantaneous regret bound yields the final cumulative regret:

$$\text{Regret}(T) \leq \tilde{O}(\sqrt{T}) + O(k \log\log T) = \tilde{O}(\sqrt{T})$$

□

**Algorithm 6** CLCB-Frozen-Cont

**Require:** Cache size $k$, kernel $k(\cdot,\cdot)$, ridge $\lambda$, distance $d(\cdot,\cdot)$, mismatch factor $\phi$, horizon $T$, confidence schedule $\{\delta_t\}$, Lipschitz constant $L_g$, constant $C_p$

**Initialize:** $X_{\text{obs}}, Y_{\text{obs}}, \mathcal{U}, \bar{\mathcal{U}}, \mathcal{M}_0 \leftarrow \emptyset$; Stage $s \leftarrow 1$; Length $n \leftarrow 0$; SWITCH $\leftarrow$ True
**Tolerances and Radius:** $e_s \leftarrow 2^{-2^s}$; $\rho_s \leftarrow e_s/(2L_g)$; **Counts:** $\forall u \in \bar{\mathcal{U}}, N(u) \leftarrow 0, N(\perp) \leftarrow 0$
**for** $t = 1, 2, \ldots, T$ **do**
  Receive query $x_t$; $\mathcal{U} \leftarrow \mathcal{U} \cup \{x_t\}$
  $(N(\cdot), n, \hat{w}_t(\cdot), \text{SWITCH}, U_t^{(p)}) \leftarrow$
    **Handle_Stage_Frozen**$(x_t, \bar{\mathcal{U}}, n, e_s, t, \delta, \rho_s, C_p, N(\cdot))$
  **if** SWITCH **then** ▷ **Switch_Cache_Continuous**
    $\bar{\mathcal{U}} \leftarrow \mathcal{U}$;  $\forall u \in \bar{\mathcal{U}}, w^{\text{opt}}(u) \leftarrow \hat{w}_t(u)/(\sum_{v\in\bar{\mathcal{U}}} \hat{w}_t(v))$
    $\mathcal{M}_t \leftarrow$ `Cont_Reverse_Greedy`$(\bar{\mathcal{U}}, w^{\text{opt}}, \hat{c}_t^{\text{LCB}}, k, d, \phi)$
    Query LLM for $u \in \mathcal{M}_t \setminus \mathcal{M}_{t-1}$ and store responses
    $s \leftarrow s+1$;  $e_s \leftarrow 2^{-2^s}$;  $\rho_s \leftarrow e_s/(2L_g)$;  $n \leftarrow 0$; SWITCH $\leftarrow$ False
    Reset counts: $\forall u \in \bar{\mathcal{U}}, N(u) \leftarrow 0$;  $N(\perp) \leftarrow 0$
  **else**
    $\mathcal{M}_t \leftarrow \mathcal{M}_{t-1}$
  **end if**
  $X_{\text{obs}}, Y_{\text{obs}} \leftarrow$ **Exec_Frozen_Act**$(x_t, \mathcal{M}_t, \hat{c}_{t-1}^{\text{LCB}}, \phi, d, X_{\text{obs}}, Y_{\text{obs}})$
**end for**

**Algorithm 7** Execute_Frozen_Action

**Require:** Query $x_t$, Current Cache $\mathcal{M}_t$, LCB costs $\hat{c}_t^{\text{LCB}}$, mismatch factor $\phi$, distance $d$, Observations $X_{\text{obs}}, Y_{\text{obs}}$

Let $d(x_t, \mathcal{M}_t) = \min_{u\in\mathcal{M}_t} d(x_t, u)$ (where $d(x_t, \emptyset) = \infty$)
**if** $\hat{c}_t^{\text{LCB}}(x_t) \leq \phi\,(d(x_t, \mathcal{M}_t))$ **then**
  Query LLM on $x_t$, observe true cost $C_t(x_t)$
  $X_{\text{obs}} \leftarrow X_{\text{obs}} \cup \{x_t\}$
  $Y_{\text{obs}} \leftarrow Y_{\text{obs}} \cup \{C_t(x_t)\}$
**else**
  Serve cached response at $s(x_t, \mathcal{M}_t) = \arg\min_{u\in\mathcal{M}_t} d(x_t, u)$
  Carry over unmodified: $X_{\text{obs}} \leftarrow X_{\text{obs}}$, $Y_{\text{obs}} \leftarrow Y_{\text{obs}}$
**end if**
**return** $X_{\text{obs}}, Y_{\text{obs}}$

**Algorithm 8** Handle_Stage_Frozen

**Require:** Query $x_t$, Frozen pool $\bar{\mathcal{U}}$, stage length $n$, round $t$, stage tolerance $e_s$, confidence $\delta$, dynamic radius $\rho_s$, constant $C_p$, boundary counts $N(\cdot)$

$n \leftarrow n + 1$
**if** $\bar{\mathcal{U}} \neq \emptyset$ **and** $\min_{u \in \bar{\mathcal{U}}} d(x_t, u) \leq \rho_s$ **then**
  $z_t \leftarrow \arg\min_{u \in \bar{\mathcal{U}}} d(x_t, u)$
**else**
  $z_t \leftarrow \perp$
**end if**
Update count: $N(z_t) \leftarrow N(z_t) + 1$
Compute empirical weights: $\forall u \in \bar{\mathcal{U}}, \hat{w}_t(u) \leftarrow N(u)/n$; $\hat{w}_t(\perp) \leftarrow N(\perp)/n$
Compute local stage bounds: $m_s \leftarrow |\bar{\mathcal{U}}| + 1$
$U_t^{(p)} \leftarrow C_p \sqrt{\frac{m_s \log 2 + \log(2t^2/\delta)}{n}}$
SWITCH $\leftarrow$ False
$U_t^{(c)} \leftarrow \beta_t \sqrt{\frac{2\gamma_t}{t}}$
**if** $U_t^{(c)} \leq e_s$ **and** $U_t^{(p)} \leq e_s$ **then**
  SWITCH $\leftarrow$ True
**end if**
**return** $N(\cdot), n, \hat{w}_t(\cdot)$, SWITCH

## A.3 Additional Experimental Results

In this section, we include and discuss additional empirical results we omitted from the main paper due to a lack of space. We first give details about the stream generation we applied for the NQ+TriviaQA dataset.

To accurately simulate the non-uniform, heavy-tailed arrival patterns often observed in real LLM systems, we introduced a skewed popularity distribution when sampling from the combined dataset. Let $Q_D$ denote the set of all queries in dataset $D$, where $D \in \{\text{NQ, TriviaQA}\}$. For each query $q_i \in Q_D$, we independently sample a raw popularity score $s_i$ from a standard log-normal distribution:

$$s_i \sim \text{Lognormal}(\mu = 0, \sigma = 1)$$

These raw scores are then normalized over the respective dataset to obtain the final sampling probability $p_i$ for each query $q_i$:

$$p_i = \frac{s_i}{\sum_{q_j \in Q_D} s_j}$$

During the stream generation, when a burst is assigned to dataset $D$, the next arriving query is sampled with replacement from $Q_D$ according to the probability mass function defined by $p_i$. This log-normal formulation ensures that a small subset of queries appears with high frequency, while the long tail of the dataset consists of unique or rare queries, presenting a rigorous challenge for the caching algorithms.

To evaluate our incoming data stream generation, we applied two dimensionality reduction techniques to visualize the incoming data streams in Fig. 3 and Fig. 4. Principal Component Analysis (PCA) was used to map synthetic cluster centers and simulated stream points onto a two-dimensional manifold. The generated stream points remain closely grouped around their designated universe cluster centers (marked with an 'X'), which verifies that the incoming data behaves as expected without drifting. Second, Uniform Manifold Approximation and Projection (UMAP) was utilized to map the more complex structures of the combined dataset. The UMAP visualization shows a clear separation between the two sources, naturally separating the NQ data into one distinct region and the TriviaQA data into another. Importantly, both figures are demonstrating that the sampled stream data accurately overlaps with the underlying background data. This provides strong verification that our streaming code successfully maintains the original structural patterns of the datasets and these two plots also show that our Assumption 2 is justified.

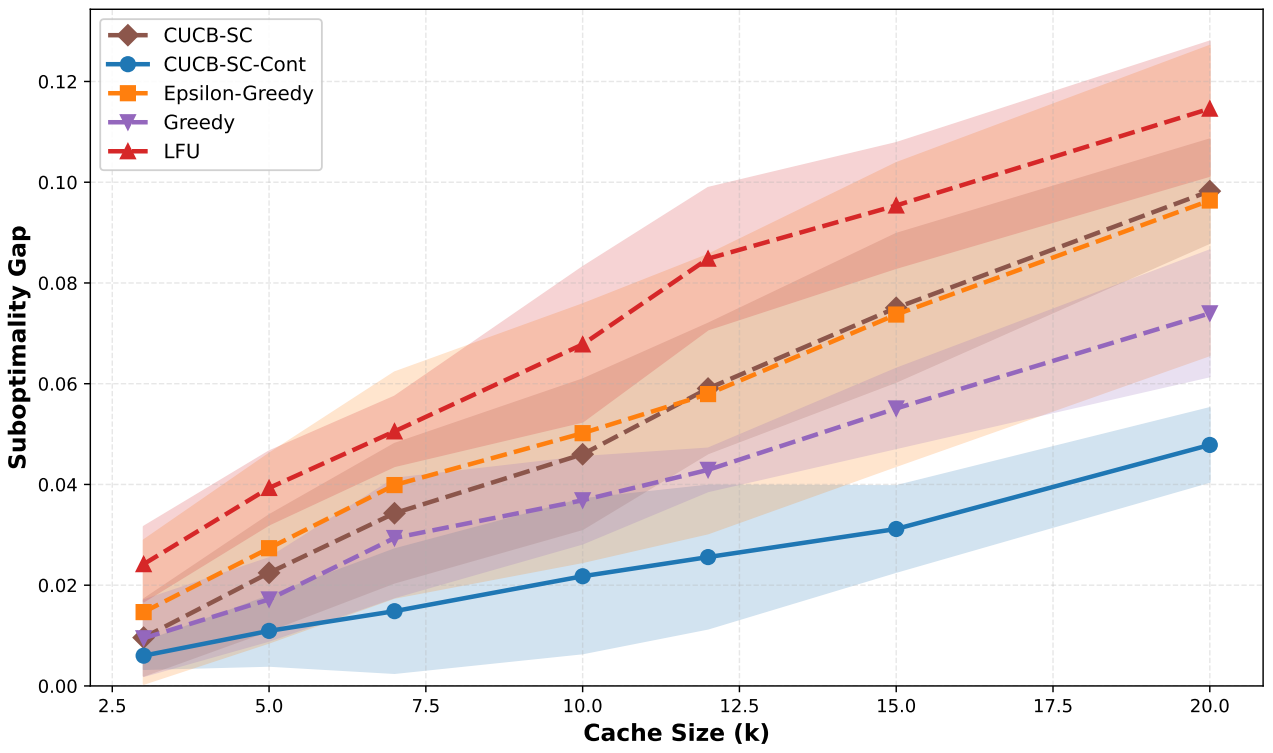

**Figure 5: Cache size vs Suboptimality Gap Ablation for the Offline setting**

To evaluate how cache size affects system performance in the offline, we conducted an ablation study by varying the cache size ($k$) from 3 to 20 items. The experiment simulated a stream of data over 500 time steps ($t = 500$) for the synthetic dataset. For a fair comparison, we held hyperparameters constant across the trials. The results of this study are illustrated in Fig. 5 and Fig. 6. The True Loss figure shows that as the cache size increases, the overall loss drops consistently for all methods. This is an expected result, as a larger cache can hold more items simultaneously, which naturally lowers the penalty of missing a needed item. CUCB-SC-Cont maintains the lowest absolute loss across every tested cache size. The Suboptimality Gap figure provides deeper insight into how efficiently each method uses this extra capacity. As the cache grows, the task of selecting the best set of items to cache becomes much harder. Consequently, the gap between the algorithms and the theoretical Oracle increases across the board. However, CUCB-SC-Cont proves to be much more robust to this increasing complexity. Its suboptimality gap grows at a significantly slower rate compared to other baselines.

Now, we present again the results in Fig. 2 by adding the performance of the CLCB-Frozen-Cont algorithm in the online setting for the synthetic and combined datasets which can be seen in Fig. 7. Subplots (c) and (d) show that the Frozen algorithm is highly competitive with the discrete baseline in the synthetic setting both in terms of final average regret and runtime, while achieving a higher final average regret but a much lower runtime in the combined NQ+TriviaQA dataset. Notably, the Frozen baseline also has much lower variance in its final average regret compared to the other baselines, indicated by the small vertical error bars.

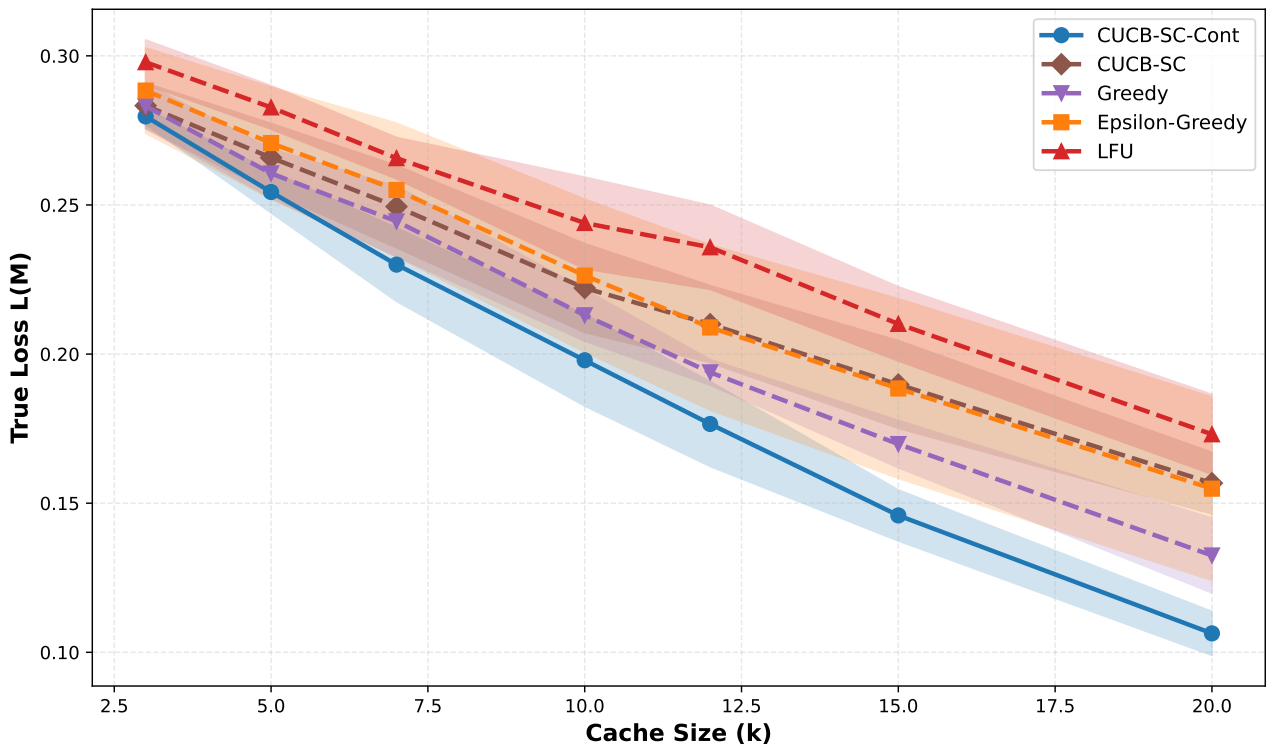

**Figure 6: Cache size vs Absolute Loss Ablation for the Offline setting**

To better understand the geometry of the embedding space for the simulated synthetic dataset, we analyze several distance distributions between query embeddings. We simulate a stream by sampling queries from the original query universe and perturbing their embeddings with Gaussian noise, thereby creating semantically similar but previously unseen inputs. We then compute pairwise distances among stream queries (Fig. 8), pairwise distances in the original query universe (Fig. 9), nearest-neighbor distances within the stream (Fig. 10), and the distance from each stream query to its closest point in the universe (Fig. 11). The resulting distributions reveal the structure that while global pairwise distances in both the stream and universe are large and concentrated (median $\approx$ 0.88), indicating that most queries are well-separated in the embedding space, local distances are significantly smaller, with median nearest-neighbor distances around 0.074. Moreover, the distance from unseen queries to their nearest universe neighbor shows a similar scale (median $\approx 0.074$), confirming that the jittered queries remain close to existing semantic clusters. This contrast between large global distances and small local neighborhoods justifies the use for $\epsilon$-net–based discretization: a moderate $\epsilon$ can effectively cover local neighborhoods while still preserving global diversity.

Next, we consider how the cost prediction error evolves over time for the synthetic dataset, measured using RMSE, MAE, and maximum absolute error in Fig. 12. Initially, all error metrics are relatively high due to limited observations, but they decrease rapidly as more data is collected and the model learns the underlying cost function. After this phase, the errors stabilize at low levels, indicating that the model has converged to accurate predictions. The RMSE and MAE remain consistently small, reflecting good average performance, while the maximum error exhibits larger fluctuations early on before settling, suggesting that worst-case predictions improve more slowly but eventually stabilize. The cost estimation error plateaus instead of converging to 0 because the target cost per center is itself evolving over time (since as more arrivals are assigned to a cluster, the cost is the average of all the points in the cluster and these points have slightly different true costs due to different token lengths) and new centers are introduced online.

Next, we look at the token distribution for the synthetic dataset as displayed in Fig. 13 and Fig. 14. We can see that the CDF hits 0.5 around 10 tokens and reaches 1 around 40 tokens. The CDF figure shows that there seems to be two main clusters for the token lengths, concentrated around approximately 0-15 tokens and 25-40 tokens.

The token-length plots (Fig. 15 and Fig. 16) for the combined dataset show a clear difference in query length between the two datasets. Natural Questions queries are substantially longer and more spread out, while TriviaQA queries are shorter and more concentrated as expected since TriviaQA is a reading comprehension dataset whereas Natural Questions is formed by real user questions issued to Google search. This separation is visible both in the histogram and in the CDF, where the combined curve lies between the two dataset curves and reflects the heterogeneous cost structure caused by mixing the two sources.

The embedding-space distance plots (Fig. 17, Fig. 18, Fig. 19 and Fig. 20) reveal that, although most query pairs in the combined dataset are far apart globally, local neighborhoods remain significantly closer. In particular, the pairwise distance distribution is concentrated at relatively large values with median around 0.854, whereas the nearest-neighbor distances are much smaller with median around 0.537. This suggests that queries are far apart overall but still form meaningful local clusters, which supports using similarity-based caching and $\epsilon$-net discretization.

Next, we report additional experiments that (i) empirically validate the modeling assumptions underlying our analysis, (ii) quantify the covering complexity of a real query workload, (iii) compare against an additional nearest-neighbor baseline, and (iv) evaluate sensitivity to the choice of embedding model. Unless otherwise stated, all experiments use 5K real queries from Natural Questions (NQ) and TriviaQA, with the same setup as in Section 6.

*A.3.1 Empirical Validation of Modeling Assumptions.* Our approach relies on kernel-based generalization across nearby queries, which is useful only if semantic distance is informative about serving cost. To test this, we sample 200,000 query pairs $(x, y)$ uniformly at random and compute the Spearman rank correlation between the embedding distance $d(x, y)$ and the absolute cost difference $|c(x) - c(y)|$. We obtain $\rho = 0.412$, indicating a clear positive monotone relationship. Grouping the pairs into bins by distance, the mean cost variation $|c(x) - c(y)|$ in the farthest bin is approximately 3× that in the closest bin. Nearby queries therefore have substantially more similar costs than distant ones, supporting distance-based generalization.

Assumption 1 requires the expected cost $g_M$ to be Lipschitz in the embedding distance with constant $L_g$. We estimate the empirical Lipschitz ratio

$$R_M(x, y) = \frac{|g_M(x) - g_M(y)|}{d(x, y)} \quad (16)$$

over 20 million sampled (query pair, cache configuration) tuples $(x, y, M)$. The 95th percentile of $R_M(x, y)$ is 0.645, and more than 99.999% of the sampled tuples satisfy $R_M(x, y) \leq 1$, i.e., the assumption holds with $L_g = 1$. This indicates that, in practice, $L_g$ is upper-bounded by a small constant. Table 1 summarizes these statistics.

*A.3.2 Covering Complexity of Real Query Workloads.* Our $\epsilon$-net is not constructed over the entire embedding space. Representatives are constructed only over the observed query support, and in the online algorithm the representative set grows only when a newly observed query lies at distance more than $\epsilon$ from all existing centers. Consequently, our theoretical bounds are most informative when the query distribution has low *effective* covering complexity, i.e., when the number of representatives remains much smaller than the number of observed queries. We quantify this behavior on NQ+TriviaQA (Table 2).

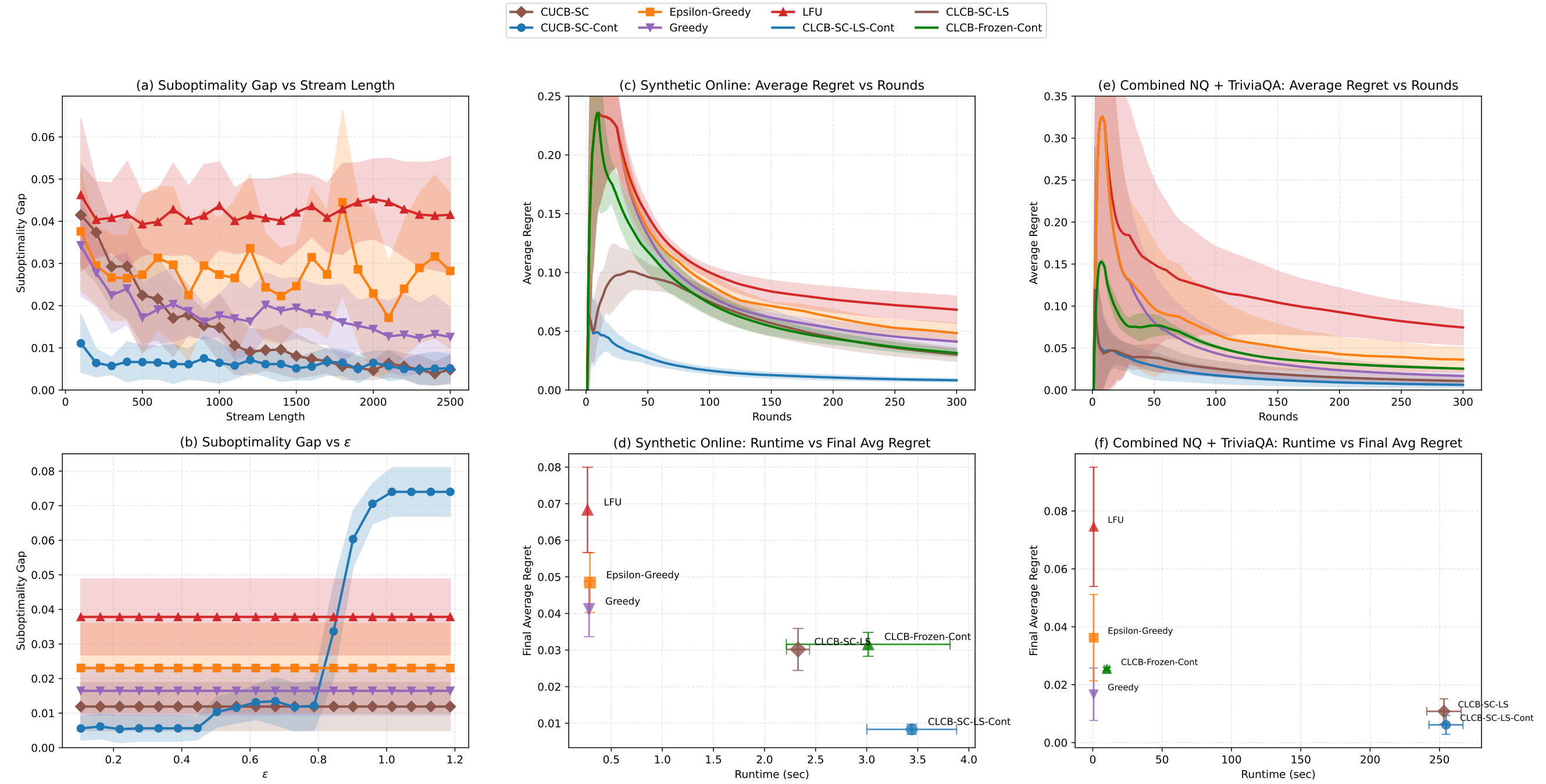

**Figure 7: (a) Variation of the suboptimality gap versus stream length in the offline setting (b) Sensitivity analysis showing the variation of the suboptimality gap with respect to the $\epsilon$-net radius $\epsilon$ (c) Online performance on the synthetic dataset showing the average regret across rounds (d) Efficiency trade-off for synthetic experiments, showing total runtime against the final average regret (e) Online performance on the combined dataset showing the average regret across rounds (f) Efficiency trade-off for real dataset experiments, showing total runtime against the final average regret**

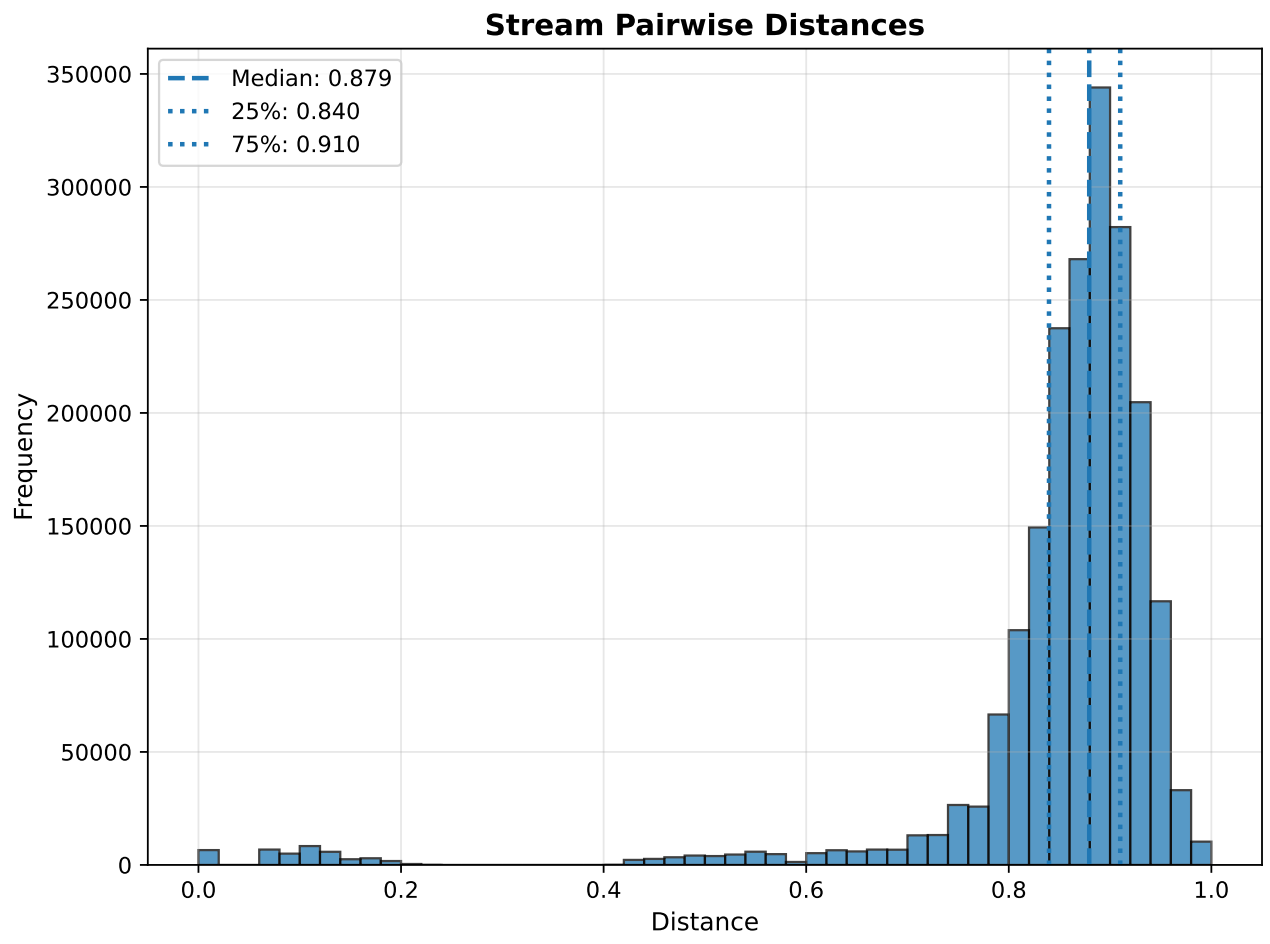

**Figure 8: Distribution of pairwise distances among stream queries for the synthetic dataset.**

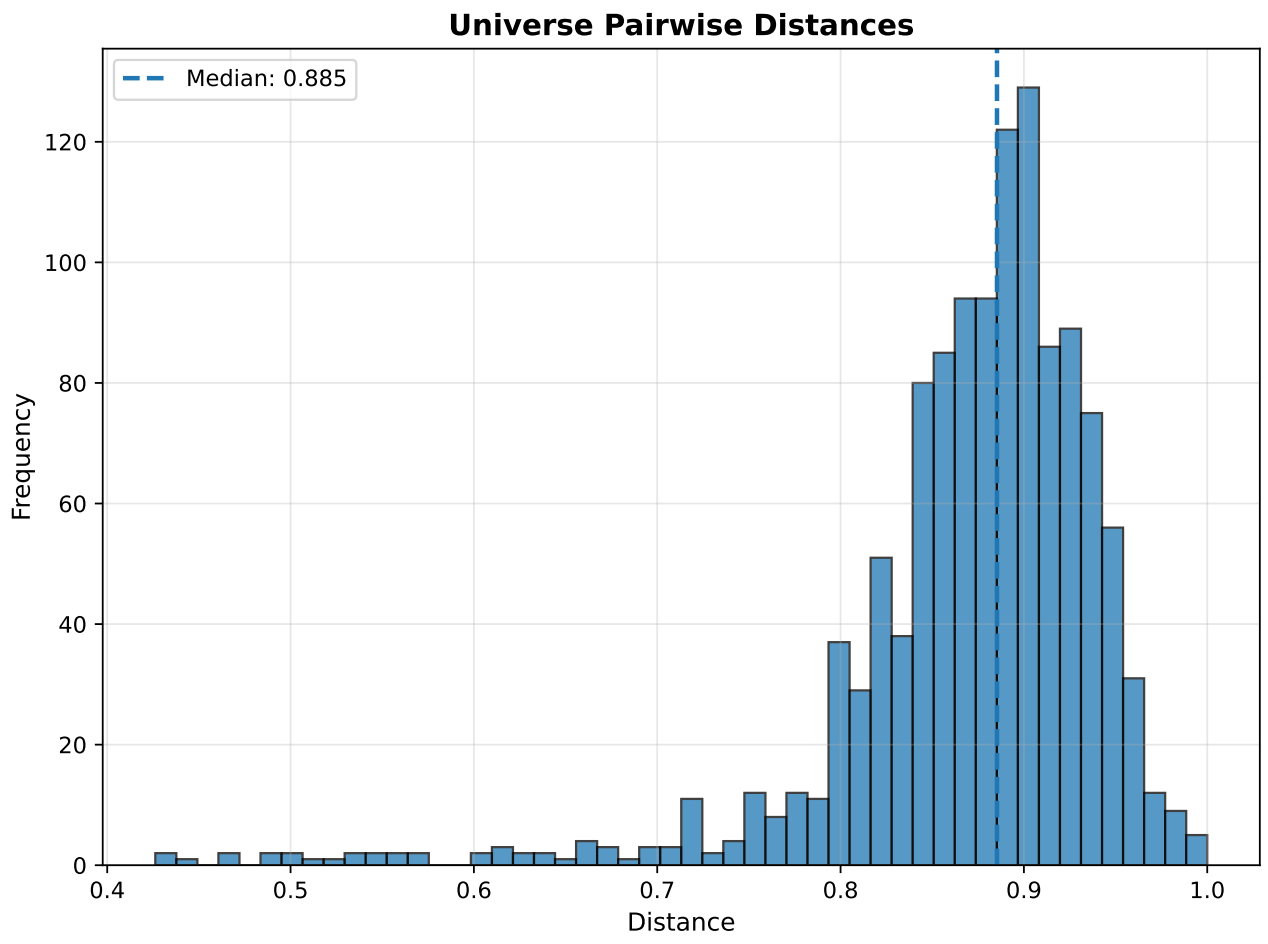

**Figure 9: Distribution of pairwise distances in the original query universe for the synthetic dataset.**

We select the smallest radius achieving at least 90% coverage on a validation set, which gives $\epsilon = 0.575$. At this radius, only 520 representatives (10.4% of the observed queries) suffice for 92% of an independent test set of 2,500 queries to lie within distance $\epsilon$ of some representative.

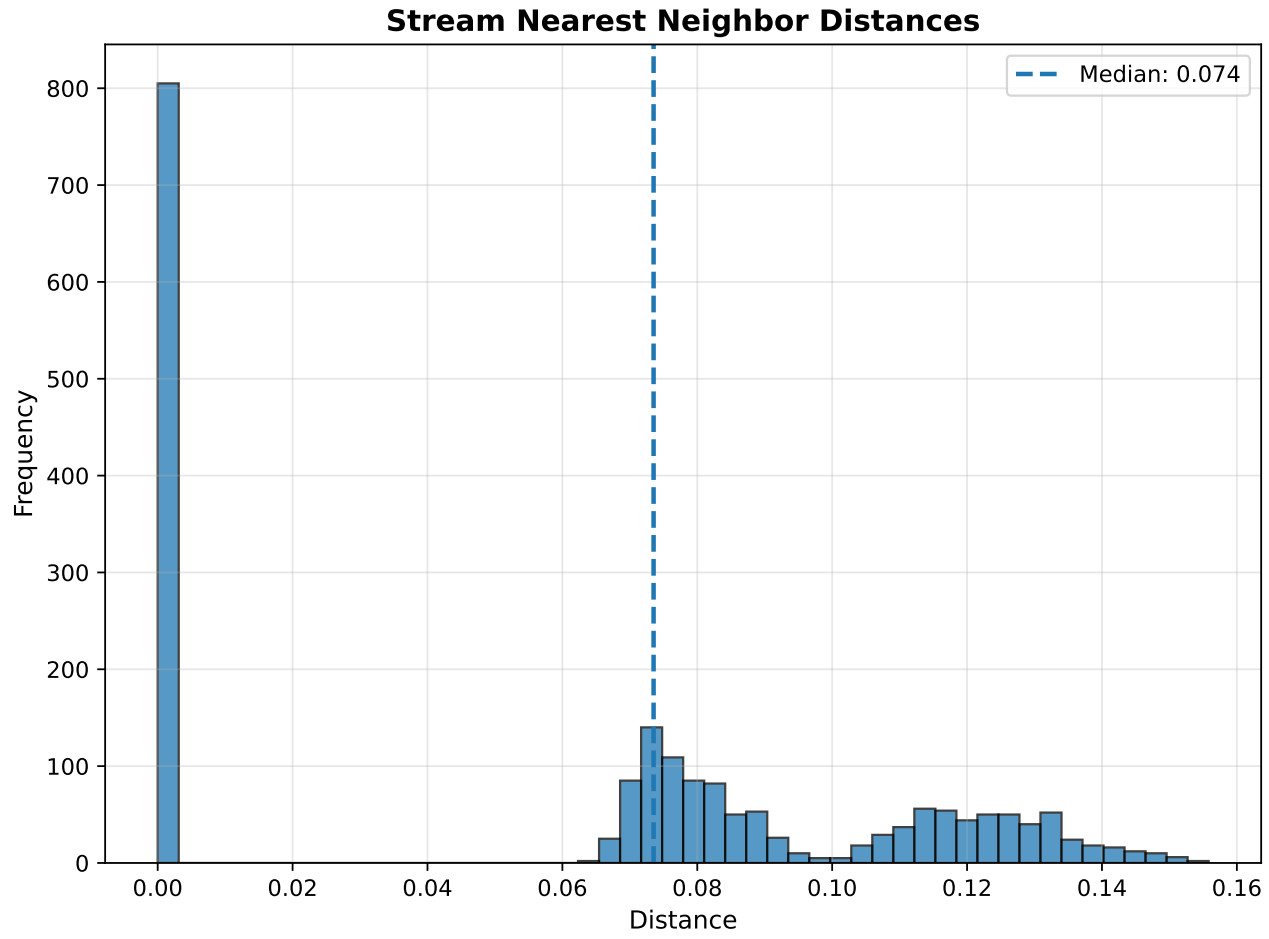

Figure 10: Nearest-neighbor distance distribution within the stream queries for the synthetic dataset.

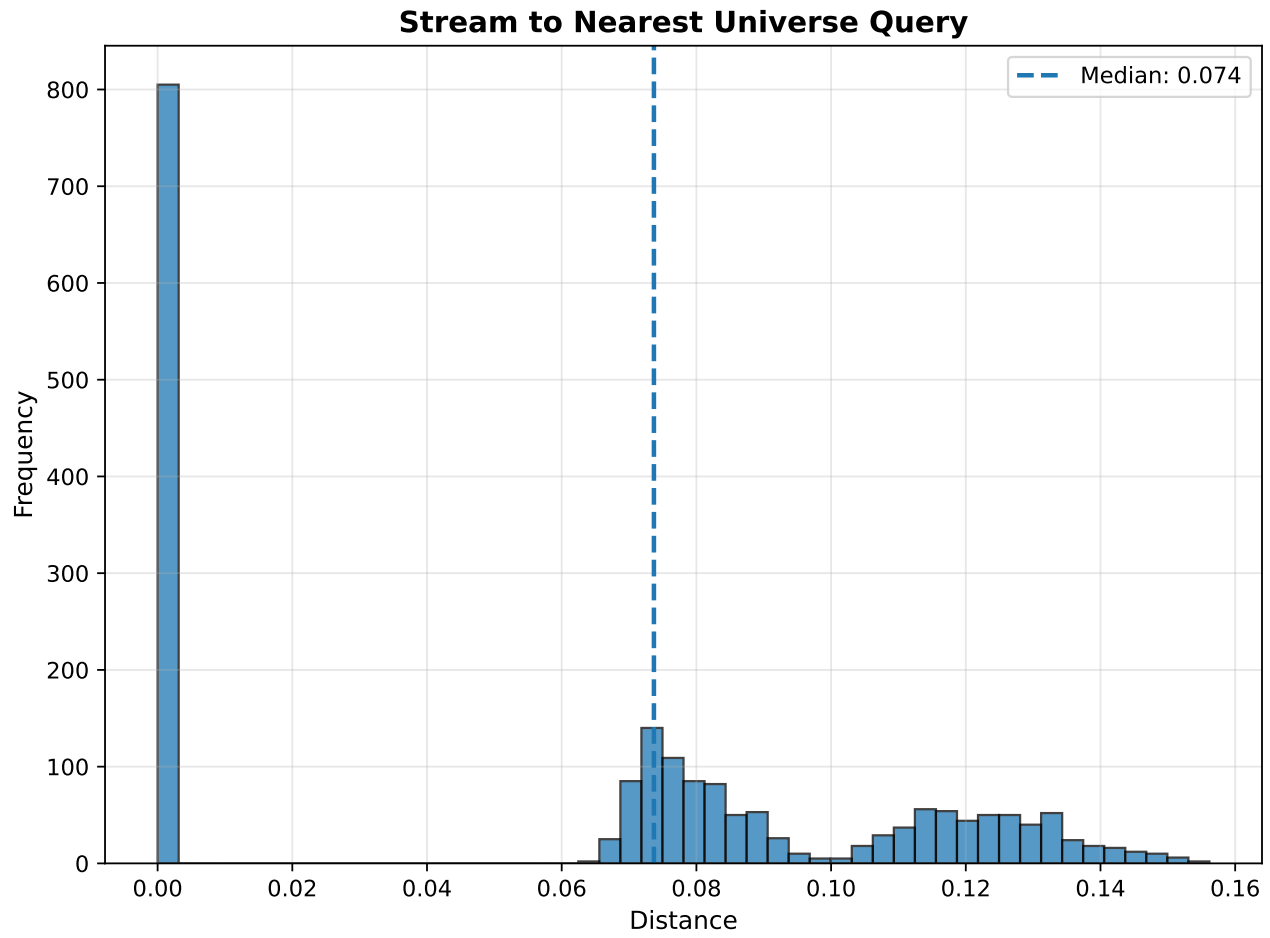

Figure 11: Distance from each stream query to its nearest neighbor in the query universe for the synthetic dataset.

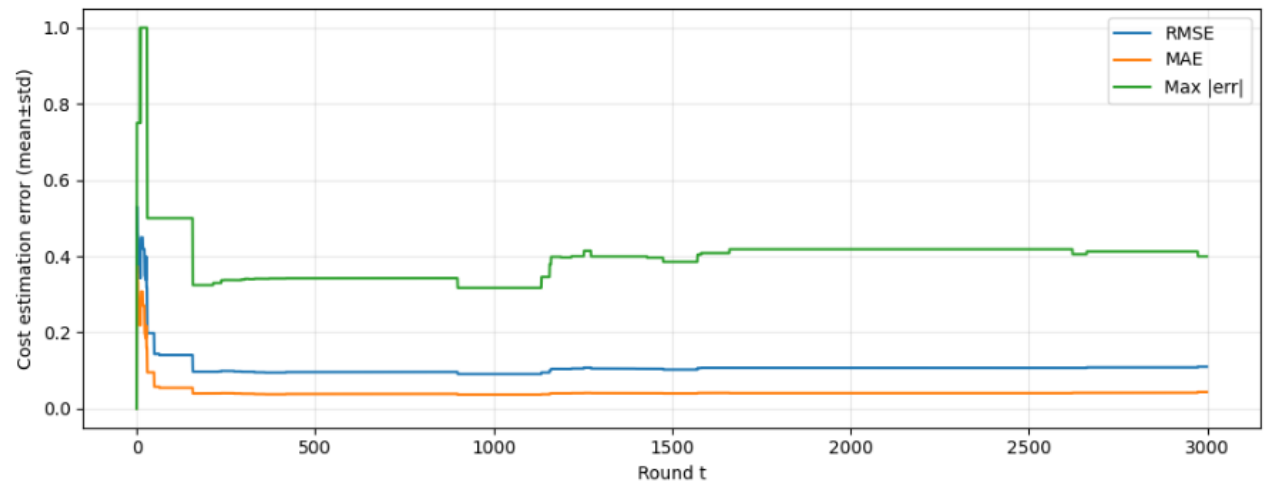

Figure 12: Cost prediction error of the KRR model

A power-law fit of the number of representatives $m(n)$ against the number of observed queries $n$ gives

$$m(n) \approx 3.92\, n^{0.576}, \tag{17}$$

which is substantially sublinear in $n$. Together, these measurements show that on this real workload the effective covering complexity

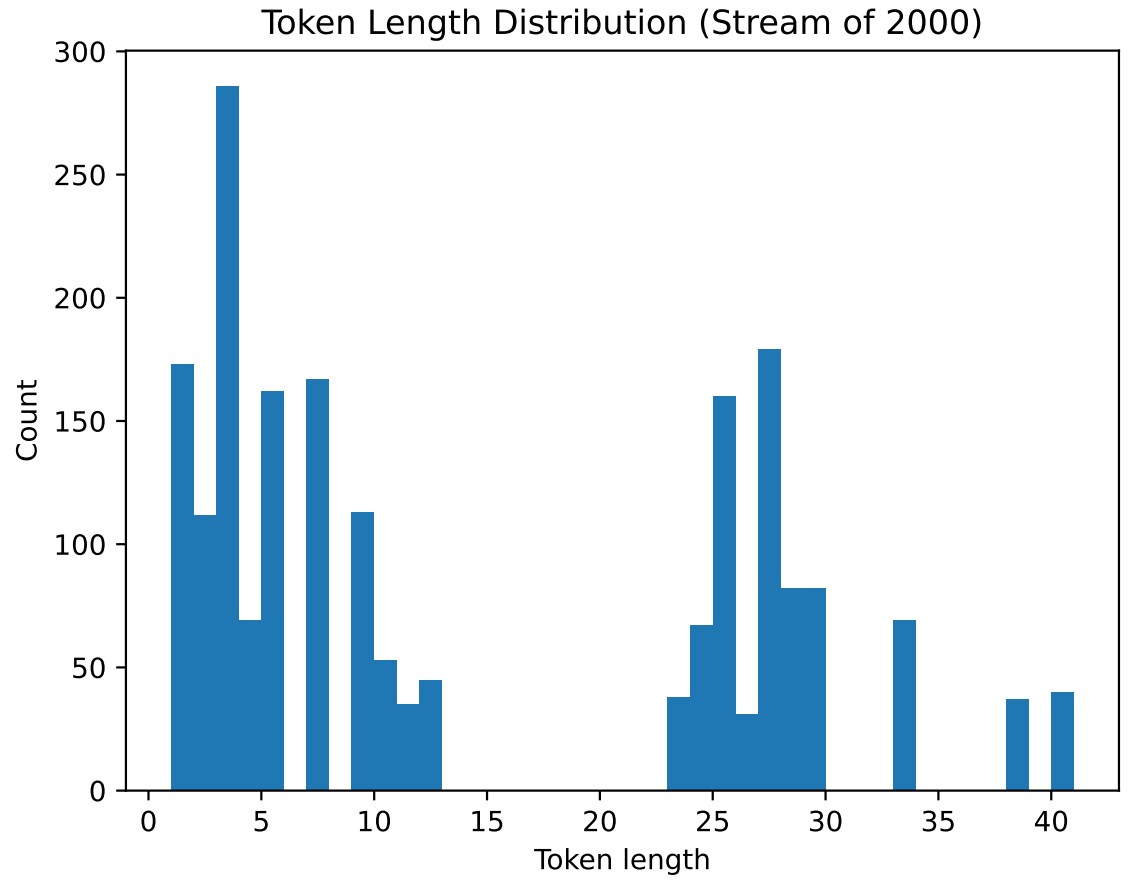

Figure 13: Token length distribution for the synthetic dataset

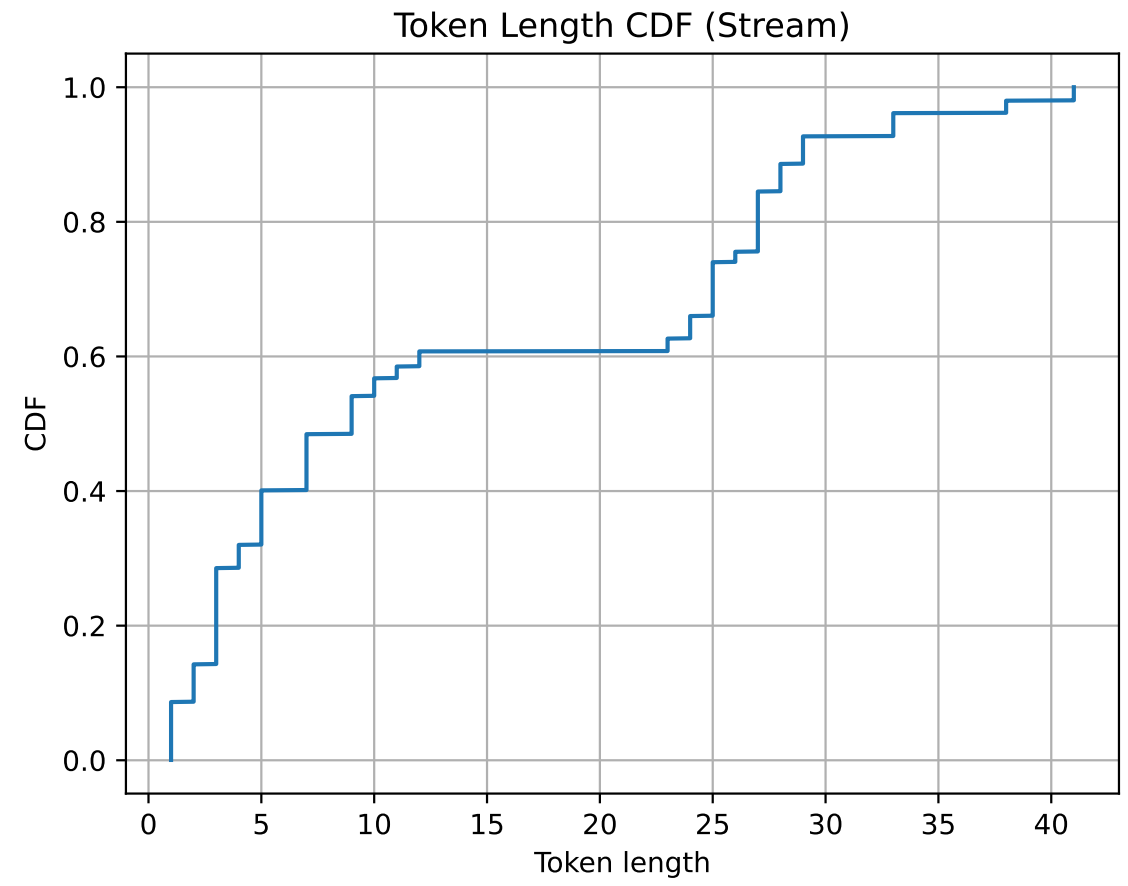

Figure 14: CDF of the Token length distribution for the synthetic dataset

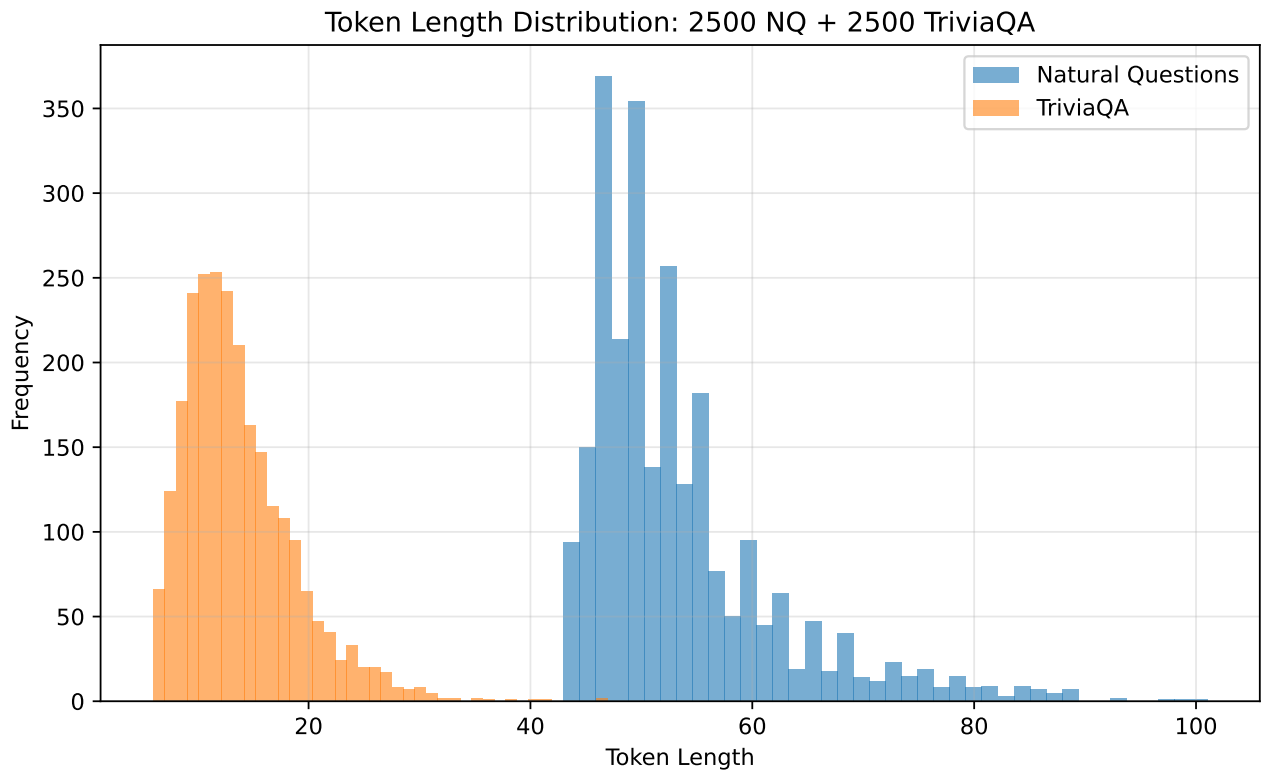

Figure 15: Token length distribution for the combined dataset consisting of 2500 Natural Questions and 2500 TriviaQA queries.

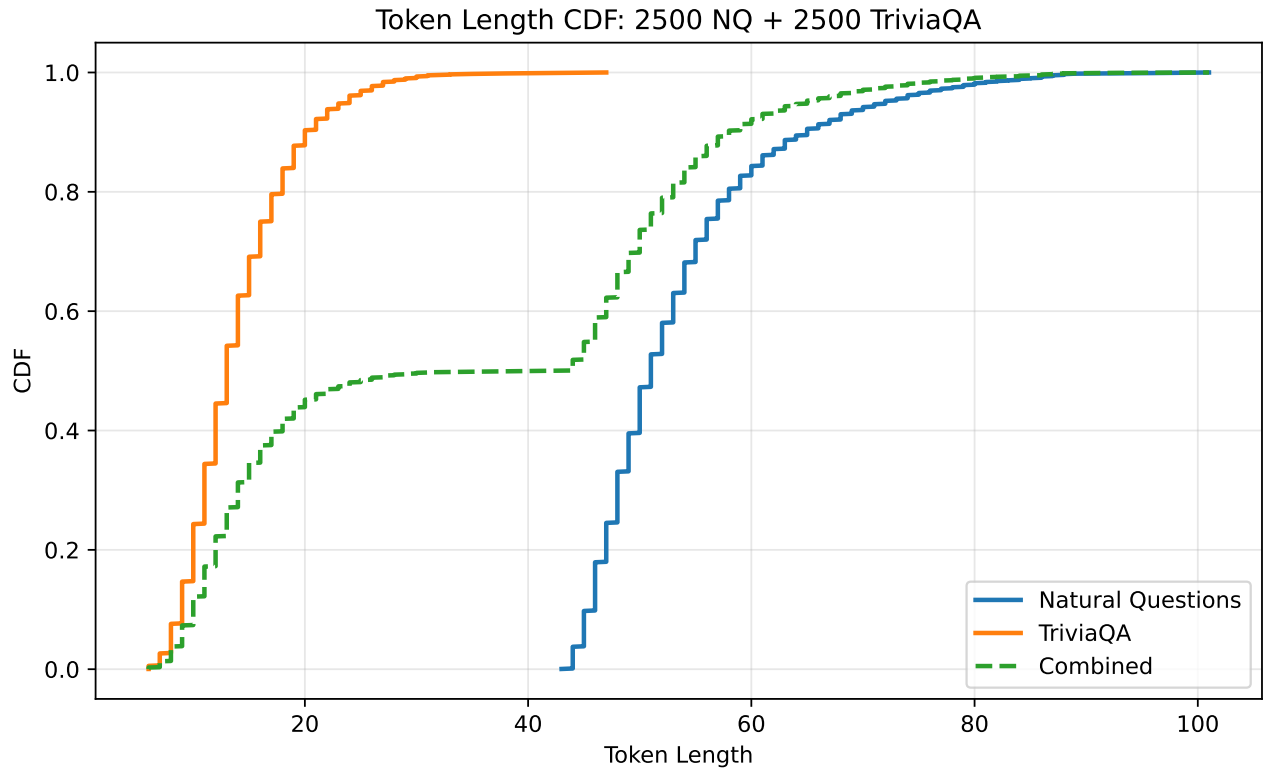

**Figure 16: Token length CDF for Natural Questions, TriviaQA, and the combined dataset.**

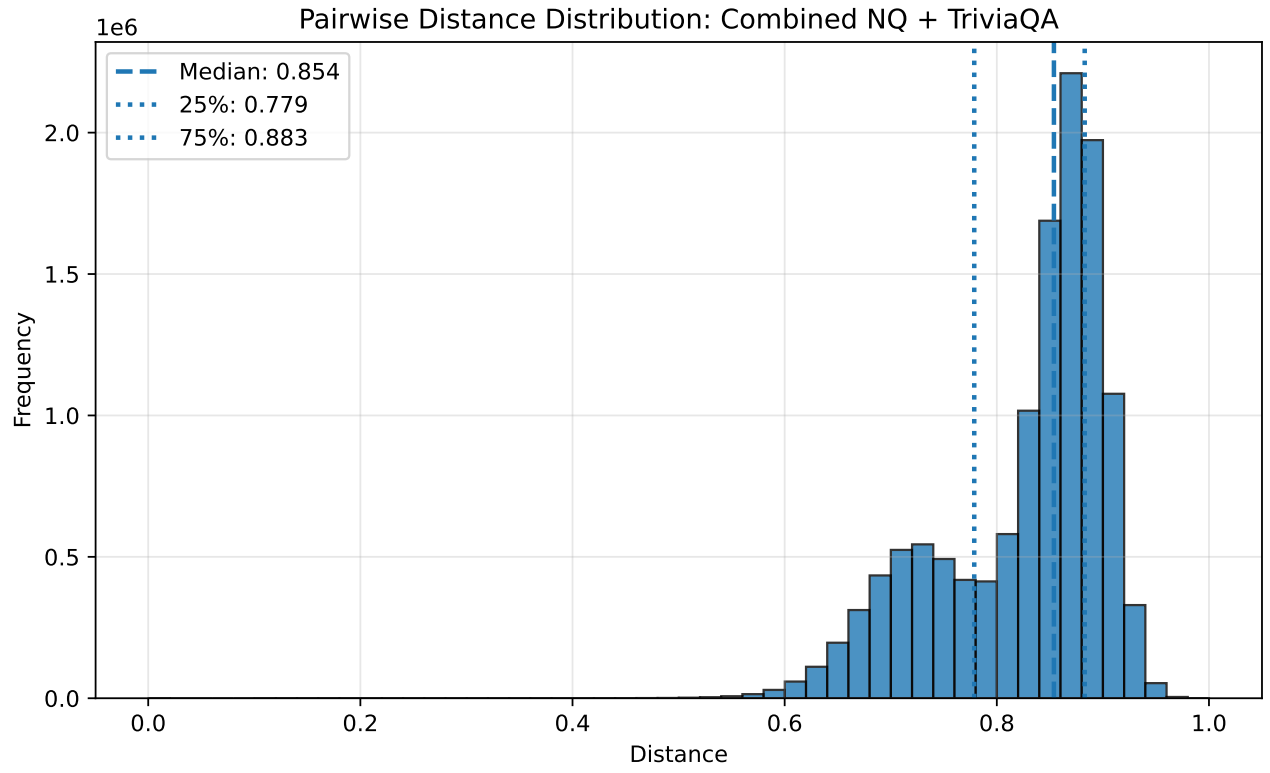

**Figure 17: Pairwise distance distribution in embedding space for the combined dataset.**

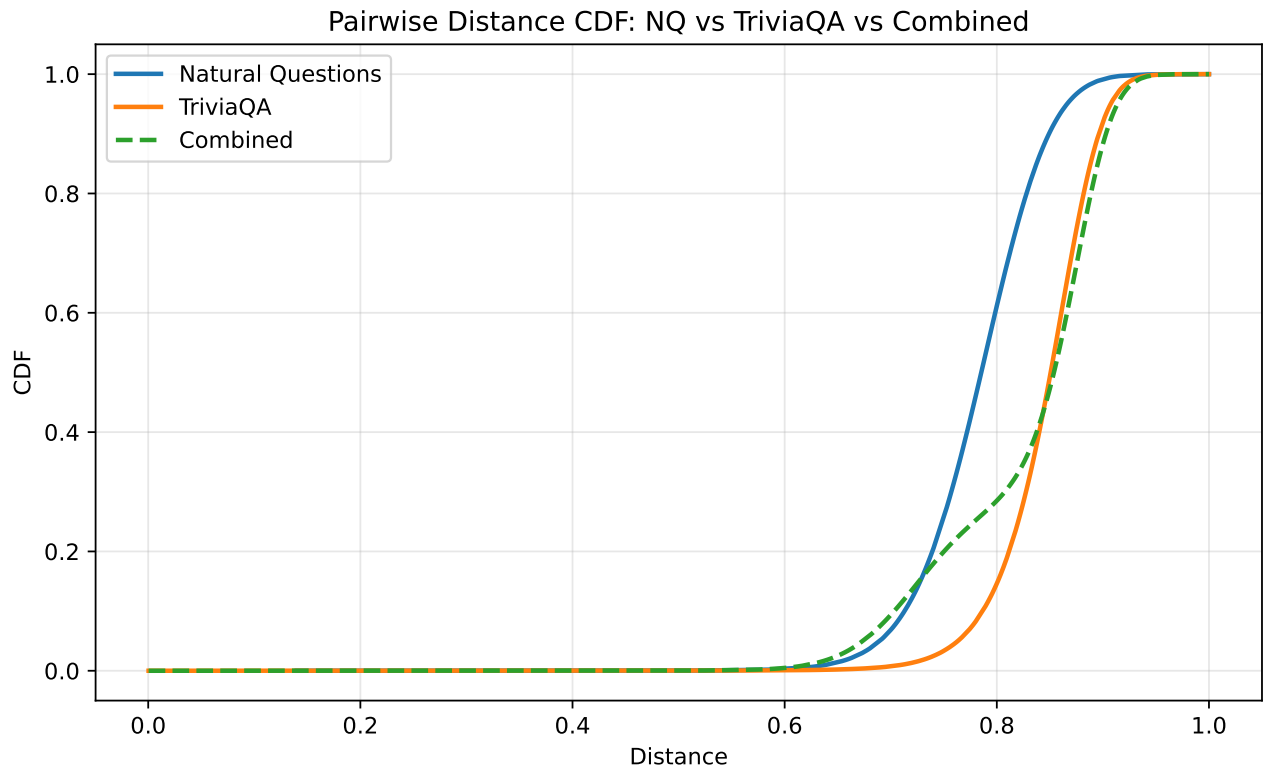

**Figure 18: Pairwise distance CDF in embedding space for Natural Questions, TriviaQA, and the combined dataset.**

is far smaller than that of a worst-case covering of the embedding space.

*A.3.3 Computational Scalability.* Our theoretical analysis uses exact kernel ridge regression (KRR) to obtain rigorous uncertainty bounds. Training exact KRR on $\ell$ observations requires $O(\ell^3)$ time. For computational scalability, our implementation instead uses a

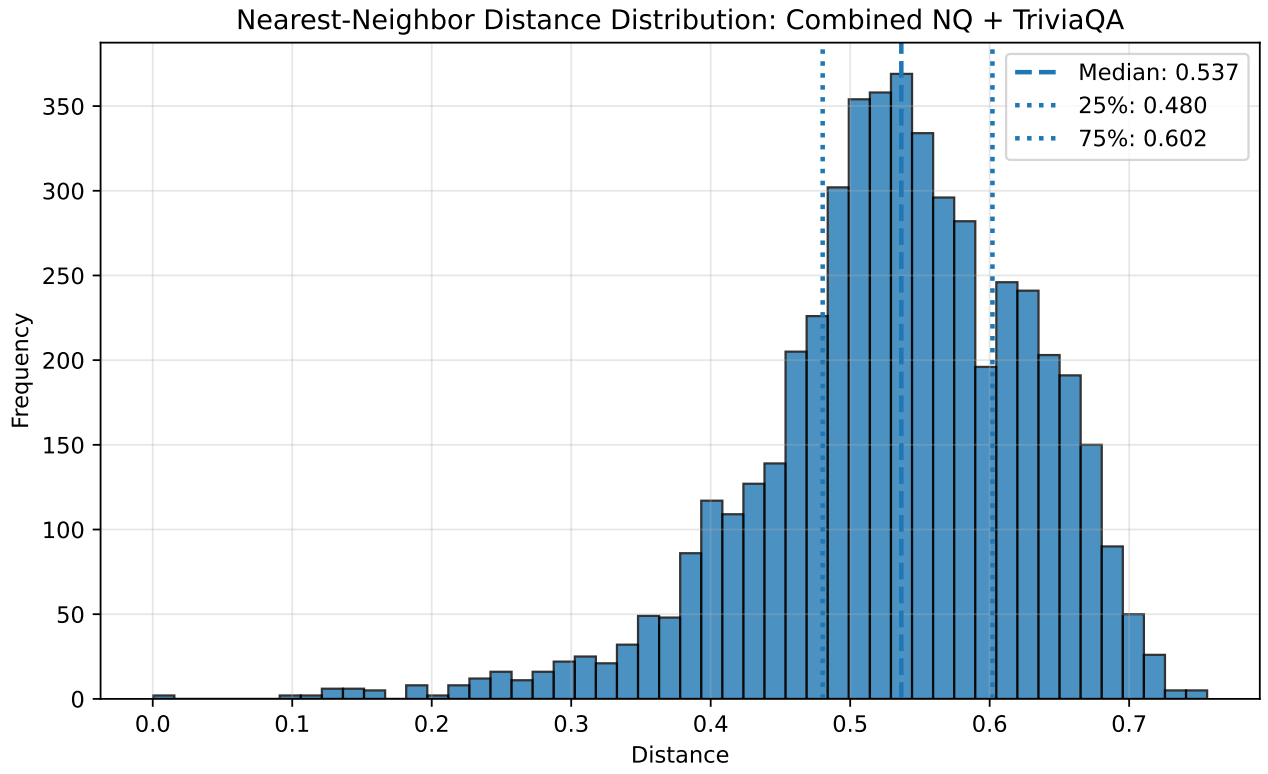

**Figure 19: Nearest-neighbor distance distribution in embedding space for the combined dataset.**

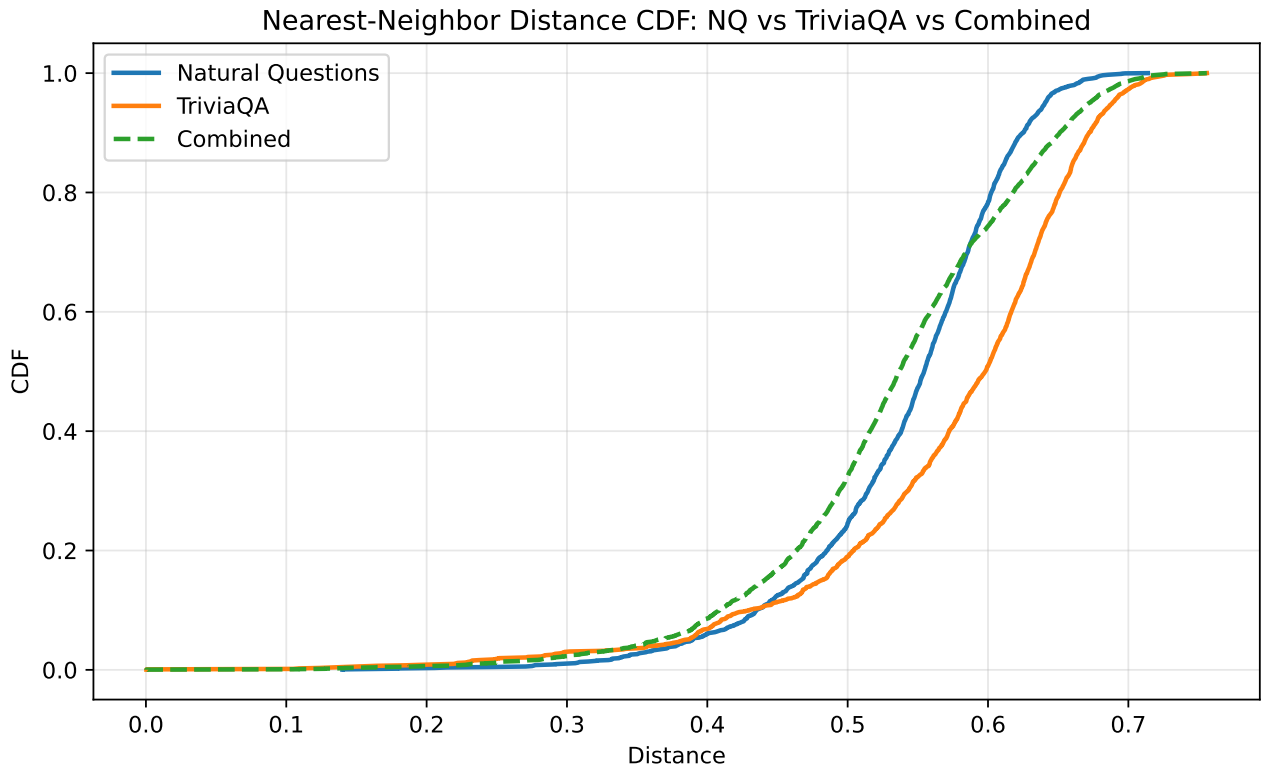

**Figure 20: Nearest-neighbor distance CDF in embedding space for Natural Questions, TriviaQA, and the combined dataset.**

**Table 1: Empirical validation of modeling assumptions on NQ+TriviaQA.**

| Statistic | Value |
|---|---|
| *Distance vs. serving cost (200K query pairs)* | |
| Spearman $\rho\big(d(x,y),\ \lvert c(x)-c(y)\rvert\big)$ | 0.412 |
| Mean $\lvert c(x)-c(y)\rvert$, farthest / closest bin | $\approx 3\times$ |
| *Lipschitz ratio $R_M(x,y)$ (20M tuples)* | |
| 95th percentile | 0.645 |
| Fraction with $R_M(x,y) \le 1$ | $> 99.999\%$ |

**Table 2: Covering complexity of the NQ+TriviaQA query workload.**

| Quantity | Value |
|---|---|
| Radius $\epsilon$ (smallest with $\ge$ 90% val. coverage) | 0.575 |
| Number of representatives | 520 (10.4%) |
| Test coverage (2,500 held-out queries) | 92% |
| Empirical growth $m(n)$ | $3.92\, n^{0.576}$ |

rank-$r$ Nyström approximation of the kernel matrix, which reduces the training time to $O(\ell r^2 + r^3)$ with $r \ll \ell$. Extending our regret guarantees to such low-rank approximations is an interesting direction for future work.

**Table 3: Sensitivity of CLCB-SC-LS-Cont to the embedding model.**

| Encoder | Final avg. regret | Reduction vs. CLCB-SC-LS |
|---|---|---|
| `all-MiniLM-L6-v2` | 0.00613 | 45.5% |
| `multi-qa-MiniLM-L6-cos-v1` | 0.00924 | 44.4% |

*A.3.4 Additional Baseline: Nearest-Neighbor Reuse with LRU Eviction.* We compare against an additional heuristic baseline, NN-LRU, which mirrors the reuse rule of practical semantic caches. For each arriving query, NN-LRU returns the nearest cached response if its embedding distance to the query is below a threshold; otherwise, it calls the LLM and caches the response, evicting the least recently used entry when the cache is full. NN-LRU attains a final average regret of 0.0329, compared to 0.00613 for CLCB-SC-LS-Cont, so our method achieves approximately 5.4× lower regret. The runtime of NN-LRU is similar to that of Greedy.

*A.3.5 Sensitivity to the Embedding Model.* To assess robustness to the choice of embedding model, we replace the general-purpose encoder `all-MiniLM-L6-v2` used in our main experiments with `multi-qa-MiniLM-L6-cos-v1`, an encoder oriented toward question answering and semantic search that is trained under a different objective and data distribution. All other settings are unchanged. As shown in Table 3, CLCB-SC-LS-Cont remains the best-performing method under both encoders, with final average regret of 0.00613 and 0.00924, respectively. Its relative regret reduction over CLCB-SC-LS is also nearly unchanged (45.5% vs. 44.4%), indicating that our conclusions are robust to the choice of embedding model.

*A.3.6 Additional Discussion.*

*Design choices.* We select the discretization radius $\epsilon$ using Fig. 2, which shows that performance degrades when the discretization becomes too coarse. We use the RBF kernel because its dependence on the Euclidean distance between embeddings enables our posterior-width bound; the framework itself, however, supports other kernels. The kernel bandwidth is selected on a validation set.

*Embedding-based similarity.* Embedding distance is not a perfect measure of response equivalence: two prompts that are close in embedding space may still warrant different responses. Nevertheless, embedding-based retrieval is the mechanism used by practical semantic caching systems such as GPTCache [2] and MeanCache [8], which identify reusable responses through embedding similarity. Our framework formalizes this idea by modeling distance through a general embedding mismatch function rather than assuming that nearby prompts are identical.

*Switching cost.* We adopt a conservative switching-cost model in which populating a new cache entry requires incurring its serving cost. If previously generated responses remain available, this term can instead represent a retrieval cost, which is typically smaller. The low-switching objective remains relevant in this case, and our regret analysis can be adjusted to accommodate these lower costs.